\documentclass[10pt,twocolumn,letterpaper]{article}

\usepackage{cvpr}    

\newcommand{\Cow}[1][]{\includegraphics[width=10pt,trim={6cm 9cm 5cm 6cm},clip]{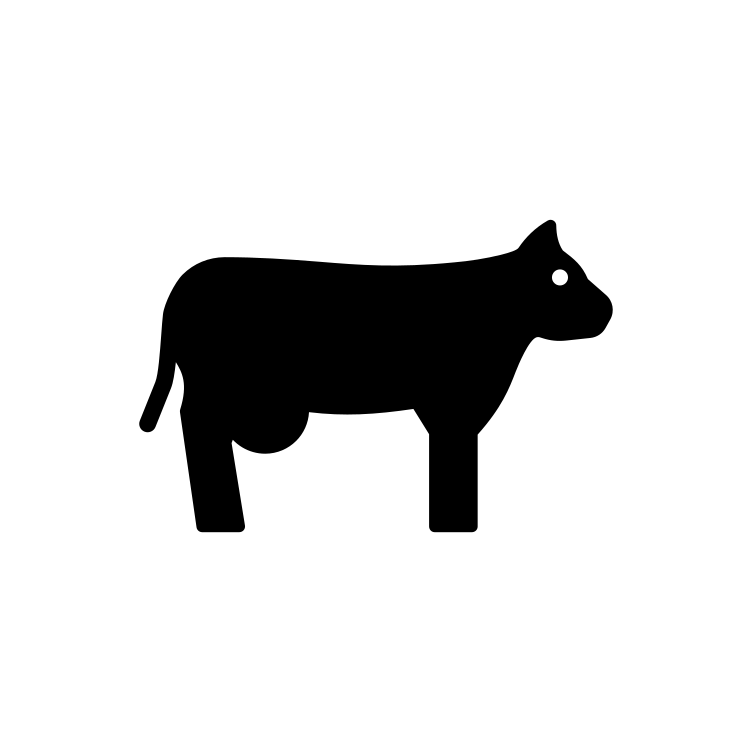}}
\newcommand{\Plant}[1][]{\includegraphics[width=10pt,trim={7cm 6cm 5cm 2cm},clip]{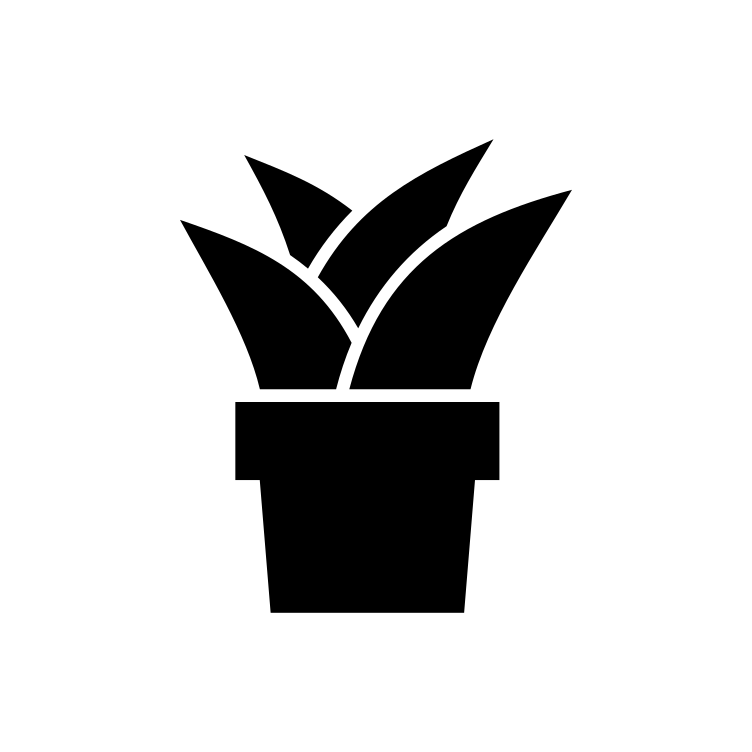}}
\newcommand{\Sheep}[1][]{\includegraphics[width=10pt,trim={6cm 7cm 5cm 6cm},clip]{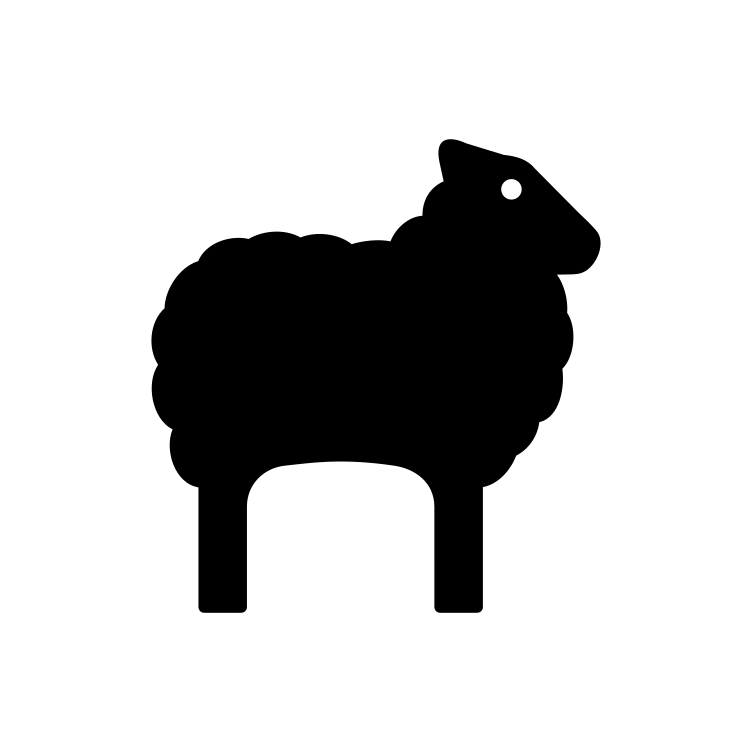}}
\newcommand{\myrowcolour}{\rowcolor[gray]{0.925}}
\newcommand{\best}[1]{\textbf{#1}}
\newcommand{\second}[1]{\underline{#1}}

\newcommand{\oursTab}{{Ours}}
\newcommand{\oursAbbrv}{{SoY}}

\definecolor{cvprblue}{rgb}{0.21,0.49,0.74}
\usepackage[accsupp]{axessibility}
\usepackage[pagebackref,breaklinks,colorlinks,allcolors=cvprblue]{hyperref}
\usepackage{booktabs}
\usepackage{xcolor}
\usepackage{colortbl}
\usepackage{fontawesome5}
\usepackage{graphicx}
\usepackage{multirow}
\usepackage{cuted}
\usepackage{capt-of}
\usepackage{algorithm}
\usepackage{algpseudocode}
\usepackage{listings}
\usepackage{adjustbox}
\usepackage{array}
\newcolumntype{R}{c}
\usepackage{makecell}
\usepackage{amssymb}
\usepackage{mathtools}
\usepackage{tikz}
\usepackage{xspace}
\usetikzlibrary{decorations.pathreplacing,calligraphy}

\large

\newcommand{\soyicon}{%
 \raisebox{-0.6ex}{\includegraphics[height=1.3em]{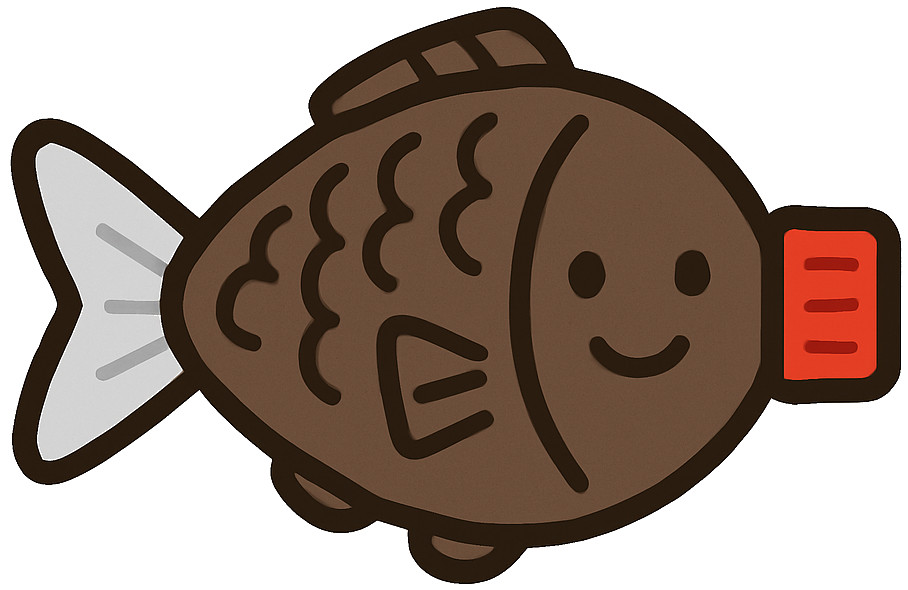}}%
 \xspace
}

\def\eqref#1{(\ref{eq:#1})}
\def\eqlabel#1{\label{eq:#1}}

\def\norm#1{\left\lVert#1\right\rVert}
\def\l2#1{\norm{#1}_2}

\title{\soyicon~~Shape-of-You: Fused Gromov-Wasserstein Optimal Transport for Semantic Correspondence in-the-Wild}

\author{
Jiin Im$^{1}$ \quad Sisung Liu$^{2}$ \quad Je Hyeong Hong$^{1,2}$\setcounter{footnote}{1}\thanks{Corresponding author.}\\
$^{1}$Dept. Electronic Engineering, Hanyang University \quad
$^{2}$Dept. Artificial Intelligence, Hanyang University\\
}

\begin{document}
\maketitle
\begin{abstract}
Semantic correspondence is essential for handling diverse in-the-wild images lacking explicit correspondence annotations.
While recent 2D foundation models offer powerful features, adapting them for unsupervised learning via nearest-neighbor pseudo-labels has key limitations: it operates locally, ignoring structural relationships, and consequently its reliance on 2D appearance fails to resolve geometric ambiguities arising from symmetries or repetitive features. 
In this work, we address this by reformulating pseudo-label generation as a Fused Gromov-Wasserstein (FGW) problem, which jointly optimizes inter-feature similarity and intra-structural consistency. 
{Our framework, Shape-of-You (SoY),} leverages a 3D foundation model to define this intra-structure in the geometric space, resolving abovementioned ambiguity. 
However, since FGW is a computationally prohibitive quadratic problem, we approximate it through anchor-based linearization.
The resulting probabilistic transport plan provides a structurally consistent but noisy supervisory signal.
Thus, we introduce a soft-target loss dynamically blending guidance from this plan with network predictions to build a learning framework robust to this noise.
SoY achieves state-of-the-art performance on SPair-71k and AP-10k datasets, establishing a new benchmark in semantic correspondence without explicit geometric annotations. 
Code is available at \href{https://github.com/SpatialAILab/shapeofyou}{Shape-of-You}. 
\end{abstract}

\begin{figure}[t!]
\centering
\includegraphics[width=0.95\columnwidth]{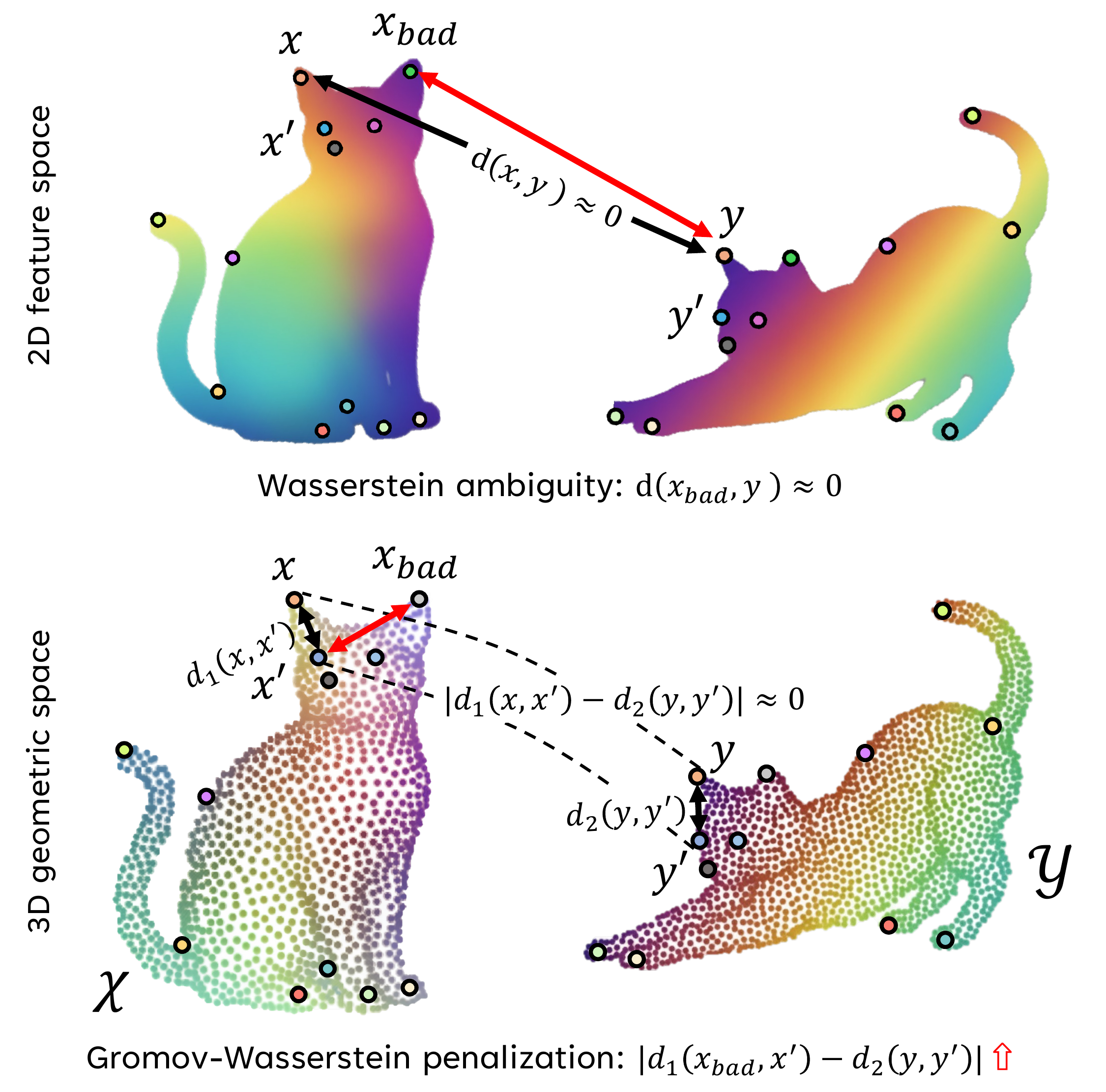}
\vspace{-2mm}
\caption{
Our Fused Gromov-Wasserstein approach combines inter-feature matching with intra-geometric consistency.
(Top) Feature matching yields false correspondences (red) when distinct points ($x, x_{bad}$) share similar features to $y$.
(Bottom) 3D Gromov-Wasserstein penalizes distortions to filter invalid matches.
}
\label{fig:background}
\vspace{-4mm}
\end{figure}

\vspace{-3mm}
\section{Introduction}
\label{sec:intro}
Semantic correspondence, the task of establishing meaningful pixel-level alignments between different instances within the same category, is fundamental to various computer vision applications such as object pose estimation~\cite{wang2024gs}, robotic manipulation~\cite{ju2024robo}, and visual content editing~\cite{zhu2025scene, chen2024zero}. 
However, finding semantic correspondences in ``in-the-wild" settings, where extreme variations in viewpoint, illumination, and intra-class shape exist, presents a significant challenge for current methods.

This challenge is compounded by practical constraints.
While fully and weakly supervised methods achieve strong performance, they inherently rely on explicit pixel-level annotations or auxiliary geometric metadata.
Since acquiring such data at scale can be challenging for real-world scenarios, learning semantic correspondence without explicit geometric annotations (e.g., camera poses, 3D models, viewpoints) has emerged as a practical alternative~\cite{gupta2023asic, tang2023emergent}.

This prevalence has been fueled by the recent rise of 2D vision foundation models such as DINO~\cite{caron2021emerging,oquab2024dinov} demonstrating effective zero-shot correspondence performance. 
To further improve this zero-shot capability, current methods lacking explicit metadata~\cite{gupta2023asic, shtedritski2024shic} leverage these powerful features by generating pseudo-labels via simple nearest-neighbor (NN) matching in feature space. 
However, this reliance on NN matching suffers from two fundamental issues.
First, NN matching operates locally in feature space ignoring global information necessary to verify the geometric consistency of matches across the entire image. 
Second, compounding this issue, these models are trained purely on 2D appearance cues and thus fail to reflect the true 3D geometric structure of objects. 
This produces correspondences that are semantically plausible but geometrically incorrect, introducing training noise that degrades performance.

We observe that these issues stem from two inter-connected factors: local matching and the 2D space limitation.
Local matching fails to preserve structural relationships within each image (intra-information).
Based on this insight, we reformulate the semantic correspondence problem as a Fused Gromov-Wasserstein (FGW) optimal transport problem, which jointly optimizes inter feature similarity and intra structural consistency for globally consistent matching.
However, applying FGW in 2D feature space still leaves the fundamental geometric ambiguity unresolved, as 2D models trained purely on appearance allow points close in feature space to be far apart in actual 3D space.
This motivates the need to utilize 3D geometric constraints, which can be effectively achieved by leveraging recent 3D foundation models.
By defining the intra-information required by FGW in geometric space rather than feature space (Fig.~\ref{fig:background}), the matching process can preserve 3D structure while resolving ambiguities arising from 2D appearance.

To this end, we propose \textbf{Shape-of-You (\oursAbbrv)}, a novel framework that incorporates geometric constraints into pseudo-label generation. 
Our approach lifts 2D images into 3D point cloud representations using a pretrained 3D foundation model (VGGT~\cite{wang2025vggt}) and formulates the correspondence problem as a Fused Gromov-Wasserstein (FGW) optimal transport problem that uses this 3D geometric structure as its intra-structure.
Through iterative refinement with anchor-based approximations, we generate geometrically-aware pseudo-labels that accommodate partial visibility and occlusions. 
We then train a lightweight adapter network using a soft-target loss derived from the probabilistic transport plan, enabling robust learning despite pseudo-label noise.

In summary, our contributions are as follows:
\begin{itemize}
 \item We \emph{formulate semantic correspondence as a Fused Gromov-Wasserstein optimal transport problem}, jointly optimizing inter feature similarity and intra geometric structure for globally consistent matching.
  
 \item We propose a \emph{geometry-aware unbalanced optimal transport method for pseudo-labeling} that enforces global 3D consistency while efficiently \emph{approximating Gromov-Wasserstein through anchor-based linearization} (Sec.~\ref{sec:pseudo}).
 
\item We introduce a \emph{soft-target loss} operating on probabilistic transport plans rather than hard labels to handle correspondence ambiguity and noise during training (Sec.~\ref{sec:train}).
\end{itemize}

Our method achieves state-of-the-art performance on the SPair-71k~\cite{min2019spair} and AP-10k benchmarks, with a PCK@0.10 of 67.9\% and {68.0\%}~(intra-species), respectively.
\vspace{-1mm}
\subsection{Related work}
\label{sec:related_work}
\vspace{-1mm}

\paragraph{Semantic correspondence.} 
Foundation models like DINO~\cite{oquab2024dinov} and Stable Diffusion~\cite{rombach2022high} enable semantic correspondence via dense features~\cite{tang2023emergent}.
Building on these, supervised approaches such as DHF~\cite{luo2023diffusion} and TLR~\cite{zhang2024telling} or weakly supervised methods like SphMap~\cite{mariotti2024improving} and DIY-SC~\cite{dunkel2025yourself} achieve strong performance but rely on explicit annotations or 3D geometric metadata.
To address this, methods without explicit geometric annotations (e.g., viewpoints or 3D models) leverage foundation model features to generate pseudo-labels through nearest-neighbor matching (e.g., ASIC~\cite{gupta2023asic}), reframe the problem as image-to-shape correspondence (e.g., SHIC~\cite{shtedritski2024shic}), or fine-tune features via distillation (e.g., DistillDIFT~\cite{fundel2025distillation}).
While dispensing with such explicit metadata, they often rely on implicit or weak supervision such as CLIP priors, curated category-specific sets (e.g., Neural Congealing~\cite{ofri2023neural}) or SAM masks~\cite{gupta2023asic}.
However, their reliance on local matching or teacher distillation can make it challenging to fully capture global object structure, potentially introducing geometric ambiguities.

\vspace{-5mm}
\paragraph{Optimal transport for structural matching.}
The failure of local matching highlights the need for methods that capture global structure. 
Early works formulated semantic correspondence as an optimal transport (OT) problem~\cite{liu2020semantic} extensively studied in 3D non-rigid shape matching~\cite{le2024integrating, eisenberger2020deep}. 
Optimal transport made practical by solvers like the Sinkhorn algorithm~\cite{cuturi2013sinkhorn} is a powerful assignment module. 
Its classical Wasserstein formulation uses direct costs like feature distances for correspondence learning~\cite{sarlin2020superglue, puy2020flot} and for video–text alignment~\cite{lin2024multi}.
More recently GECO~\cite{hartwig2025geco} integrated unbalanced OT for supervised geometrically consistent feature learning. 
For structure-based matching Gromov-Wasserstein (GW) extends OT to consider internal structures of matched entities making it ideal for tasks like graph matching~\cite{xu2019gromov}, video action segmentation~\cite{xu2024temporally} and point cloud alignment~\cite{ryner2023globally}.
Yet, GW is a non-convex quadratic problem hindering guaranteed convergence and prompting various approximations~\cite{li2022fast, xu2019gromov, memoli2011gromov, sato2020fast}.

\vspace{-5mm}
\paragraph{3D geometric foundation models.}
Computing GW's structural costs requires extracting 3D geometry from images.
Recent approaches use feed-forward transformers for direct 3D prediction, replacing multi-stage pipelines.
DUSt3R~\cite{dust3r_cvpr24} predicts camera poses and 3D point maps, while MASt3R~\cite{mast3r_eccv24} demonstrates robust same-scene matching. 
We employ VGGT~\cite{wang2025vggt}, predicting multiple 3D attributes in a single forward pass without post-processing.

\vspace{-5mm}
\paragraph{Remark.}
Methods capturing global geometric structure beyond local matching are needed.
We bridge this gap by combining 3D lifting with \textit{Gromov-Wasserstein optimal transport}, approximated via anchor-based linearization.

\section{Preliminaries}
\vspace{-1mm}
\label{sec:preliminaries}
This section reviews the optimal transport concepts foundational to our method, from classical optimal transport to unbalanced formulations for partial matching, and the Gromov-Wasserstein extension for structural comparison.
\vspace{-1mm}
\subsection{Classical optimal transport}
\vspace{-1mm}
The classical optimal transport problem, also known as the Kantorovich problem, seeks to find the most efficient plan to transport mass between two discrete distributions.
Given source and target sets with $N$ and $M$ points respectively, each point is assigned a \textit{mass} representing its relative importance, expressed as discrete probability distributions $a \in \Delta_N$ and $b \in \Delta_M$. 
Here, $\Delta_N \coloneqq \{ x \in \mathbb{R}_+^N : \sum_i x_i = 1 \}$ is the probability simplex for the $N$ source points, and similarly $\Delta_M$ is the simplex for the $M$ target points.
The ground cost matrix $C \in \mathbb{R}_+^{N \times M}$ defines the cost $C_{ij}$ of moving mass from the $i$-th source point to the $j$-th target point (e.g., cosine distance between features).
A transport plan $\pi \in \mathbb{R}_+^{N \times M}$ specifies the mass $\pi_{ij}$ moved from source $i$ to target $j$. 
For balanced transport, $\pi$ must satisfy the marginal constraints, ensuring all mass from $a$ is transferred to $b$:
\begin{equation}
\Pi(a, b) = \{ \pi \in \mathbb{R}_+^{N \times M} \mid \pi \mathbf{1}_M = a, \; \pi^\top \mathbf{1}_N = b \}
\eqlabel{pi_def}
\end{equation}
where $\mathbf{1}$ is a vector of ones. 
The optimal transport problem finds the plan $\pi$ that minimizes the total transportation cost:
{
\small
\begin{equation}
\min_{\pi \in \Pi(a,b)} \langle C, \pi \rangle, \quad \text{where} \quad \langle C, \pi \rangle = \sum_{i,j} C_{ij} \pi_{ij}
\eqlabel{ot}
\end{equation}
}
This linear program is made differentiable and efficiently solved using the Sinkhorn algorithm~\cite{cuturi2013sinkhorn} by adding an entropic regularizer $-\varepsilon H(\pi)$. 

\begin{figure*}[t!]
\centering
\includegraphics[width=\textwidth]{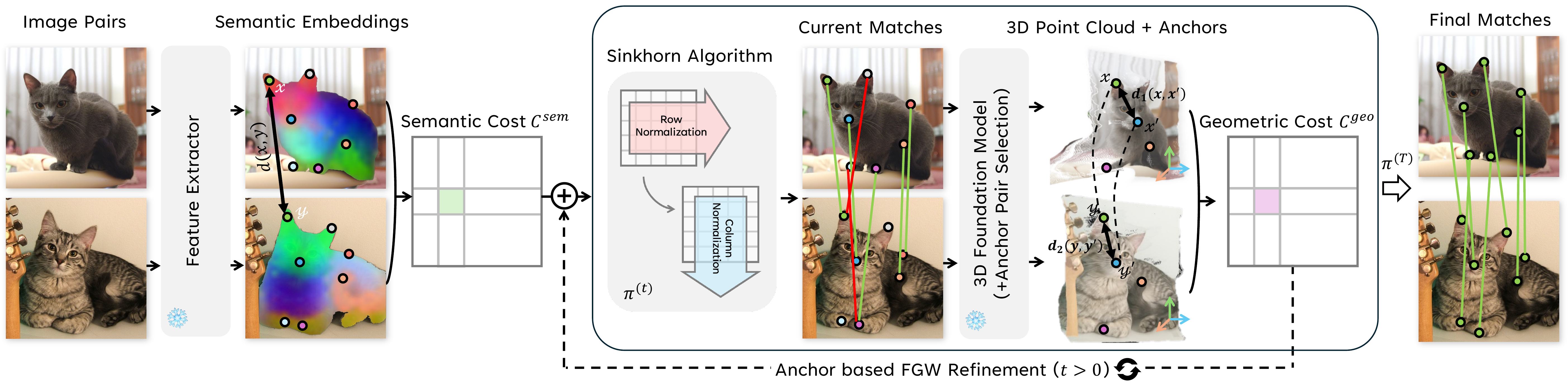}
\vspace{-7mm}
\caption{Overview of our pseudo-label generation pipeline. We first compute an initial semantic match to identify high-confidence anchors. These anchors are then used to create a tractable, linear approximation of the otherwise intractable quadratic Gromov-Wasserstein (GW) geometric cost. This approximated geometric cost is fused with the semantic cost, yielding a final {fused cost matrix}. This cost matrix is then used to solve an UOT problem, producing a transport plan $\pi^{(T)}$ that serves as our pseudo-label robust to geometric ambiguities.}

\label{fig:main}
\vspace{-4mm}
\end{figure*}
\vspace{-1mm}
\subsection{Unbalanced optimal transport}
The balanced constraint $\pi \in \Pi(a,b)$ (Eq.~\eqref{pi_def}), which assumes all mass must be transported, is often too strict for partial matching tasks (e.g., matching only overlapping regions between viewpoints). 
Instead of searching within $\Pi(a,b)$, unbalanced optimal transport (UOT)~\cite{chizat2018scaling} relaxes the marginal constraints by optimizing over non-negative matrices $\pi \in \mathbb{R}_+^{N \times M}$ with penalty terms for deviating from marginals $a$ and $b$. 
A common penalty is the Kullback-Leibler (KL) divergence:
{\small
\begin{equation}
\begin{aligned}
\min_{\pi \in \mathbb{R}_+^{N \times M}} \; \langle C, \pi \rangle
 + \rho D_{KL}(\pi \mathbf{1}_M \| a)
 + \rho D_{KL}(\pi^\top \mathbf{1}_N \| b)
\end{aligned}
\eqlabel{uot}
\end{equation}
}
where $D_{KL}(p \| q)$ is the Kullback-Leibler divergence and $\rho > 0$ is a regularization parameter that penalizes deviation of the marginals ($\pi \mathbf{1}_M, \pi^\top \mathbf{1}_N$) from the original distributions ($a, b$). 
The parameter $\rho$ controls the trade-off between cost minimization and marginal enforcement: as $\rho \to \infty$, the problem reverts to balanced OT, while finite $\rho$ allows mass exclusion for partial matching.

\subsection{Gromov-Wasserstein (GW) optimal transport}
\label{sec:gwot}
While classical optimal transport successfully computes a cost $C$ between points in a shared, aligned space, this direct comparison fails for domains that are not naturally aligned. 
For instance, a direct distance $C_{ij}$ is meaningless for geometric spaces where two object instances exist in arbitrary, unaligned coordinate frames.
The Gromov-Wasserstein (GW) optimal transport~\cite{peyre2016gromov} extends the classical formulation to this setting by comparing the \textit{internal structure} of each space, rather than the points themselves.

Instead of the cross-domain cost $C$, the GW problem is formulated using two \textit{intra-domain} distance matrices: $D_1 \in \mathbb{R}_+^{N \times N}$ for the source and $D_2 \in \mathbb{R}_+^{M \times M}$ for the target, whose entries $D_{1,ii'}$ and $D_{2,jj'}$ represent distances between point pairs $(i,i')$ and $(j,j')$ within their respective spaces.
The GW problem then finds a transport plan $\pi \in \Pi(a,b)$ that best preserves these internal structures, formulated as a quadratic optimization problem:
\begin{equation}
\eqlabel{gw_discrete}
\min_{\pi \in \Pi(a,b)} \sum_{i,j} \sum_{i',j'} |D_{1,ii'} - D_{2,jj'}| \pi_{ij} \pi_{i'j'}
\end{equation}
This objective penalizes pairings that break the internal geometry. 
When $i\!\to\!j$ and $i'\!\to\!j'$ are matched (high $\pi_{ij}, \pi_{i'j'}$), the objective ensures that the intra-source distance $D_{1,ii'}$ aligns with the intra-target distance $D_{2,jj'}$.

However, the above GW formulation presents a significant practical challenge. Eq.~\eqref{gw_discrete} involves a quadratic term in $\pi$ (i.e., $\pi_{ij}\pi_{i'j'}$), resulting in a non-convex optimization problem that prevents use of standard linear OT solvers such as the Sinkhorn algorithm. Finding a tractable approximation for this objective is a key contribution of our method.

\section{Pseudo label generation via FGW}
\vspace{-1mm}
\label{sec:pseudo}
Our goal is to leverage both semantic feature similarity and geometric structure to generate robust pseudo-labels for semantic correspondence.
While feature similarity, residing in a shared embedding space, can be directly compared using a classical OT cost $\mathcal{F}_{\text{OT}}(\pi)$ = $\langle C, \pi \rangle$, geometric structures exist in arbitrary, unaligned coordinate frames, making a direct cost comparison infeasible.
We therefore address this using the Gromov-Wasserstein (GW) cost, $\mathcal{F}_{\text{GW}}(\pi)$ (Eq.~\eqref{gw_discrete}), which preserves intra-domain structural relationships.
To jointly optimize both, we formulate the task as a Fused Gromov-Wasserstein (FGW) problem~\cite{titouan2019optimal}, defined as $\min_{\pi} (1-\alpha) \mathcal{F}_{\text{OT}}(\pi) + \alpha \mathcal{F}_{\text{GW}}(\pi)$, where the hyperparameter $\alpha=0.3$ is used to balance the two terms.

However, optimizing this full FGW objective is a NP-hard problem---
As noted in the original work~\cite{titouan2019optimal}, the problem is provably a non-convex quadratic program. 
This non-convexity introduces significant challenges beyond the complexity mentioned in Sec.~\ref{sec:gwot}, as standard solvers are not only computationally expensive but also only guaranteed to converge to a local stationary point.

Therefore, we propose a two-stage process (Fig.~\ref{fig:main}): first, we solve a classic optimal transport problem using only semantic information to obtain initial anchors, then progressively integrate geometric costs based on these anchors.
\vspace{-1mm}
\subsection{Problem formulation}
\vspace{-1mm}
Given a source image $I^A$ and a target image $I^B$, we decompose them into uniform grid patches and select patches corresponding to \textit{instance regions}, which are segmented using SAM~\cite{kirillov2023segment}, obtaining $N$ and $M$ patch sets $\mathcal{X} = \{x_i\}_{i=1}^N$ and $\mathcal{Y} = \{y_j\}_{j=1}^M$ respectively.
Each patch is represented by two complementary representations: semantic features $\mathbf{f}_i^A \in \mathbb{R}^d$ extracted from a foundation model and 3D coordinates $\mathbf{v}_i^A \in \mathbb{R}^3$ obtained by lifting the image into 3D space using a pretrained 3D foundation model~\cite{wang2025vggt} followed by bilinear interpolation. 
These are aggregated into feature matrices $\mathbf{F}^A \in \mathbb{R}^{N \times d}$, $\mathbf{F}^B \in \mathbb{R}^{M \times d}$ and coordinate matrices $\mathbf{V}^A \in \mathbb{R}^{N \times 3}$, $\mathbf{V}^B \in \mathbb{R}^{M \times 3}$ respectively.
To establish correspondence, we define uniform probability distributions over the two patch sets $\mathcal{X}$ and $\mathcal{Y}$ as $a = \frac{1}{N}\mathbf{1}_N$ and $b = \frac{1}{M}\mathbf{1}_M$ formulating the problem as optimal transport.

\vspace{-1mm}
\subsection{Semantic UOT for anchor initialization (\texorpdfstring{$t=0$}{t=0})}
\label{sec:uot}
\vspace{-1mm}

This initial stage aims to solve the classical optimal transport component of the FGW problem, relying purely on semantic feature similarity.
To approximate the GW term, we perform anchor selection in the next stage (Sec.~\ref{sec:anchor}) where anchors are defined from the transport plan $\pi$, thus requiring its initialization via semantic UOT.
We compute a semantic cost matrix $C^{\text{sem}} \in \mathbb{R}^{N \times M}$, where $C^{\text{sem}}_{ij} = 1 - S_{ij}$ and $S_{ij}$ is the cosine similarity between patch features $\mathbf{f}_i^A$ and $\mathbf{f}_j^B$.

A key challenge in semantic correspondence is that not all patches may be matched due to occlusions and/or non-overlapping regions. 
Hence, the classical OT formulation (Eq.~\eqref{ot}) fails in this scenario, so we adopt \textit{unbalanced optimal transport (UOT)} as formulated in Eq.~\eqref{uot} relaxing the marginal constraints.
We obtain the initial transport plan $\pi^{(0)}$ by solving this problem using our semantic cost $C^{\text{sem}}$.

\vspace{-2mm}
\subsection{Anchor-based FGW refinement (\texorpdfstring{$t>0$}{t>0})}
\label{sec:anchor}
This stage refines the initial plan by solving the full FGW problem. 
To overcome the intractability of the quadratic GW term, we linearize it using the anchors from stage 1 and then iteratively fuse it with the semantic cost.
\vspace{-4mm}
\paragraph{Anchor pair selection.}
We define the internal structure of each instance by computing the intra-instance pairwise distance matrices $D^{A}\!\in\!\mathbb{R}^{N \times N}$ and $D^{B}\!\in\!\mathbb{R}^{M \times M}$ using 3D coordinates, where $D^{A}_{ik}\!=\!\| \mathbf{v}_i^{A} - \mathbf{v}_k^{A} \|_2$.
In each iteration $t$, we select $K{=}64$ high-confidence anchor pairs $\mathcal{A}^{(t)} = \{(i_k, j_k)\}_{k=1}^K$ from the transport plan $\pi^{(t-1)}$ to serve as geometric reference points for linearizing the GW cost.
To ensure these anchors are stable and mutually consistent, we enforce cycle-consistency in both matching space and 3D geometry. 
Specifically, for each source patch, we identify its best forward match in the target, then trace backward from this target to find its best match in the source.
We compute the 3D cycle error between the original source patch and this backward-matched patch, retaining only pairs where this error falls below a threshold $\delta$.
From these cycle-consistent candidates, we select the top $K$ pairs with the highest matching confidence as our final anchors.

\vspace{-4mm}
\paragraph{Anchor-based GWD linearization.}
With these structures ($D^A, D^B$) and the anchor set $\mathcal{A}^{(t)}$ defined, we can now address the geometric cost. 
The original objective for geometric consistency is the Gromov-Wasserstein distance (GWD) as defined in Eq.~\eqref{gw_discrete}:
\begin{equation}
\mathcal{L}_{\mathrm{GW}}(\pi) = \sum_{i,j} \sum_{i',j'} \big| D^{A}_{ii'} - D^{B}_{jj'} \big| \pi_{ij} \pi_{i'j'}
\eqlabel{gw_ideal}
\end{equation}
To create a tractable objective, we linearize the problem by approximating one of the quadratic terms (indexed by $i', j'$) with a fixed anchor transport plan $\hat{\pi}$, following prior graph alignment~\cite{tang2023fused}.
This anchor-based linearization is consistent with the GW view of comparing intra-instance distance patterns~\cite{memoli2011gromov} and with anchor-based GW surrogates~\cite{sato2020fast}, which replace the quadratic distortion with convex Wasserstein or energy costs to obtain efficient and robust approximations to full GW.
Specifically, this anchor transport plan $\hat{\pi}$ is a sparse probability distribution that assigns uniform mass $\hat{\pi}_{i'j'} = 1/K$ to each of the $K$ anchor pairs $(i',j') \in \mathcal{A}^{(t)}$ and zero otherwise, treating all selected anchors as equally reliable.
By substituting $\hat{\pi}$ for the $\pi_{i'j'}$ term in Eq.~\eqref{gw_ideal}, we transform the quadratic objective into a linear one with respect to $\pi$:
{\small
\begin{align}
\mathcal{L}_{\mathrm{GW}}(\pi) 
&\approx \sum_{i,j} \pi_{ij} \left( \frac{1}{K}\sum_{(a_A,a_B) \in \mathcal{A}^{(t)}} \big| D^{A}_{i,a_A} - D^{B}_{j,a_B} \big| \right) 
\eqlabel{geo-cost} 
\end{align}
}
This yields a linear objective of the form $\sum_{i,j} \pi_{ij} C^{\mathrm{geo}}_{ij}$, where $C^{\mathrm{geo}}_{ij}$ is the parenthesized term in Eq.~\eqref{geo-cost}. 
Intuitively, $C^{\mathrm{geo}}_{ij}$ computes the average geometric distortion of candidate match $(i,j)$ measured against all $K$ anchors, quantifying whether the 3D distance from patch $i$ to each source anchor $a_A$ is preserved by candidate match $j$ relative to target anchor $a_B$.
Under the GW perspective that encodes structure through intra-instance distances, preserving distances to sufficient anchors implies approximate preservation of the full geometry, so if a candidate match agrees with all reliable anchors, it is likely to preserve the overall structure.

\begin{figure}[t!]
\centering
\includegraphics[width=0.9\columnwidth]{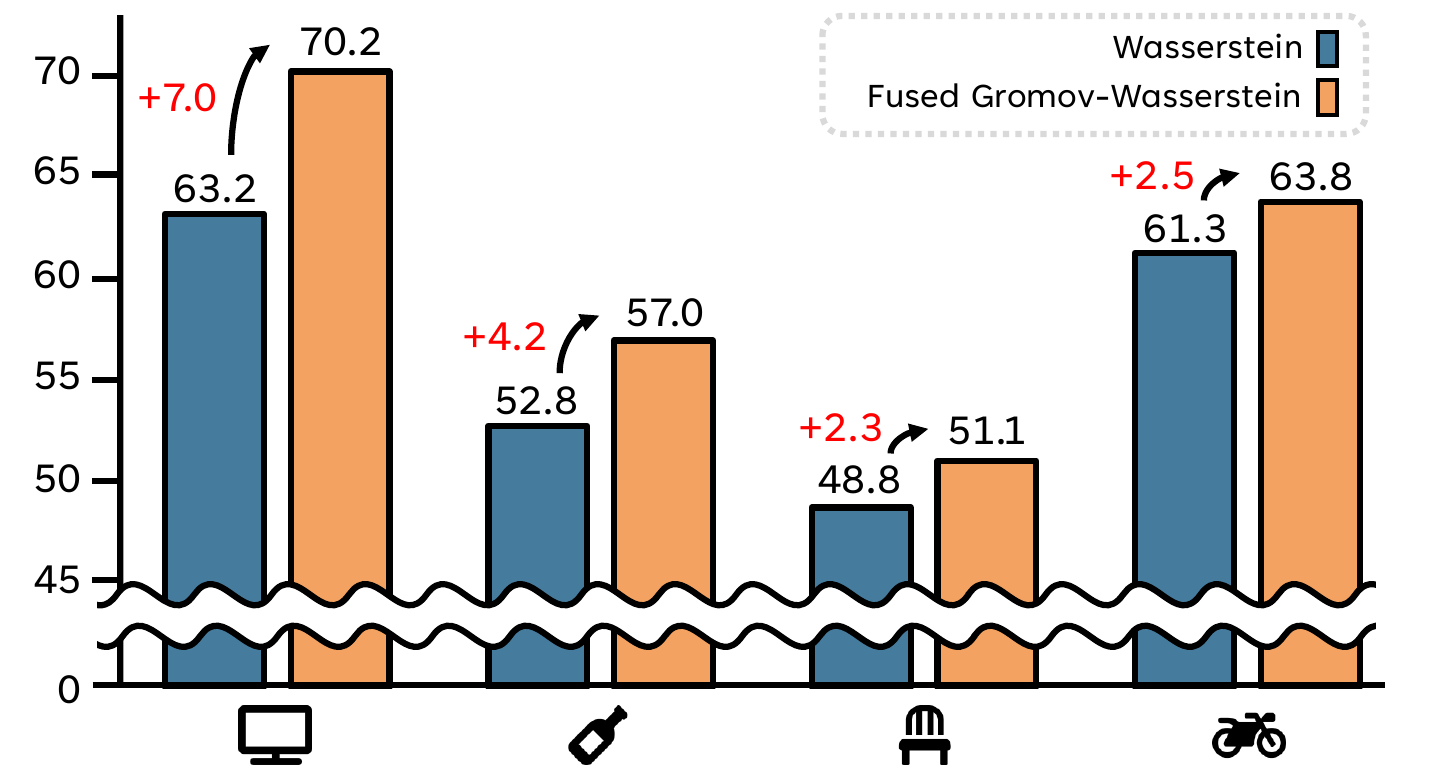}
\vspace{-3mm}
\caption{Impact of the Gromov-Wasserstein term. Wasserstein baseline (blue) vs. Fused Gromov-Wasserstein (orange). Incorporating geometric structural consistency leads to consistent improvements (+2.3 $\sim$ +7.0\%p) across categories.}
\label{fig:w_fgw}
\vspace{-4mm}
\end{figure}

\vspace{-3mm}
\paragraph{Fused cost update and iteration.}
We now fuse the semantic cost $C^{\text{sem}}$ and the linearized geometric cost $C^{\mathrm{geo}}$. 
To ensure both costs contribute on a comparable scale, we normalize them into ($\tilde{C}^{\text{sem}}, \tilde{C}^{\text{geo}}$) and define the total cost:
\begin{equation} 
C^{\text{total}} = (1-\alpha) \, \tilde{C}^{\text{sem}} + \alpha\, \tilde{C}^{\text{geo}}
\eqlabel{fused-cost}
\end{equation}
We resolve the UOT problem (Eq.~\eqref{uot}) using this geometry-aware cost $C^{\text{total}}$ to find the refined transport plan $\pi^{(t)}$:
{\small
\begin{equation}
\eqlabel{update-step}
\mathop{\arg\min}_{\pi \in \mathbb{R}_+^{N \times M}} \;
\langle C^{\text{total}}, \pi \rangle
+ \rho \mathcal{D}_{\mathrm{KL}}(\pi \mathbf{1}_M \| a)
+ \rho \mathcal{D}_{\mathrm{KL}}(\pi^\top \mathbf{1}_N \| b)
\end{equation}
}
We iterate this refinement for $T$ iterations. 
At the first iteration ($t=1$), the anchor set $\mathcal{A}^{(1)}$ is extracted from the initial semantic plan $\pi^{(0)}$. For $t>1$, anchors $\mathcal{A}^{(t)}$ are iteratively updated from $\pi^{(t-1)}$.
In each iteration, this new anchor set helps compute $C^{\mathrm{geo}}$, which is then fused (Eq.~\eqref{fused-cost}) to update the transport plan $\pi^{(t)}$ (Eq.~\eqref{update-step}).
We observe this alternation between anchor selection and transport optimization gradually converges to correspondences that satisfy both appearance similarity and geometric consistency. 
The final transport matrix $\pi^{(T)}$ serves as the basis for our pseudo-label.

As shown in Fig.~\ref{fig:w_fgw}, incorporating the GW term consistently improves the pseudo-labelling accuracy, implicitly demonstrating the importance of enforcing geometric structural consistency in semantic correspondences.

\section{Training pipeline}
\label{sec:train}
We now use the pseudo-labels $\pi^{(T)}$ generated through iterative geometric refinement in Sec.~\ref{sec:pseudo} to train a light refinement network $f_\theta$ (same architecture as~\cite{zhang2024telling}).
The network takes multi-scale features from pretrained DINOv2~\cite{oquab2024dinov} and Stable Diffusion~\cite{rombach2022high} models as input, which are projected and fused through learnable weights to produce a refined feature map.
Once trained, the network enables efficient correspondence estimation at inference time without requiring the iterative optimization used for label generation.

To further robustify the supervision signal before training, we apply a relaxed cycle consistency constraint~\cite{dunkel2025yourself} to filter the generated pseudo-labels. 
Extending their approach, we construct a richer candidate set by retaining the top-$k$ matches and enforcing a symmetric relaxed cycle consistency, rather than relying on a single match.

\subsection{Training objectives}
Our training objective combines two losses: our proposed \emph{soft target loss} ($\mathcal{L}_{\text{soft}}$) and the dense correspondence loss ($\mathcal{L}_{\text{dense}}$).
The final objective is $\mathcal{L}_{\text{total}} = \mathcal{L}_{\text{soft}} + \mathcal{L}_{\text{dense}}$.

Our generated pseudo-label $\pi^{(T)}$ is a probabilistic matrix that is structurally consistent but contains noise and ambiguity. 
Using this noisy distribution directly as target would force the network to learn imperfections. 
We therefore distill $\pi^{(T)}$ into a multi-hot binary target $\pi^{\mathrm{hard}}$ representing the filtered top-$k$ candidates. 
However, $\pi^{\mathrm{hard}}$ remains overconfident, creating a challenge our soft target loss addresses.

\begin{table*}[t!]
 \centering
 \renewcommand{\arraystretch}{1.1}
 \resizebox{\textwidth}{!}{
 \begin{tabular}{lc|RRRRRRRRRRRRRRRRRR|R}
{Method} & & { \faIcon{plane}} & { \faIcon{bicycle}} & { \faIcon{crow}} & { \faIcon{ship}} & { \faIcon{wine-bottle}} & { \faIcon{bus}} & { \faIcon{car}} & { \faIcon{cat}} & { \faIcon{chair}} & { \Cow} & { \faIcon{dog}} & { \faIcon{horse}} & { \faIcon{motorcycle}} & {\faIcon{walking}} & { \Plant} & { \Sheep} & { \faIcon{train}} & { \faIcon{tv}} & {avg~($\uparrow$)} \\
  \midrule
  \myrowcolour
  {ASIC} & \cite{gupta2023asic} & 57.9 & 25.2 & 68.1 & 24.7 & 35.4 & 28.4 & 30.9 & 54.8 & 21.6 & 45.0 & 47.2 & 39.9 & 26.2 & 48.8 & 14.5 & 24.5 & 49.0 & 24.6 & 36.9 \\
  {DINOv2} & \cite{oquab2024dinov} & 72.7 & 62.4 & 85.2 & \best{41.4} & 40.3 & 52.5 & \second{51.5} & 71.3 & 36.1 & 67.2 & 65.0 & 67.6 & 61.1 & \second{68.5} & 30.6 & 61.9 & 54.3 & 24.3 & 55.7 \\
  \myrowcolour
  {DIFT} & \cite{tang2023emergent} & 63.5 & 54.5 & 80.8 & 34.5 & 46.2 & 52.7 & 48.3 & 77.7 & 39.0 & 76.0 & 54.9 & 61.3 & 53.3 & 46.0 & \second{57.8} & 57.1 & \best{71.1} & \second{63.4} & 57.7 \\
  {DistillDIFT}$^\dagger$ & \cite{fundel2025distillation} & 70.3 & 55.4 & 85.9 & 36.8 & 51.9 & 51.7 & 50.5 & 78.0 & 40.5 & 73.8 & \second{65.3} & 65.2 & 54.5 & 66.0 & 47.8 & 59.4 & 55.6 & 54.4 & 59.8 \\
  \myrowcolour
  {DINOv2 + SD}$^\dagger$ & \cite{zhang2023tale} & \second{72.9} & \second{63.4} & \second{86.4} & 40.5 & \second{52.6} & \best{55.4} & \best{53.3} & \second{78.4} & \second{45.2} & \second{77.1} & 64.7 & \second{69.4} & \second{62.9} & \second{68.5} & 56.8 & \second{67.0} & 65.9 & 51.8 & \second{63.5} \\
  {\oursTab} & & \best{73.5} & \best{66.7} & \best{89.9} & \second{40.8} & \best{58.0} & \best{55.4} & 51.1 & \best{84.8} & \best{52.1} & \best{81.0} & \best{71.0} & \best{75.3} & \best{64.1} & \best{71.8} & \best{62.1} & \best{70.2} & \second{70.2} & \best{64.4} & \best{67.9} \\
  \bottomrule
 \end{tabular}
 }
 \vspace{-2mm}
    \caption{\textbf{Per-category PCK@0.1 scores (per-keypoint) on SPair-71k} (\textit{higher is better}~$\uparrow$). \best{Best} and \second{second best} are highlighted. $\dagger$ denotes results re-evaluated under identical standard evaluation settings from~\cite{min2019spair}.}
 \label{tab:spair_results}
 \vspace{-4mm}
\end{table*}

\vspace{-4mm}
\paragraph{Soft target loss.}
A key component of our training is the \textit{soft target loss}, designed to handle the overconfidence of the multi-hot binary mask $\pi^{\mathrm{hard}}$ derived from the top-$k$ candidates. 
A naive loss using only $\pi^{\mathrm{hard}}$ would over-penalize unselected but semantically similar candidates as hard negatives, forcing an artificial separation that can disrupt the learning of a coherent semantic feature space.
To mitigate this, we introduce a \textit{dynamic} label smoothing strategy, where the smoothing target is created based on the network's current semantic understanding.
We achieve this by adapting a strategy from the video-text alignment domain~\cite{lin2024multi}, blending our geometric guide $\pi^{\mathrm{hard}}$ with a `soft' semantic plan $\pi^{\mathrm{curr}}$ derived from the network's features.

We compute this `current' soft target $\pi^{\text{curr}}$ as the optimal plan minimizing $\langle C^{\text{curr}}, \pi \rangle$ found by solving an optimal transport problem using the semantic cost $C^{\text{curr}}$ derived from the network's current features. 
Crucially, we stop the gradient from flowing back into the network during this plan computation, treating $\pi^{\text{curr}}$ as a fixed target.
Our final soft target $\pi^{\text{soft}}$ is a blend of these two:
\begin{equation}
\pi^{\text{soft}} = (1-\beta) \pi^{\text{hard}} + \beta \pi^{\text{curr}}
\eqlabel{soft_target}
\end{equation}
where $\beta=0.5$ is a mixing hyperparameter. 
This acts as dynamic label smoothing by guiding the network with the structurally consistent $\pi^{\text{hard}}$ while using $\pi^{\text{curr}}$ (the network's own semantic judgment) to soften penalties on semantically similar matches, thus preventing over-penalization.

We then train the network by minimizing a symmetric soft supervised contrastive loss between the predicted similarity distribution $\mathbf{S}$ and this robust soft target $\pi^{\text{soft}}$: 

\vspace{-2mm}
\begin{equation}
\mathcal{L}_{\text{soft}} =
\frac{1}{2}\left[
\smash{
\underbrace{\text{CE}(\tau \mathbf{S}, \pi^{\text{soft}})}_{\text{\scriptsize source$\rightarrow$target}} +
\underbrace{\text{CE}(\tau \mathbf{S}^\top, (\pi^{\text{soft}})^\top)}_{\text{\scriptsize target$\rightarrow$source}}
}
\right]
\eqlabel{soft_target_loss}
\vspace{2mm}
\end{equation}
where $\text{CE}(\tau \mathbf{S}, \pi^{\text{soft}})$:=$-\sum_{i}\frac{1}{Z_i}\sum_{j} \pi^{\text{soft}}_{ij} \log p_{ij}$, with $Z_i = \sum_j \pi^{\text{soft}}_{ij}$ acting as the row-wise normalization factor, $p_{ij}$=
$\exp(\tau S_{ij})$$/$$\sum_{k} \exp(\tau S_{ik})$,
and $\tau$ is a learnable parameter.

\vspace{-4mm}
\paragraph{Dense correspondence loss.}
To leverage the full spatial structure, we also adopt a standard dense correspondence loss, similar to~\cite{zhang2024telling}. 
This loss propagates gradients to all feature map locations, not just those with sparse pseudo-labels. 
For each source patch $p_i^s$, we compute a predicted target location $\hat{p}_i^t$ by applying a differentiable soft-argmax operator over the similarity map $\mathbf{S}$: 
\begin{equation}
\begin{split}
    \mathcal{L}_{\text{dense}} = \sum_{i \in \mathcal{P}} \bigl\|\hat{p}_i^t - \bigl(p_i^t + \epsilon\bigr)\bigr\|_2 \quad 
\end{split}
\eqlabel{dense_loss}
\end{equation}
where $\mathcal{P}$ is the set of pseudo-label pairs $(p_i^s, p_i^t)$ and $\epsilon$ is Gaussian noise, which serves as a regularization to prevent overfitting to the exact pseudo-label locations.
\subsection{Inference}
\label{sec:inference}
At test time, we use the trained network $f_\theta$, discarding the iterative optimization pipeline. 
Correspondence is established via nearest-neighbor matching on the cosine similarity, followed by soft-argmax to achieve sub-pixel accuracy.

\section{Experiments}
\label{sec:experiments}

\subsection{Experimental setup}
\paragraph{Datasets.}
SPair-71k~\cite{min2019spair} is a challenging benchmark consisting of 70,958 image pairs with significant variations in viewpoint, scale, occlusion and truncation; we evaluate on its test set of 12,234 pairs across 18 categories.
To specifically evaluate performance on cases with high geometric ambiguity, we adopt the \textit{Geometry-aware subset} protocol from~\cite{zhang2024telling}, which isolates keypoints that are semantically similar but geometrically distinct within the same part-group.
Evaluations on this subset measure robustness to geometric ambiguity, while the full test set measures overall performance.
We also use AP-10K~\cite{yu2021ap}, an animal pose estimation dataset recently repurposed by~\cite{zhang2024telling} as a benchmark for geometry-aware semantic correspondence.

\begin{table}
  \centering
  \resizebox{\columnwidth}{!}{
  \begin{tabular}{lcllcccc}
    \toprule
    &&   \multicolumn{3}{c}{SPair-71k}& \multicolumn{3}{c}{AP-10k (PCK@0.1)}\\
        \cmidrule(lrr){3-5}\cmidrule(llr){6-8}
        Models &&  0.1 & 0.05 & 0.01 &  I.S & C.S. & C.F.\\
    \midrule
    \myrowcolour
    DistillDIFT$^\dagger$ & \citep{fundel2025distillation} & 59.8 & 42.7 & 5.7 &  65.5 & 62.8 & 52.8\\
    DINOv2 + SD$^\dagger$ & \citep{zhang2023tale} & 63.5 & 48.3 & 8.8 & 65.5 & 63.3 & 51.1 \\
    \myrowcolour
    \oursTab\  &&  \textbf{67.9}  & \textbf{50.8} &\textbf{10.0} & \textbf{68.0} & \textbf{65.8} & \textbf{52.9} \\
    \bottomrule
  \end{tabular}
  }
  \vspace{-2mm}
    \caption{\textbf{Results for different PCK levels (per-keypoint) on SPair-71k and AP-10k.} Results for AP-10K are intra-species (I.S.), cross-species (C.S.) and cross-family (C.F.), following \cite{zhang2024telling}.}
  \label{tab:per-image-pck}
  \vspace{-5mm}
\end{table}

\vspace{-3mm}

\begin{figure*}[t]
    \centering
    \begin{subfigure}[t]{0.495\textwidth}
        \centering
        \scriptsize
        \makebox[0.24\linewidth]{Source}\hfill
        \makebox[0.24\linewidth]{DistillDIFT}\hfill
        \makebox[0.24\linewidth]{DINOv2 + SD}\hfill
        \makebox[0.24\linewidth]{Ours}\\[2pt]
        \includegraphics[width=0.24\linewidth]{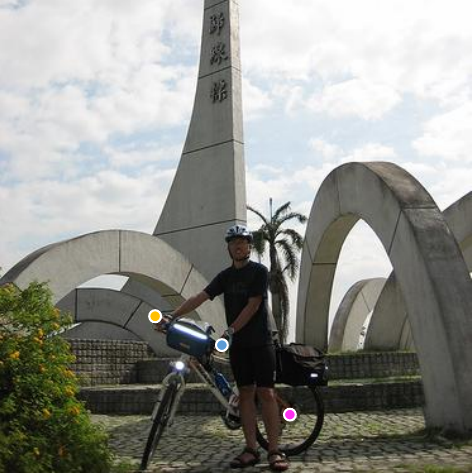}
        \includegraphics[width=0.24\linewidth]{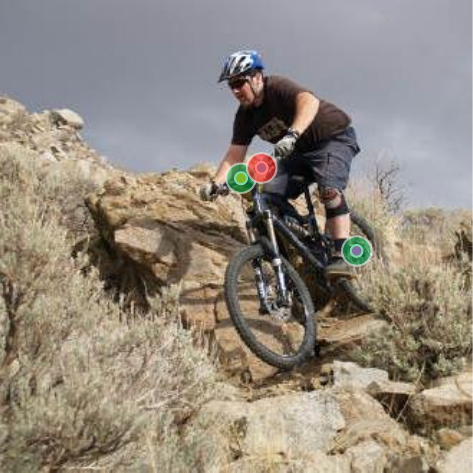}
        \includegraphics[width=0.24\linewidth]{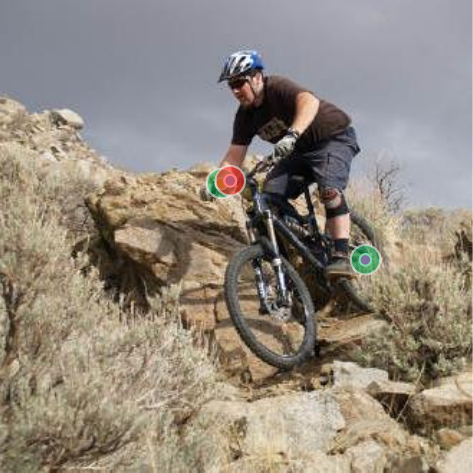}
        \includegraphics[width=0.24\linewidth]{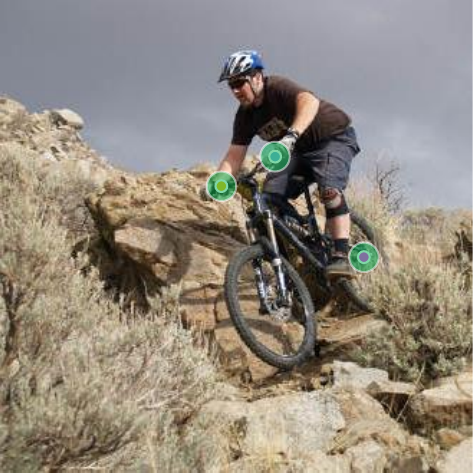}
        \caption{Viewpoint change, severe occlusion.}
        \label{fig:failure_a}
    \end{subfigure}
    \hfill
    \begin{subfigure}[t]{0.495\textwidth}
        \centering
        \scriptsize
        \makebox[0.24\linewidth]{Source}\hfill
        \makebox[0.24\linewidth]{DistillDIFT}\hfill
        \makebox[0.24\linewidth]{DINOv2 + SD}\hfill
        \makebox[0.24\linewidth]{Ours}\\[2pt]
        \includegraphics[width=0.24\linewidth]{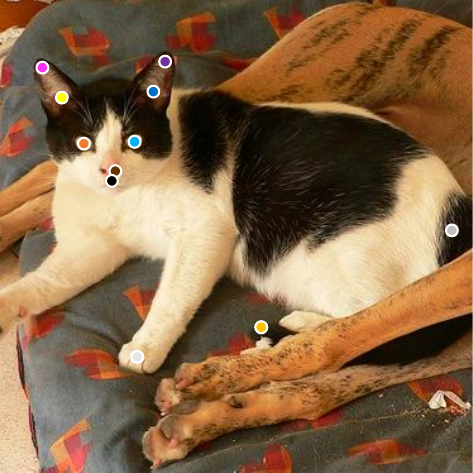}
        \includegraphics[width=0.24\linewidth]{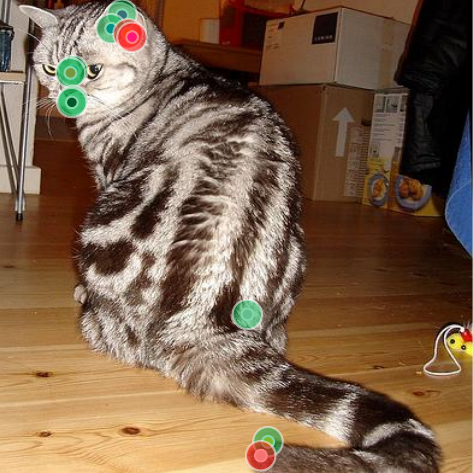}
        \includegraphics[width=0.24\linewidth]{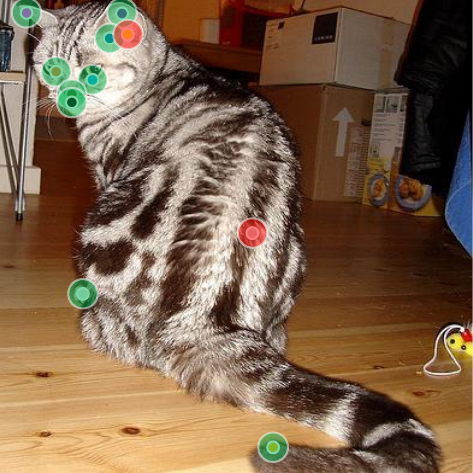}
        \includegraphics[width=0.24\linewidth]{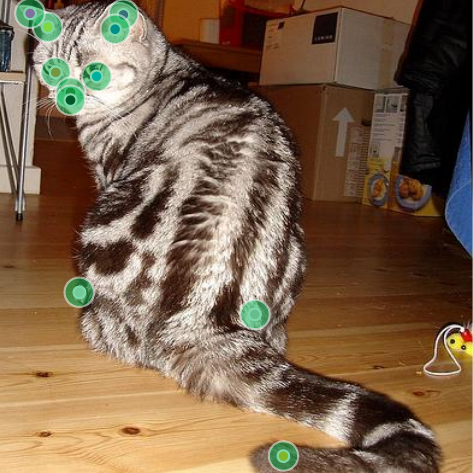}
        \caption{Extreme pose variation, self-occlusion.}
        \label{fig:failure_b}
    \end{subfigure}
    
    \vspace{0.3cm}
    
    \begin{subfigure}[t]{0.495\textwidth}
        \centering
        \scriptsize
        \makebox[0.24\linewidth]{Source}\hfill
        \makebox[0.24\linewidth]{DistillDIFT}\hfill
        \makebox[0.24\linewidth]{DINOv2 + SD}\hfill
        \makebox[0.24\linewidth]{Ours}\\[2pt]
        \includegraphics[width=0.24\linewidth]{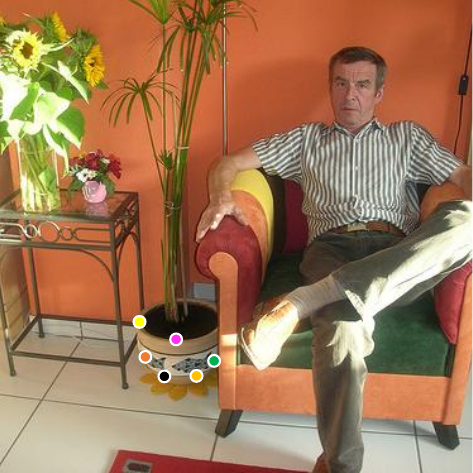}
        \includegraphics[width=0.24\linewidth]{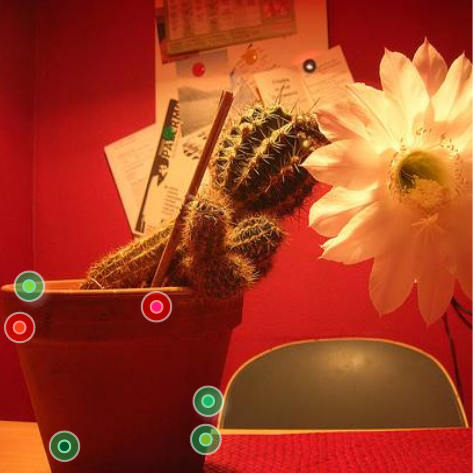}
        \includegraphics[width=0.24\linewidth]{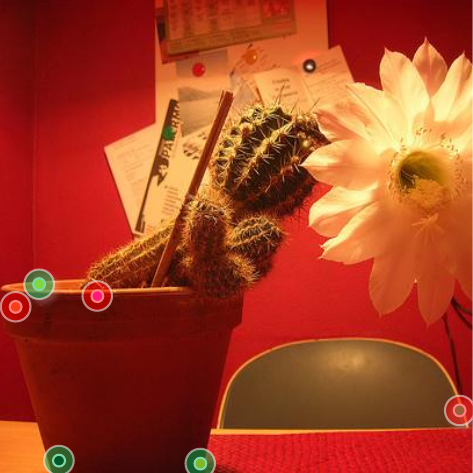}
        \includegraphics[width=0.24\linewidth]{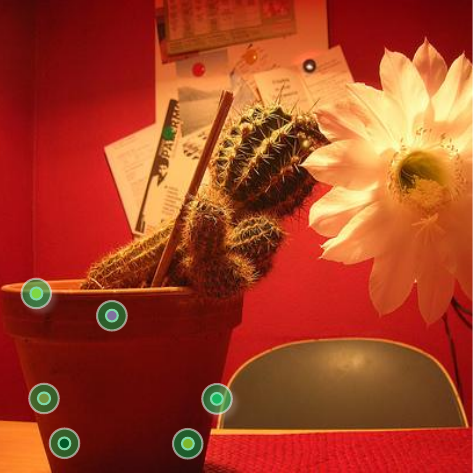}
        \caption{Large scale change, textureless region.}
        \label{fig:failure_c}
    \end{subfigure}
    \hfill
    \begin{subfigure}[t]{0.495\textwidth}
        \centering
        \scriptsize
        \makebox[0.24\linewidth]{Source}\hfill
        \makebox[0.24\linewidth]{DistillDIFT}\hfill
        \makebox[0.24\linewidth]{DINOv2 + SD}\hfill
        \makebox[0.24\linewidth]{Ours}\\[2pt]
        \includegraphics[width=0.24\linewidth]{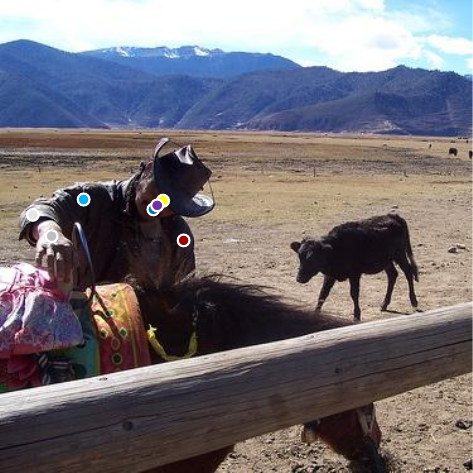}
        \includegraphics[width=0.24\linewidth]{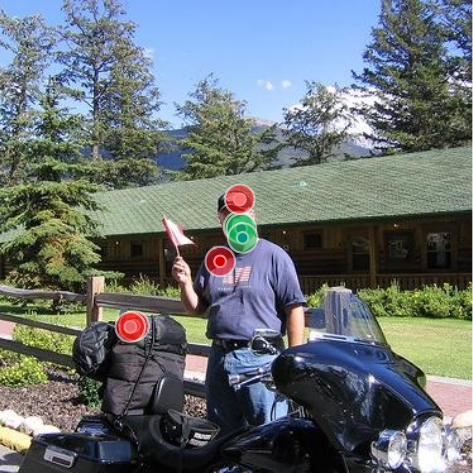}
        \includegraphics[width=0.24\linewidth]{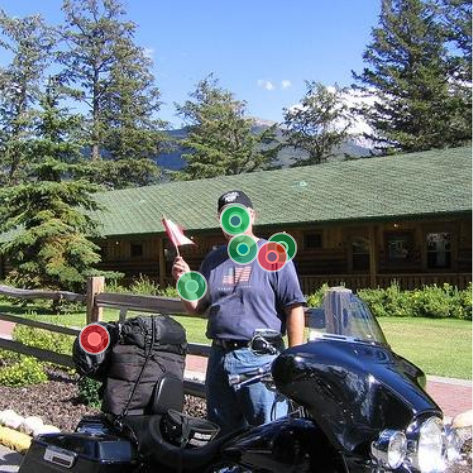}
        \includegraphics[width=0.24\linewidth]{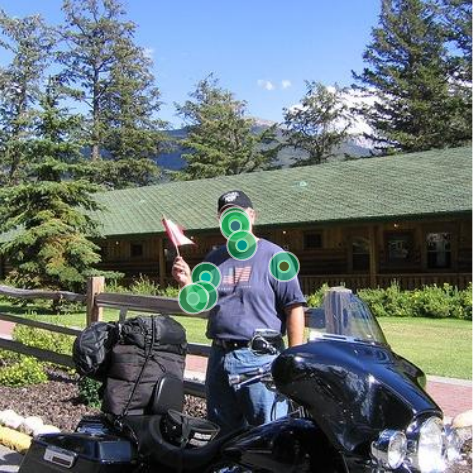}
        \caption{Background clutter, semantic ambiguity.}
        \label{fig:failure_d}
    \end{subfigure}
    \vspace{-2mm}
    \caption{\textbf{Qualitative comparison on challenging cases from SPair-71k.}
    Four challenging scenarios: 
    (a) \textit{geometric ambiguity} from extreme viewpoint changes and occlusion; 
    (b) \textit{extreme pose variation} causing self-occlusion in non-rigid objects; 
    (c) \textit{large scale changes} with feature ambiguity in textureless regions; and 
    (d) \textit{fine-grained semantic ambiguity} from background clutter and intra-class variations.
    \textcolor{green}{Green circles} indicate correct matches (PCK@0.10), \textcolor{red}{red circles} indicate errors.}
    \label{fig:failure_cases}
    \vspace{-4mm}
\end{figure*}

\paragraph{Metric.}\looseness=-1
To measure correspondence accuracy, we use the per-keypoint Percentage of Correct Keypoints (PCK). In this metric, PCK is the percentage of predicted keypoints falling within a specified distance threshold of their ground-truth counterparts.
For all evaluations on SPair-71k and AP-10K, this threshold is defined by the formula $\mathsf{a} \cdot \max(h, w)$, where $h$ and $w$ represent the height and width of the object bounding box and we set $\mathsf{a}\!=\!0.1$.
Following standard evaluation~\cite{min2019spair, zhang2024telling}, the final reported score is the mean per-keypoint PCK across all categories, computed by averaging over all keypoints within each category.
Additionally, to evaluate the quality of the pseudo-labels, we introduce the per-keypoint $\text{PCK}_{\text{label}}$. 
This zero-shot metric applies our pseudo-label generation algorithm directly to SAM~\cite{kirillov2023segment} extracted instances without network training.
Matching is constrained to these regions, evaluating only the subset of ground-truth pairs where the target keypoint lies within the SAM mask.

\subsection{Results}
\paragraph{Quantitative results.}
Tab.~\ref{tab:spair_results} summarizes the quantitative evaluation on SPair-71k. 
Note that despite lacking explicit geometric annotations these methods share implicit or weak supervision: baselines utilize Stable Diffusion or DINOv2 priors whereas we additionally leverage SAM masks and category prompts. 
To ensure a unified comparison, we re-evaluated baselines ($\dagger$ in Tab.~\ref{tab:spair_results}) under the standard $\max(h, w)$ threshold, standardizing DistillDIFT's~\cite{fundel2025distillation} y-scale variant and establishing DINOv2+SD~\cite{zhang2023tale} as a direct zero-shot baseline.
Under identical settings, our method achieves 67.9\% PCK@0.1 (vs. 63.5\% for this zero-shot baseline) and achieves best or second-best performance on 17 out of 18 categories, showing that our 3D geometric structure helps address correspondence challenges difficult to resolve with 2D features alone.
As shown in Tab.~\ref{tab:per-image-pck}, this strong performance extends to stricter thresholds (PCK@0.05 and PCK@0.01) for measuring more precise correspondence capabilities.
To validate generalization, we conduct zero-shot evaluation on the AP-10k dataset~\cite{yu2021ap}, where our method outperforms existing methods across all three challenging settings introduced in~\cite{zhang2024telling}: achieving 68.0\% (intra-species), 65.8\% (cross-species), and 52.9\% (cross-family), surpassing the DINOv2+SD baseline.
This consistent performance on unseen categories demonstrates that our approach avoids overfitting and learns a generalizable representation for geometry-aware correspondence.

\vspace{-3mm}
\paragraph{Qualitative results.}
Fig.~\ref{fig:failure_cases} compares correspondence results on challenging cases with geometric ambiguities. 
DistillDIFT distills features from Stable Diffusion and DINOv2, while DINOv2+SD combines them directly; both rely solely on 2D similarity.
This causes failures under extreme viewpoint and pose variations with occlusions, geometric ambiguities on non-rigid objects, textureless regions with scale variations, and fine-grained semantic ambiguity combined with background clutter.
DistillDIFT's student inherits these limitations from its teachers.
In contrast, our method integrates 3D geometric constraints into pseudo-label generation via the Fused Gromov-Wasserstein framework, jointly optimizing 2D feature similarity and 3D geometric structure.
Results show our approach accurately locates correspondences despite occlusion by leveraging 3D topology, resolves scale changes and texture scarcity using depth information, and distinguishes foreground from background through geometric consistency.
This demonstrates that our geometry-aware pseudo-labeling overcomes 2D limitations, enabling geometrically consistent predictions using only 2D features at inference time.

\begin{figure}[t!]
\centering
\includegraphics[width=1\columnwidth]{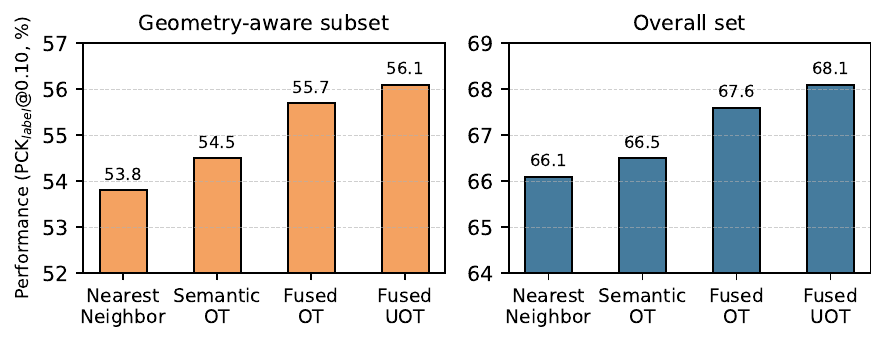}
\vspace{-6mm}
\caption{
\textbf{Ablation study on pseudo-label generation strategies on SPair-71k.}
Performance \textbf{(PCK$_{\text{label}}$@0.1)} on the \textit{Geometry-aware subset} (left) and \textit{Overall set} (right). 
(1) Nearest Neighbor, (2) Semantic OT (feature similarity only), (3) Fused OT (semantic and geometric costs with balanced OT), and (4) our Fused UOT.
}
\label{fig:pseudo_label_analysis}
\vspace{-6mm}
\end{figure}

\begin{figure}[t]
  \centering
  \begin{minipage}[b]{0.49\columnwidth}
    \centering
    \includegraphics[width=\linewidth]{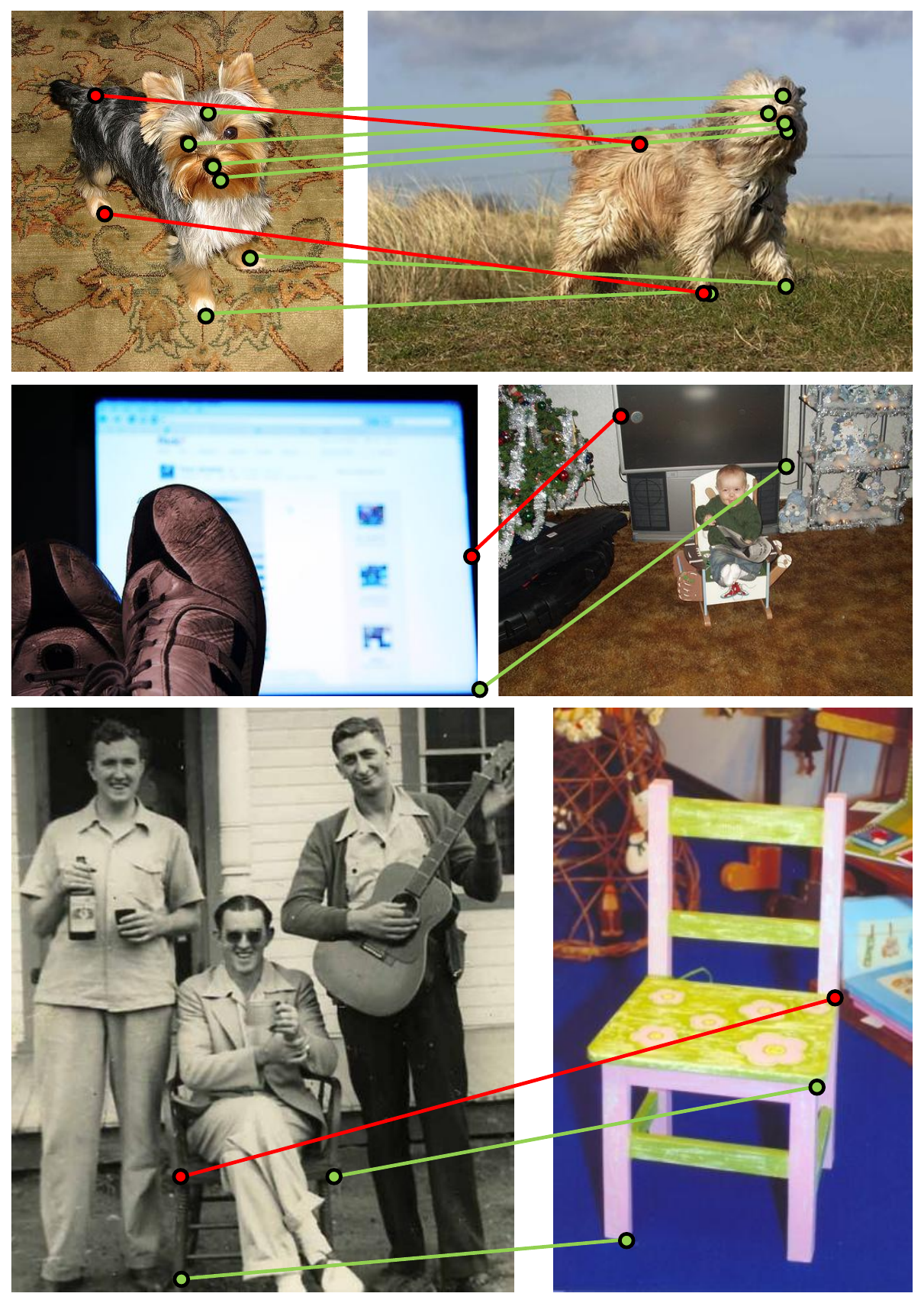}
    \vspace{-6mm}
    \caption*{(a) Pseudo label: NN}
    \vspace{-3mm}
  \end{minipage}
  \hfill
  \begin{minipage}[b]{0.49\columnwidth}
    \centering
    \includegraphics[width=\linewidth]{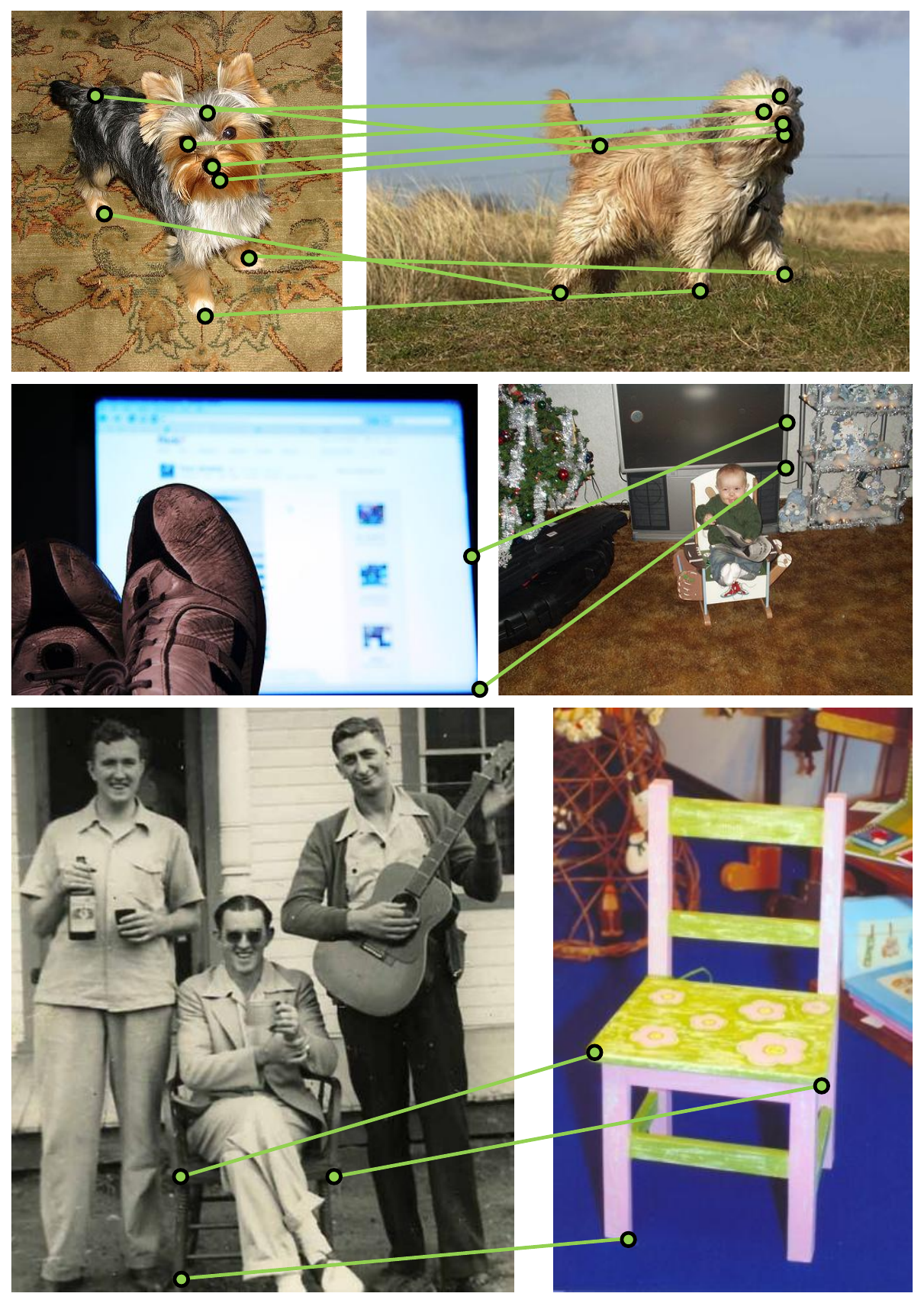}
    \vspace{-6mm}
    \caption*{(b) Pseudo label: Ours}
    \vspace{-2mm}
  \end{minipage}
  \caption{\textbf{Qualitative comparison of pseudo-label generation methods.} 
  (a) Nearest Neighbor matching produces geometrically inconsistent correspondences with multiple errors (\textcolor{red}{red lines}). 
  (b) Our method (\oursAbbrv) generates globally consistent correspondences (\textcolor{green}{green lines}) by leveraging 3D geometric structure.}
  \label{fig:qual_pseudo_label}
\end{figure}

\subsection{Ablation study} 
\paragraph{Pseudo-label strategy comparison.} 
Fig.~\ref{fig:pseudo_label_analysis} provides an in-depth analysis of pseudo-label generation strategies. 
To evaluate the quality of the pseudo-labels themselves, we measure their performance using our previously defined $\text{PCK}_{\text{label}}@0.1$. 
We then compare the performance of the baseline Nearest Neighbor (NN) against our Optimal Transport (OT) based approaches.
As shown in the figure, starting from Nearest Neighbor (53.8\%), we observe improvements through Semantic OT (54.5\%), Fused OT (55.7\%), and finally Fused UOT (56.1\%).
Notably, improvements are larger on the Geometry-aware subset (53.8\%→56.1\%, +2.3\%p), demonstrating that geometric constraints are important when ambiguity exists.

\begingroup
\begin{table}[t!]
\centering
\fontsize{8pt}{9.5pt}\selectfont
\begin{tabular}{l l l c}
    \toprule
    Intra-structure distance for Fused OT & & & PCK$_{\text{label}}$@0.1  \\
    \midrule
    \myrowcolour
    Baseline: Semantic OT (w/o GW term) & & & 66.5 \\
    (1) w/ 2D distance & & & 65.7 \\
    \myrowcolour
    (2) w/ Semantic distance (intra-feature) & & & 66.2 \\
    (3) w/ \textit{3D geometric distance (ours)} & & & \textbf{67.6} \\
    \bottomrule
\end{tabular}
\vspace{-2mm}
\caption{
    \textbf{Ablation on the intra-structure for pseudo-label generation.}
    We compare the \textit{3D geometric structure} (ours) against 2D based 
    alternatives to justify the \textit{Fused OT} strategy (Fig.~\ref{fig:pseudo_label_analysis}).
}
\vspace{-5mm}
\label{tab:ablation_structure}
\end{table}
\endgroup

To further understand the role of geometric information in our fusion strategy, Tab.~\ref{tab:ablation_structure} compares different intra-structure distances for the Fused OT framework. 
Using only semantic similarity without any GW term achieves 66.5\%. 
We find that incorporating 2D distance (65.7\%) or semantic distance for intra-feature cosine similarity (66.2\%) does not improve performance. 
However, leveraging 3D geometric distance achieves 67.6\%, showing that 3D geometric structure is important for resolving correspondence ambiguities. 
This validates our core hypothesis that lifting 2D matches to 3D space enables more reliable pseudo-label generation.

Fig.~\ref{fig:qual_pseudo_label} presents a qualitative comparison between NN and our Fused UOT method. 
While NN produces multiple geometrically inconsistent errors (red lines), our method generates globally consistent correspondences (green lines) by leveraging 3D geometric structure. 
This visual confirmation supports our quantitative findings.

\vspace{-4mm}
\paragraph{Component ablation.}
Tab.~\ref{tab:ablation_component} analyzes the contribution of our method's core components. 
To ensure a fair comparison, all configurations were re-implemented and evaluated within our unified experimental pipeline.
We first measure the zero-shot performance of the base DINOv2 + SD backbone, which yields 63.5\% PCK@0.1 in our setup. 
We then establish adapter baselines trained with standard Nearest Neighbor (NN) pseudo-labels (64.6\%) and a relaxed cycle consistency (c.c.) variant (64.8\%).
While c.c. provides a modest +0.2\%p gain, the overall improvement remains limited, which we attribute to geometrically inconsistent supervision from local matching~\cite{dunkel2025yourself}.
In contrast, training the adapter with our \textit{FGW labels} reaches 66.8\%, outperforming the strongest NN baseline by +2.0\%p and demonstrating the clear benefit of capturing geometric structure beyond 2D cycle consistency.
Finally, incorporating both relaxed c.c. and our \textit{soft target loss} further boosts performance to \textbf{67.9\%}. 
This final gain validates the effectiveness of our proposed loss function for robustly handling the inherent uncertainty in pseudo-label matching, showing that all components synergize to achieve the optimal performance.

\begingroup
\begin{table}[t!]
\centering
\fontsize{8pt}{9.5pt}\selectfont
\begin{tabular}{l l l c}
    \toprule
    Method configuration & & & PCK@0.1 \\
    \midrule
    \myrowcolour
    Backbone (DINOv2 + SD) (zero-shot) & & & 63.5 \\
    \midrule
    Adapter w/ NN labels$^*$ & & & 64.6 \\
    \myrowcolour
    \quad + relaxed c.c.$^*$ & & & 64.8 \\
    \midrule
    Adapter w/ \textit{FGW} labels & & & 66.8 \\
    \myrowcolour
    \quad + relaxed c.c. & & & 67.1 \\
    \quad + \textit{soft target loss} & & & 67.5 \\
    \myrowcolour
    \quad + relaxed c.c. + \textit{soft target loss} & & & \textbf{67.9} \\ 
    \bottomrule
\end{tabular}
\vspace{-2mm}
    \caption{
        \textbf{Ablation study of average PCK@0.1 (per-keypoint) on SPair-71k.} 
        We compare the zero-shot backbone against adapters trained with (1) Nearest Neighbor (NN) and (2) our proposed FGW labels, along with incremental gains from (3) relaxed cycle consistency (c.c.) and (4) soft target loss. ($^*$: baseline variants)
    }
\vspace{-5mm}
\label{tab:ablation_component}
\end{table}
\endgroup

\section{Conclusion} 
\label{sec:conclusion}  
We presented \textbf{Shape-of-You (\oursAbbrv)}, a novel semantic correspondence framework formulating the problem as Fused Gromov-Wasserstein optimal transport, jointly optimizing inter-feature similarity and intra-geometric structure.
By leveraging 3D geometric constraints from foundation models and introducing an efficient anchor-based linearization, we generate structurally consistent pseudo-labels through iterative refinement.
Combined with our soft-target loss handling probabilistic matching uncertainty, our method enables effective learning \textit{without explicit geometric annotations}.
\oursAbbrv~achieves 67.9\% PCK@0.1 on SPair-71k outperforming baselines by 4.4\%p, with consistent gains on AP-10k zero-shot evaluation and particular effectiveness in challenging cases with geometric ambiguity.
Ablation studies validate the contributions of 3D geometric distances, pseudo-label generation via FGW, and soft-target loss.
\vspace{-4mm}
\paragraph{Limitations.}
Our framework relies on 3D foundation models which can occasionally yield flat reconstructions or fail on transparent surfaces. 
Severe geometric ambiguities such as symmetric car parts at mid-range viewpoints can cause erroneous anchor matching and soft targets may sometimes over-smooth geometric signals.

\section*{Acknowledgments}
This work was supported by the National Research Foundation of Korea (NRF) grant funded by the Korea government (MSIT) (No. RS-2025-16068784, 50\%; No. RS-2022-NR070832, 25\%), the Institute of Information \& communications Technology Planning \& Evaluation (IITP) under the artificial intelligence semiconductor support program to nurture the best talents (IITP-(2025)-RS-2023-00253914, 25\%) grant funded by the Korea government (MSIT), and Hanyang University (No. HY202500000003991).

{\small
\bibliographystyle{ieeenat_fullname}
\bibliography{main}

@String(CVPR = {Proceedings of the IEEE Conference on Computer Vision and Pattern Recognition})

@String(ICCV = {Proceedings of the IEEE International Conference on Computer Vision})

@String(ECCV = {Proceedings of the European Conference on Computer Vision})

@String(NIPS = {Advances in Neural Information Processing Systems})

@String(ICML = {International Conference on Machine Learning})

@String(WACV = {IEEE Winter Conference on Applications of Computer Vision})

@String(ICLR = {International Conference on Learning Representations})

@String(SIGGRAPH = {ACM Transactions on Graphics})

@inproceedings{ofri2023neural,
  title={Neural Congealing: Aligning Images to a Joint Semantic Atlas},
  author={Ofri-Amar, Dolev and Geyer, Michal and Kasten, Yoni and Dekel, Tali},
  booktitle = CVPR,
  year={2023}
}

@inproceedings{caron2021emerging,
  title={Emerging Properties in Self-Supervised Vision Transformers},
  author={Caron, Mathilde and Touvron, Hugo and Misra, Ishan and J\'egou, Herv\'e  and Mairal, Julien and Bojanowski, Piotr and Joulin, Armand},
  booktitle= ICCV,
  year={2021}
}

@inproceedings{dust3r_cvpr24,
      title={{DUSt3R: Geometric 3D Vision Made Easy}}, 
      author={Shuzhe Wang and Vincent Leroy and Yohann Cabon and Boris Chidlovskii and Jerome Revaud},
      booktitle = CVPR,
      year = {2024}
}

@inproceedings{mast3r_eccv24,
      title={{Grounding Image Matching in 3D with MASt3R}}, 
      author={Vincent Leroy and Yohann Cabon and Jerome Revaud},
      booktitle = ECCV,
      year = {2024}
}

@inproceedings{wang2025vggt,
  title={{VGGT: Visual Geometry Grounded Transformer}},
  author={Wang, Jianyuan and Chen, Minghao and Karaev, Nikita and Vedaldi, Andrea and Rupprecht, Christian and Novotny, David},
  booktitle= CVPR,
  year={2025}
}

@inproceedings{zhu2025scene,
  title={{Scene-Level Appearance Transfer with Semantic Correspondences}},
  author={Zhu, Liyuan and Cai, Shengqu and Huang, Shengyu and Wetzstein, Gordon and Khosravan, Naji and Armeni, Iro},
  booktitle=SIGGRAPH,
  year={2025}
}

@article{oquab2024dinov,
  title={{DINOv2: Learning Robust Visual Features without Supervision}},
  author={Oquab, Maxime and Darcet, Timoth{\'e}e and Moutakanni, Th{\'e}o and Vo, Huy V and Szafraniec, Marc and Khalidov, Vasil and Fernandez, Pierre and HAZIZA, Daniel and Massa, Francisco and El-Nouby, Alaaeldin and others},
  journal={Transactions on Machine Learning Research},
  year={2024}
}

@inproceedings{wang2024gs,
  title={{GS-Pose: Category-Level Object Pose Estimation via Geometric and Semantic Correspondence}},
  author={Wang, Pengyuan and Ikeda, Takuya and Lee, Robert and Nishiwaki, Koichi},
  booktitle=ECCV,
  year={2024},
}

@article{chen2024zero,
  title={{Zero-shot Image Editing with Reference Imitation}},
  author={Chen, Xi and Feng, Yutong and Chen, Mengting and Wang, Yiyang and Zhang, Shilong and Liu, Yu and Shen, Yujun and Zhao, Hengshuang},
  journal=NIPS,
  year={2024}
}

@inproceedings{puy2020flot,
  title={{Flot: Scene flow on point clouds guided by optimal transport}},
  author={Puy, Gilles and Boulch, Alexandre and Marlet, Renaud},
  booktitle=ECCV,
  year={2020},
}

@inproceedings{sarlin2020superglue,
  title={{Superglue: Learning feature matching with graph neural networks}},
  author={Sarlin, Paul-Edouard and DeTone, Daniel and Malisiewicz, Tomasz and Rabinovich, Andrew},
  booktitle= CVPR,
  year={2020}
}

@inproceedings{cuturi2013sinkhorn,
  title={{Sinkhorn distances: Lightspeed computation of optimal transport}},
  author={Cuturi, Marco},
  booktitle= NIPS,
  year={2013}
}

@inproceedings{fundel2025distillation,
  title={Distillation of diffusion features for semantic correspondence},
  author={Fundel, Frank and Schusterbauer, Johannes and Hu, Vincent Tao and Ommer, Bj{\"o}rn},
  booktitle=WACV,
  year={2025},
}

@inproceedings{tang2023emergent,
  title={Emergent correspondence from image diffusion},
  author={Tang, Luming and Jia, Menglin and Wang, Qianqian and Phoo, Cheng Perng and Hariharan, Bharath},
  booktitle= NIPS,
  year={2023}
}

@inproceedings{gupta2023asic,
  title={Asic: Aligning sparse in-the-wild image collections},
  author={Gupta, Kamal and Jampani, Varun and Esteves, Carlos and Shrivastava, Abhinav and Makadia, Ameesh and Snavely, Noah and Kar, Abhishek},
  booktitle=ICCV,
  year={2023}
}

@inproceedings{zhang2023tale,
  title={{A Tale of Two Features: Stable Diffusion Complements DINO for Zero-Shot Semantic Correspondence}},
  author={Zhang, Junyi and Herrmann, Charles and Hur, Junhwa and Polania Cabrera, Luisa and Jampani, Varun and Sun, Deqing and Yang, Ming-Hsuan},
  booktitle=NIPS,
  year={2023}
}

@inproceedings{rombach2022high,
  title={High-resolution image synthesis with latent diffusion models},
  author={Rombach, Robin and Blattmann, Andreas and Lorenz, Dominik and Esser, Patrick and Ommer, Bj{\"o}rn},
  booktitle=CVPR,
  year={2022}
}

@inproceedings{zhang2024telling,
  title={{Telling left from right: Identifying geometry-aware semantic correspondence}},
  author={Zhang, Junyi and Herrmann, Charles and Hur, Junhwa and Chen, Eric and Jampani, Varun and Sun, Deqing and Yang, Ming-Hsuan},
  booktitle=CVPR,
  year={2024}
}

@inproceedings{mariotti2024improving,
  title={{Improving semantic correspondence with viewpoint-guided spherical maps}},
  author={Mariotti, Octave and Mac Aodha, Oisin and Bilen, Hakan},
  booktitle=CVPR,
  year={2024}
}

@inproceedings{dunkel2025yourself,
  title={{Do It Yourself: Learning Semantic Correspondence from Pseudo-Labels}},
  author={D{\"u}nkel, Olaf and Wimmer, Thomas and Theobalt, Christian and Rupprecht, Christian and Kortylewski, Adam},
  booktitle=ICCV,
  year={2025}
}

@article{min2019spair,
   title={SPair-71k: A Large-scale Benchmark for Semantic Correspondence},
   author={Juhong Min and Jongmin Lee and Jean Ponce and Minsu Cho},
   journal={arXiv prepreint arXiv:1908.10543},
   year={2019}
}

@article{luo2023diffusion,
  title={Diffusion hyperfeatures: Searching through time and space for semantic correspondence},
  author={Luo, Grace and Dunlap, Lisa and Park, Dong Huk and Holynski, Aleksander and Darrell, Trevor},
  journal=NIPS,
  year={2023}
}

@inproceedings{ju2024robo,
  title={{Robo-ABC: Affordance Generalization Beyond Categories via Semantic Correspondence for Robot Manipulation}},
  author={Ju, Yuanchen and Hu, Kaizhe and Zhang, Guowei and Zhang, Gu and Jiang, Mingrun and Xu, Huazhe},
  booktitle=ECCV,
  year={2024},
}

@article{chizat2018scaling,
  title={Scaling algorithms for unbalanced optimal transport problems},
  author={Chizat, Lenaic and Peyr{\'e}, Gabriel and Schmitzer, Bernhard and Vialard, Fran{\c{c}}ois-Xavier},
  journal={Mathematics of computation},
  year={2018}
}

@inproceedings{shtedritski2024shic,
  title={SHIC: Shape-Image Correspondences with no Keypoint Supervision},
  author={Shtedritski, Aleksandar and Rupprecht, Christian and Vedaldi, Andrea},
  booktitle=ECCV,
  year={2024},
}

@article{tang2023fused,
  title={A fused gromov-wasserstein framework for unsupervised knowledge graph entity alignment},
  author={Tang, Jianheng and Zhao, Kangfei and Li, Jia},
  journal={Findings of the Association for Computational Linguistics: ACL},
  year={2023}
}

@inproceedings{yu2021ap,
  title={{Ap-10k: A benchmark for animal pose estimation in the wild}},
  author={Yu, Hang and Xu, Yufei and Zhang, Jing and Zhao, Wei and Guan, Ziyu and Tao, Dacheng},
  booktitle=NIPS,
  year={2021}
}

@article{ryner2023globally,
  title={Globally solving the Gromov-Wasserstein problem for point clouds in low dimensional Euclidean spaces},
  author={Ryner, Martin and Kronqvist, Jan and Karlsson, Johan},
  journal=NIPS,
  year={2023}
}

@inproceedings{xu2024temporally,
  title={{Temporally Consistent Unbalanced Optimal Transport for Unsupervised Action Segmentation}},
  author={Xu, Ming and Gould, Stephen},
  booktitle=CVPR,
  year={2024}
}

@inproceedings{lin2024multi,
  title={{Multi-granularity Correspondence Learning from Long-term Noisy Videos}},
  author={Lin, Yijie and Zhang, Jie and Huang, Zhenyu and Liu, Jia and Wen, Zujie and Peng, Xi},
  booktitle=ICLR, 
  year={2024}
}

@article{eisenberger2020deep,
  title={{Deep Shells: Unsupervised Shape Correspondence with Optimal Transport}},
  author={Eisenberger, Marvin and Toker, Aysim and Leal-Taix{\'e}, Laura and Cremers, Daniel},
  journal=NIPS,
  year={2020}
}

@inproceedings{le2024integrating,
  title={Integrating efficient optimal transport and functional maps for unsupervised shape correspondence learning},
  author={Le, Tung and Nguyen, Khai and Sun, Shanlin and Ho, Nhat and Xie, Xiaohui},
  booktitle=CVPR, 
  year={2024}
}

@inproceedings{xu2019gromov,
  title={{Gromov-Wasserstein Learning for Graph Matching and Node Embedding}},
  author={Xu, Hongteng and Luo, Dixin and Zha, Hongyuan and Duke, Lawrence Carin},
  booktitle=ICML,
  year={2019},
}

@inproceedings{kirillov2023segment,
  title={{Segment Anything}},
  author={Kirillov, Alexander and Mintun, Eric and Ravi, Nikhila and Mao, Hanzi and Rolland, Chloe and Gustafson, Laura and Xiao, Tete and Whitehead, Spencer and Berg, Alexander C and Lo, Wan-Yen and others},
  booktitle=ICCV, 
  year={2023}
}

@inproceedings{titouan2019optimal,
  title={Optimal Transport for structured data with application on graphs},
  author={Titouan, Vayer and Courty, Nicolas and Tavenard, Romain and Flamary, R{\'e}mi},
  booktitle=ICML, 
  year={2019},
}

@inproceedings{peyre2016gromov,
  title={Gromov-wasserstein averaging of kernel and distance matrices},
  author={Peyr{\'e}, Gabriel and Cuturi, Marco and Solomon, Justin},
  booktitle=ICML, 
  year={2016},
}

@article{li2022fast,
  title={{Fast and Provably Convergent Algorithms for Gromov-Wasserstein in Graph Data}},
  author={Li, Jiajin and Tang, Jianheng and Kong, Lemin and Liu, Huikang and Li, Jia and So, Anthony Man-Cho and Blanchet, Jose},
  journal={arXiv preprint arXiv:2205.08115},
  year={2022}
}

@article{sato2020fast,
  title={{Fast and Robust Comparison of Probability Measures in Heterogeneous Spaces}},
  author={Sato, Ryoma and Cuturi, Marco and Yamada, Makoto and Kashima, Hisashi},
  journal={arXiv preprint arXiv:2002.01615},
  year={2020}
}

@article{memoli2011gromov,
  title={{Gromov–Wasserstein Distances and the Metric Approach to Object Matching}},
  author={M{\'e}moli, Facundo},
  journal={Foundations of computational mathematics},
  year={2011},
}

@inproceedings{liu2020semantic,
  title={Semantic Correspondence as an Optimal Transport Problem},
  author={Liu, Yanbin and Zhu, Linchao and Yamada, Makoto and Yang, Yi},
  booktitle = CVPR,
  year={2020}
}

@inproceedings{hartwig2025geco,
  title={GECO: Geometrically Consistent Embedding with Lightspeed Inference},
  author={Hartwig, Regine and Muhle, Dominik and Marin, Riccardo and Cremers, Daniel},
  booktitle = ICCV, 
  year={2025}
}
}

\renewcommand{\thesection}{\arabic{section}}
\renewcommand{\thefigure}{\arabic{figure}}
\renewcommand{\thetable}{\arabic{table}}
\setcounter{section}{0}
\setcounter{figure}{0}
\setcounter{table}{0}
\maketitlesupplementary
\appendix
Before presenting the extended analyses, we briefly outline the structure of this supplementary material, which provides further insights to complement the main paper.
\begin{itemize}
    \item \textbf{Sec.~\ref{sec:additional}: Additional experiments.}
    \begin{itemize}
        \item Sec.~\ref{sec:backbone} examines the effect of different 2D feature backbones, comparing DINOv2~\cite{caron2021emerging} and fused DINOv2+SD~\cite{zhang2023tale} in terms of pseudo-label quality and final correspondence accuracy.
        \item Sec.~\ref{sec:pseudo_label_sensitivity} analyzes key design choices of the pseudo-label generator, including the number of anchors $K$, the feature–geometry trade-off $\alpha$, the KL regularization strength $\rho$ in UOT, and the choice of 3D foundation backbone~\cite{wang2025vggt, dust3r_cvpr24}.
        \item Sec.~\ref{sec:training_hparam_sensitivity} investigates training-time hyperparameters, focusing on the soft-target mixing weight $\beta$ and its synergy with pseudo-label quantity (top-$k$) and cycle-consistency.
        \item Sec.~\ref{sec:systematic_evaluation} provides a systematic evaluation across various challenging conditions (e.g., viewpoint, occlusion, pose) and an in-depth per-category analysis.
    \end{itemize}

    \item \textbf{Sec.~\ref{sec:failure}: Failure cases.}  
    Typical failure cases of the pseudo-label generator are summarized and discussed.

    \item \textbf{Sec.~\ref{sec:implementation_details}: Implementation details.}  
    We provide low-level implementation details and the full set of hyperparameters used in all experiments.

    \item \textbf{Sec.~\ref{sec:algorithm}: Algorithm.}  
    We present PyTorch-style pseudocode describing the complete pseudo-label generation pipeline.

    \item \textbf{Sec.~\ref{sec:visualization}: Additional visualizations.}  
    Additional qualitative visualizations on SPair-71k are provided to complement the quantitative results in the main paper.
\end{itemize}

\section{Additional experiments}
\label{sec:additional}
We provide additional experiments and analyses to complement the main paper.
This section is organized into four parts:
\textbf{(a)} Sec.~\ref{sec:backbone} studies how our method behaves under different 2D foundation feature backbones, comparing DINOv2 and fused DINOv2+SD features in terms of both pseudo-label quality and final correspondence accuracy;
\textbf{(b)} Sec.~\ref{sec:pseudo_label_sensitivity} analyzes the sensitivity of our pseudo-label generator to key design choices, including the number of anchors $K$, the feature–geometry trade-off $\alpha$, the KL regularization strength $\rho$ in UOT, and the choice of 3D foundation backbone;
\textbf{(c)} Sec.~\ref{sec:training_hparam_sensitivity} investigates training hyperparameter sensitivity, focusing on the soft-target mixing weight $\beta$ in our loss;
and \textbf{(d)} Sec.~\ref{sec:systematic_evaluation} provides a systematic evaluation across various challenging conditions and an in-depth per-category analysis.

\subsection{Backbone}
\label{sec:backbone}
\paragraph{Pseudo-labels with DINOv2 backbone.}

\begin{figure}[t] 
\centering
\includegraphics[width=1\columnwidth]{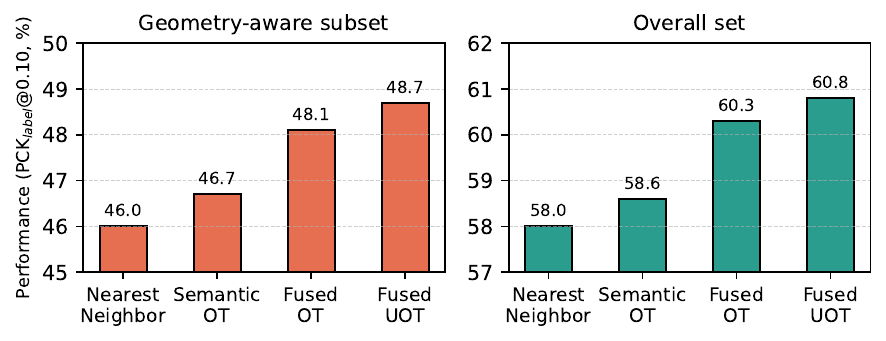}
\vspace{-7mm}
\caption{\textbf{Pseudo-label analysis with DINOv2 backbone.} 
Performance \textbf{(PCK$_{\text{label}}$@0.1)} on the \textit{geometry-aware subset} (left) and \textit{overall set} (right). 
Using only DINOv2 features, we compare four pseudo-label generation strategies: Nearest Neighbor, Semantic OT, Fused OT, and Fused UOT.
Both the geometry-aware and overall subsets show a consistent, monotonic improvement across methods, with clear gains from incorporating semantic OT and our geometry-aware matching.
This confirms that our pseudo-labeling remains effective even with a weaker backbone.}
\vspace{-3mm}
\label{fig:pseudo_label_analysis_dino}
\end{figure}

As in the main paper, the $\text{PCK}_{\text{label}}$ metric evaluates only target keypoints within the SAM mask. 
We note that evaluating the DINOv2+SD baseline across all ground-truth keypoints yields identical improvement trends (NN 62.4\% $\rightarrow$ Semantic OT 63.4\% $\rightarrow$ Fused OT 64.6\% $\rightarrow$ Fused UOT 64.9\%) demonstrating robustness beyond the masked regions.
To verify our conclusions from Fig.~\ref{fig:pseudo_label_analysis} are not tied to this backbone choice we repeat the evaluation using \emph{only} DINOv2 features.
Fig.~\ref{fig:pseudo_label_analysis_dino} reports the PCK$_{\text{label}}@0.1$ results. 
On the geometry-aware subset, performance improves from 46.0\% (Nearest Neighbor) to 46.7\% (Semantic OT), 48.1\% (Fused OT) and 48.7\% (Fused UOT), a +2.7\%p gain over the NN baseline.
A similar trend occurs on the overall set, with performance increasing from 58.0\% (NN) to 58.6\% (Semantic OT), 60.3\% (Fused OT) and 60.8\% (Fused UOT), a +2.8\%p total gain.
Although the absolute PCK$_{\text{label}}$ values are lower than with the DINOv2+SD backbone, the \emph{relative} improvements from incorporating geometric consistency remain consistent. 
This confirms our pseudo-labeling is complementary to the visual backbone and provides benefits with different feature representations.

\begin{figure*}[t]
\centering
\includegraphics[width=\textwidth]{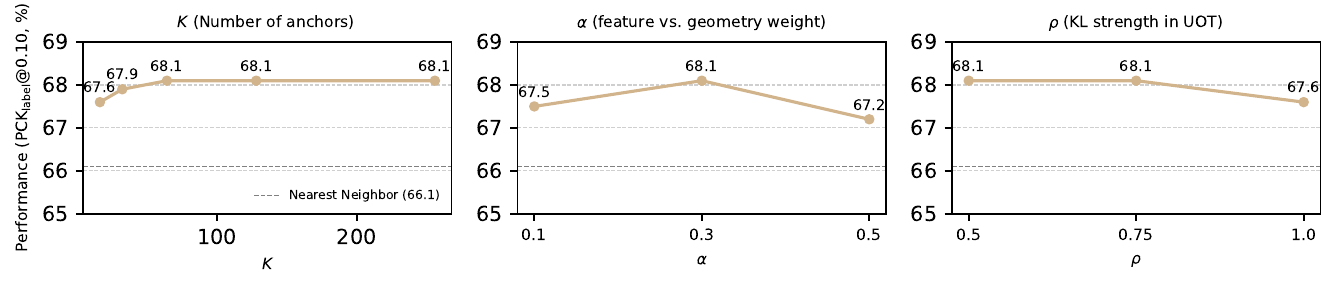}
\vspace{-9mm}
\caption{\textbf{Pseudo-label hyperparameter ablation on SPair-71k (PCK$_{\text{label}}$@0.1).}
We study three key hyperparameters of our pseudo-label generator:
\textbf{(Left)} number of anchors $K$, 
\textbf{(Middle)} feature–geometry trade-off $\alpha$, and 
\textbf{(Right)} KL regularization strength $\rho$ in UOT.
In all cases, the performance is measured as PCK$_{\text{label}}$@0.1 on the SPair-71k test set.
}
\label{fig:hyperparam_pcklabel}
\vspace{-3mm}
\end{figure*}

\vspace{-3mm}
\paragraph{Evaluation across different feature backbones.}
Tab.~\ref{tab:pck_backbones} further compares the final correspondence accuracy when training our model using pseudo-labels generated from the fused DINOv2+SD backbone, while evaluating the trained model under two different backbone choices: (1) DINOv2-only features and (2) DINOv2+SD features.
Note that these adapter evaluations were conducted prior to the integration of the relaxed cycle-consistency filtering discussed in the main paper.
Therefore, the reported PCK scores isolate the impact of the backbone and the base FGW pseudo-labels without the additional cycle-consistency boost.
Across both settings, our method consistently improves PCK at all thresholds, confirming that the geometric priors introduced by our matching remain beneficial regardless of the underlying feature strength.
With the weaker DINOv2 backbone, our approach yields substantial gains over the baseline (+1.2, +1.9, +4.8 at PCK@0.01/0.05/0.1 respectively), demonstrating that our pseudo-labels help compensate for limited semantic discriminability in the features.
When moving to the stronger DINOv2+SD backbone, the absolute accuracy increases—as expected from higher-quality visual features—yet our method continues to provide additional improvements (+0.2, +0.7, +4.0).
Interestingly, the relative gain at the higher threshold (PCK@0.1) remains particularly large in both cases, suggesting that the geometric consistency enforced by our FGW matching systematically enhances mid-to-coarse correspondence quality.
Overall, these results reinforce that our improvements do not rely on a specific backbone and that SoY complements both standard visual features and diffusion priors. 
The model consistently benefits from pseudo-labels generated with DINOv2+SD, even when the evaluation backbone differs, indicating that our geometric alignment signal generalizes robustly across feature extractors.

\begin{table}[t]
\centering
\begin{adjustbox}{width=0.95\linewidth}
\begin{tabular}{lccc}
\toprule
Method & PCK@0.01 & PCK@0.05 & PCK@0.1 \\
\midrule

\multicolumn{4}{c}{\textit{\color{white!30!black} DINOv2 Backbone}} \\
\cmidrule(lr){1-4}
\myrowcolour
DINOv2~\cite{caron2021emerging} & 6.4 & 40.2 & 55.7 \\
\oursTab~(DINOv2) 
  & 7.6 {\color{red}(+1.2)}
  & 42.1 {\color{red}(+1.9)}
  & 60.5 {\color{red}(+4.8)} \\
\midrule
\multicolumn{4}{c}{\textit{\color{white!30!black} DINOv2 + SD Backbone}} \\
\cmidrule(lr){1-4}
\myrowcolour

DINOv2 + SD~\cite{zhang2023tale} & 8.8 & 48.3 & 63.5 \\
\oursTab~(DINOv2 + SD) 
  & 9.0  {\color{red}(+0.2)}
  & 49.0 {\color{red}(+0.7)}
  & 67.5 {\color{red}(+4.0)} \\
\bottomrule
\end{tabular}
\end{adjustbox}
\caption{Comparison of PCK scores (PCK@0.01, 0.05, 0.1) with different feature backbones.
All models are trained using pseudo-labels generated from DINOv2+SD without relaxed cycle-consistency. 
Our method improves over both baselines with gains shown in red.}
\label{tab:pck_backbones}
\vspace{-3mm}
\end{table}

\subsection{Pseudo-label sensitivity}
\label{sec:pseudo_label_sensitivity}
\paragraph{Number of anchors $K$.}
Fig.~\ref{fig:hyperparam_pcklabel} (left) shows the effect of varying the number of anchors $K \in \{16, 32, 64, 128, 256\}$ used in the anchor-based FGW linearization.
PCK$_{\text{label}}$@0.1 improves from 67.6\% at $K{=}16$ to 68.1\% at $K{=}64$, and then saturates (68.1\% for $K{=}64,128,256$).
This indicates that our method is not overly sensitive to the exact choice of $K$ once a moderate number of anchors is available, and that $K{=}64$ provides a good trade-off between robustness and computational cost.
\vspace{-4mm}
\paragraph{Feature–geometry weight $\alpha$.}
Fig.~\ref{fig:hyperparam_pcklabel} (middle) studies the fusion weight $\alpha \in \{0.1, 0.3, 0.5\}$ between feature similarity and geometric cost in the fused OT objective.
We observe a clear peak at $\alpha{=}0.3$ (68.1\%), whereas both a too feature-dominated setting ($\alpha{=}0.1$, 67.5\%) and a too geometry-dominated setting ($\alpha{=}0.5$, 67.2\%) lead to lower
PCK$_{\text{label}}$.
This confirms that balancing semantic and geometric cues is important, and our default choice $\alpha{=}0.3$ lies near the optimum.
\vspace{-4mm}
\paragraph{KL strength $\rho$ in UOT.}
Fig.~\ref{fig:hyperparam_pcklabel} (right) analyzes the KL regularization strength $\rho \in \{0.5, 0.75, 1.0\}$ in the unbalanced OT formulation.
PCK$_{\text{label}}$@0.1 remains high and stable for $\rho{=}0.5$ and $\rho{=}0.75$ (both 68.1\%), but slightly decreases at $\rho{=}1.0$ (67.6\%).
This suggests that overly strong KL regularization, which enforces the marginals too strictly, can harm pseudo-label quality, while moderate relaxation yields more robust correspondences.
We therefore use $\rho{=}0.75$ in all our experiments.


\begin{table}[t]
\centering
\small
\begin{tabular}{lc}
\toprule
3D backbone & PCK$_{\text{label}}$@0.1 \\
\midrule
\myrowcolour
VGGT~\cite{wang2025vggt}   & \textbf{68.1} \\
DUSt3R~\cite{dust3r_cvpr24} & 67.1 \\
\bottomrule
\end{tabular}
\vspace{-1mm}
\caption{Effect of the 3D foundation backbone on pseudo-label quality on SPair-71k.
We report PCK$_{\text{label}}$@0.1 for our pseudo-labels when using either VGGT or DUSt3R to obtain 3D structure.}
\vspace{-3mm}
\label{tab:3d_backbone}
\end{table}

\paragraph{3D foundation backbone.}
Finally, we study the impact of the underlying 3D foundation model used to obtain geometric structure.
In all our main experiments, we adopt VGGT as the 3D backbone, which yields PCK$_{\text{label}}$@0.1 of 68.1\% for our pseudo-labels.
Tab.~\ref{tab:3d_backbone} compares this setting with an alternative 3D foundation model, DUSt3R~\cite{dust3r_cvpr24}.
DUSt3R is originally designed for multi-view 3D reconstruction and correspondence, where several views of the \emph{same} scene are jointly encoded to recover accurate geometry.
In our semantic correspondence setting, however, the source and target images typically depict different scenes or object instances, so each image is effectively processed in a single-view regime.
As also noted in the VGGT paper~\cite{wang2025vggt}, DUSt3R’s reconstruction quality degrades noticeably when only a single RGB view is available.
Consistent with this observation, replacing VGGT with DUSt3R leads to a modest drop in pseudo-label quality from 68.1\% to 67.1\% PCK$_{\text{label}}$@0.1.
Nevertheless, even with DUSt3R our method still improves over the nearest-neighbor baseline, indicating that the FGW formulation can still exploit 3D cues as long as the backbone provides reasonably stable structure.
\vspace{-1mm}
\subsection{Training hyperparameter sensitivity}
\label{sec:training_hparam_sensitivity}

\begin{table}[t]
\centering
\vspace{-1mm}
\small 
\begin{adjustbox}{width=0.85\linewidth} 
\begin{tabular}{lccc}
\toprule
$\beta$ & PCK@0.01 & PCK@0.05 & PCK@0.1 \\
\midrule
\myrowcolour
0.25 & \textbf{9.3} & \textbf{49.4} & 67.2 \\
0.50 (default) & 9.0 & 49.0 & \textbf{67.5} \\
\myrowcolour
0.75 & 6.4 & 44.9 & 65.2 \\
\bottomrule
\end{tabular}
\end{adjustbox}
\vspace{-2mm}
\caption{Sensitivity of the soft-target weight $\beta$ in the training loss on SPair-71k after 20 epochs. 
Evaluations here are conducted without relaxed cycle-consistency.
Within the range $\beta \in [0.25, 0.50]$, performance varies only mildly, whereas a larger value $\beta{=}0.75$ noticeably degrades accuracy.}
\vspace{-3mm}
\label{tab:beta_ablation}
\end{table}

\paragraph{Soft-target weight $\beta$.}
We first ablate the soft-target weight $\beta$ used in our training loss while keeping all other settings fixed.
Note that to strictly isolate the effect of $\beta$, this evaluation does not employ the relaxed cycle-consistency filtering.
Tab.~\ref{tab:beta_ablation} reports PCK scores on SPair-71k after 20 epochs for $\beta \in \{0.25, 0.50, 0.75\}$.
Within the range $\beta \in [0.25, 0.50]$, the performance is fairly stable: 
PCK@0.10 changes only slightly (67.2\% vs.\ 67.5\%) and the differences at stricter thresholds (PCK@0.01/0.05) are within 0.4 points.
In contrast, a larger value $\beta{=}0.75$ substantially degrades performance (6.4/44.9/65.2 at PCK@0.01/0.05/0.10), indicating that over-emphasizing noisy soft targets can be harmful.

We also observe that when training is extended to roughly 50 epochs, the default setting $\beta{=}0.50$ yields a slight further improvement (from 9.0/49.0/67.5 to 9.3/49.5/67.7 at PCK@0.01/0.05/0.10).
This suggests that a moderate soft-target weight may require a few more epochs to fully exploit the denoising effect of the soft supervision 
whereas a smaller value $\beta{=}0.25$ reaches its best performance earlier but tends to plateau.
Overall, these results indicate that our method is reasonably robust to the choice of $\beta$ as long as it lies in a moderate range (e.g., $0.25\text{--}0.50$) 
while too large values giving excessive emphasis to noisy labels (\emph{e.g.}, $\beta{=}0.75$) should be avoided.

\begin{table}[t]
\centering
\small
\begin{adjustbox}{width=\linewidth}
\begin{tabular}{lccccc}
\toprule
\multirow{2}{*}{Setting (hard labels, $\beta{=}0$)} & \multicolumn{4}{c}{Top-$k$ candidates} & \multirow{2}{*}{\shortstack{Ours \\ ($\beta{=}0.5, k{=}3$)}} \\
\cmidrule(lr){2-5}
& $k{=}1$ & $k{=}3$ & $k{=}5$ & $k{=}10$ & \\
\midrule
\myrowcolour
w/o relaxed c.c. & 66.8 & 66.8 & 66.7 & 66.2 & - \\
w/ relaxed c.c. & 67.1 & 67.4 & 67.4 & 66.8 & \textbf{67.9} \\
\bottomrule
\end{tabular}
\end{adjustbox}
\vspace{-2mm}
\caption{Synergy between pseudo-label quantity (top-$k$) and soft-target loss on SPair-71k (PCK@0.1). 
Filtering multiple candidates with relaxed cycle-consistency improves performance, and combining them with our soft-target loss yields the optimal result.}
\vspace{-3mm}
\label{tab:topk_ablation}
\end{table}

\vspace{-3mm}
\paragraph{Synergy with top-$k$ and cycle-consistency.}
To further investigate the relationship between pseudo-label quantity and our proposed training components, Tab.~\ref{tab:topk_ablation} analyzes the impact of retrieving multiple matches (top-$k$). 
When training with hard labels ($\beta{=}0$) without cycle-consistency, increasing the number of candidates $k$ introduces excessive noise causing accuracy to drop ($66.8\% \rightarrow 66.2\%$ for $k{=}10$). 
However, applying relaxed cycle-consistency effectively filters this noise allowing the model to benefit from the richer candidate pool and peaking at $k{=}3, 5$ ($67.4\%$). 
Finally, combining this expanded cycle-consistent candidate set with our soft-target loss ($\beta{=}0.5$) achieves the optimal performance of \textbf{67.9\%}, demonstrating the complementary benefits of geometric filtering and soft supervision.

\begin{table*}[t]
\centering
\small
\begin{adjustbox}{width=0.9\linewidth} 
\begin{tabular}{l|ccccc|ccccc|ccc|ccccc}
\toprule
& \multicolumn{10}{c|}{\textbf{(a) Azimuth Analysis (All $|$ Car (C))}} & \multicolumn{3}{c|}{\textbf{(b) Occlusion}} & \multicolumn{5}{c}{\textbf{(c) Pose}} \\
\cmidrule(lr){2-11} \cmidrule(lr){12-14} \cmidrule(lr){15-19}
Variable & 0° & 45° & 90° & 135° & 180° & 0°(C) & 45°(C) & 90°(C) & 135°(C) & 180°(C) & None & Part. & Heavy & F$\rightarrow$F & L$\rightarrow$L & R$\rightarrow$R & Un. & Cross \\
\midrule
\myrowcolour
NN & 65.3 & 65.2 & 58.9 & 57.5 & 57.3 & 88.1 & 65.0 & 39.4 & 23.1 & 15.3 & 63.8 & 58.8 & 59.4 & 62.4 & 74.8 & 76.7 & 57.7 & 67.3 \\
Ours & 69.9 & 67.9 & 59.7 & 58.3 & 58.6 & 89.7 & 61.9 & 33.6 & 20.9 & 17.1 & 66.4 & 61.4 & 61.9 & 70.6 & 75.9 & 78.7 & 60.8 & 68.9 \\
\rowcolor{gray!15}
$\Delta$ & +4.6 & +2.5 & +0.8 & +0.7 & +1.3 & +1.6 & \textcolor{red}{-3.1} & \textcolor{red}{-5.8} & \textcolor{red}{-2.2} & +1.8 & +2.6 & +2.6 & +2.5 & +8.2 & +1.1 & +2.0 & +3.1 & +1.6 \\
\bottomrule
\end{tabular}
\end{adjustbox}
\vspace{-2mm}
\caption{Systematic evaluation of pseudo-label quality ($\text{PCK}_{\text{label}}$@0.1) across various challenging conditions. 
Our framework provides consistent gains in most scenarios particularly under extreme pose variations and heavy occlusions while also revealing specific challenges in mid-range azimuths for symmetric objects.}
\label{tab:integrated_analysis}
\end{table*}

\subsection{Systematic evaluation \& per-category analysis}
\label{sec:systematic_evaluation}

\paragraph{Systematic evaluation.}
To provide a deeper understanding of our method under challenging in-the-wild conditions, we systematically evaluate pseudo-label quality categorizing the test set by azimuth difference occlusion level and pose.
Tab.~\ref{tab:integrated_analysis} summarizes these results. 
Our approach yields consistent improvements across most variations demonstrating strong robustness against severe occlusions (+2.5\%p) and challenging cross-pose alignments (+1.6\%p).
The notable gain in frontal same-pose pairs (+8.2\%p) further confirms that our geometric lifting effectively refines ambiguous 2D semantic matches.

\paragraph{Analysis of specific challenges (car and boat).}
While our framework improves pseudo-label quality across 17 out of 18 categories, we provide an in-depth analysis of specific cases where distinct challenges remain.
For the \textit{car} category, degradation primarily occurs at mid-range viewpoints (45°--135°). 
This stems from severe semantic aliasing between symmetric parts such as front and rear windshields. 
When these visually similar structures are incorrectly matched as initial anchors they often maintain structurally plausible geometric relationships. 
Consequently the linearized GW cost propagates these initial errors rather than correcting them highlighting a fundamental limitation when severe 2D ambiguity aligns with 3D structural symmetry.

For the \textit{boat} category, incorporating our relaxed cycle-consistency successfully improves the final accuracy beyond the zero-shot baseline. 
However, we hypothesize that soft targets may over-smooth the geometric signal. 
In failure cases, the trained model tends to match similar local features rather than preserving the global geometric structure. 
In both cases, we leave more detailed investigation for future work.


\section{Failure cases}
\begin{figure}[t]
\centering
\begin{subfigure}[b]{0.49\columnwidth}
    \centering
    \includegraphics[width=0.9\linewidth]{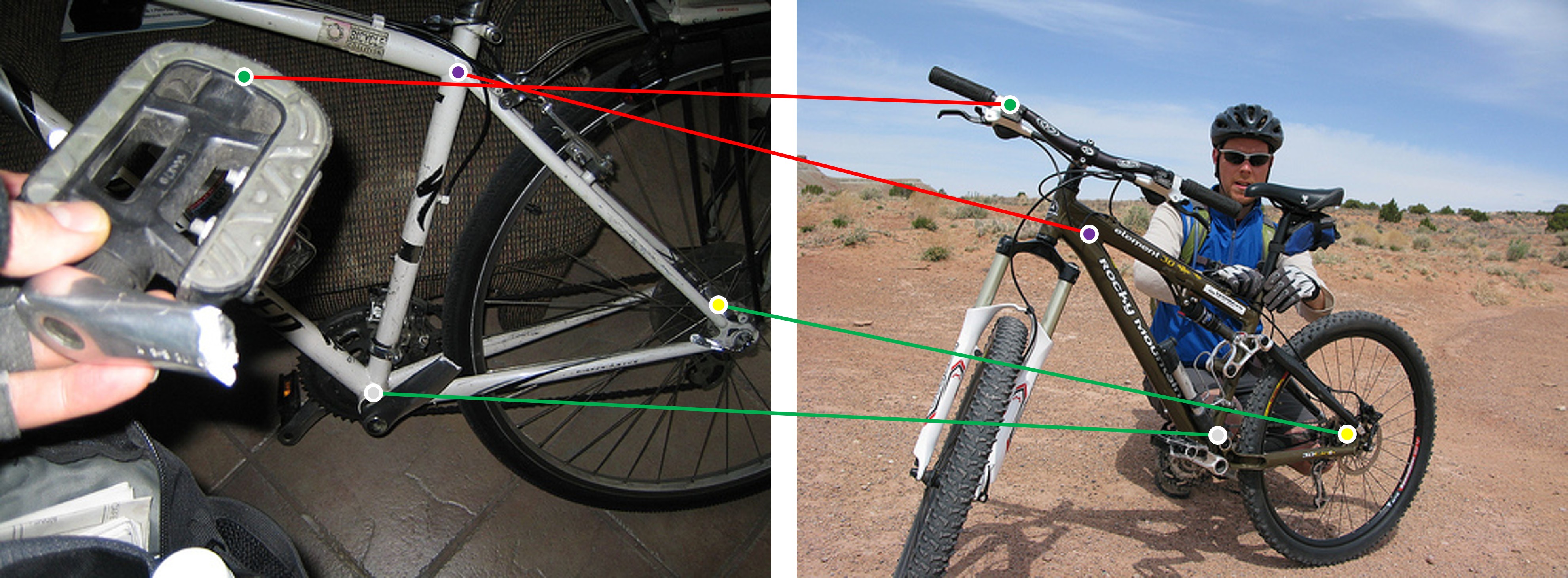}
    \caption{NN (broken structure)}
    \label{fig:failure_nn}
\end{subfigure}
\hfill
\begin{subfigure}[b]{0.49\columnwidth}
    \centering
    \includegraphics[width=0.9\linewidth]{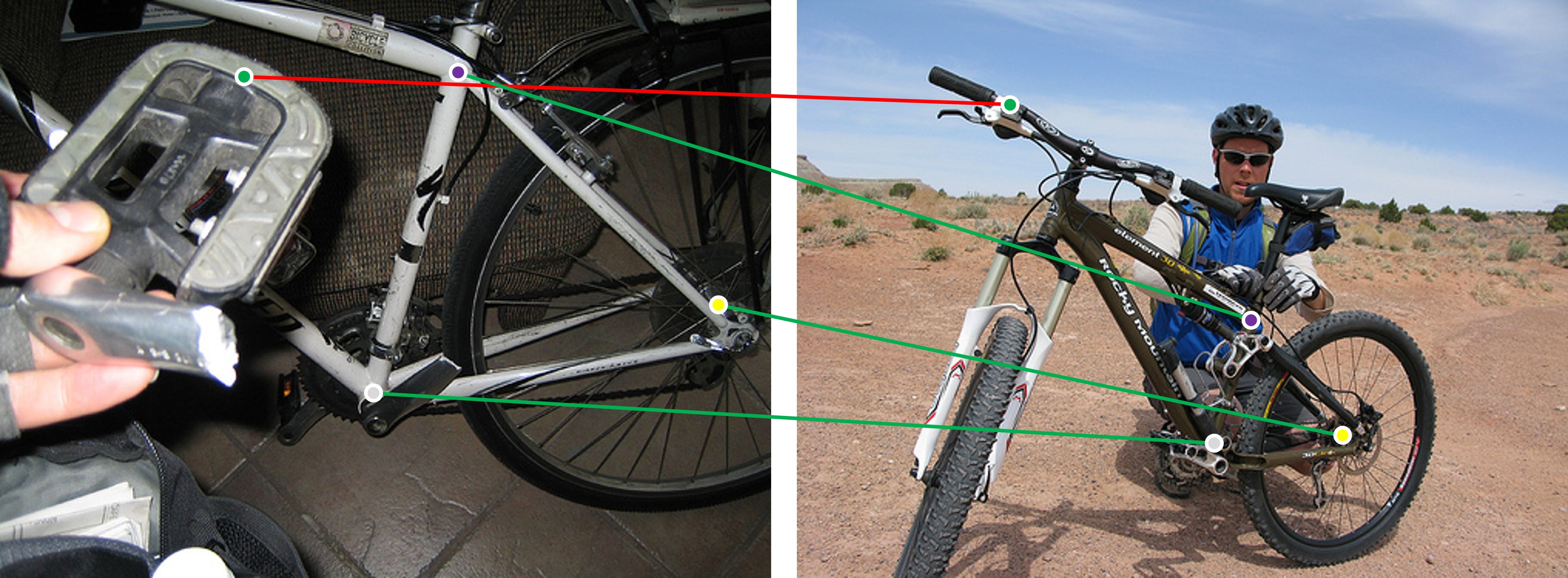}
    \caption{Ours (broken structure)}
    \label{fig:failure_ours}
\end{subfigure}

\vspace{2mm}

\begin{subfigure}[b]{0.49\columnwidth}
    \centering
    \includegraphics[width=0.9\linewidth]{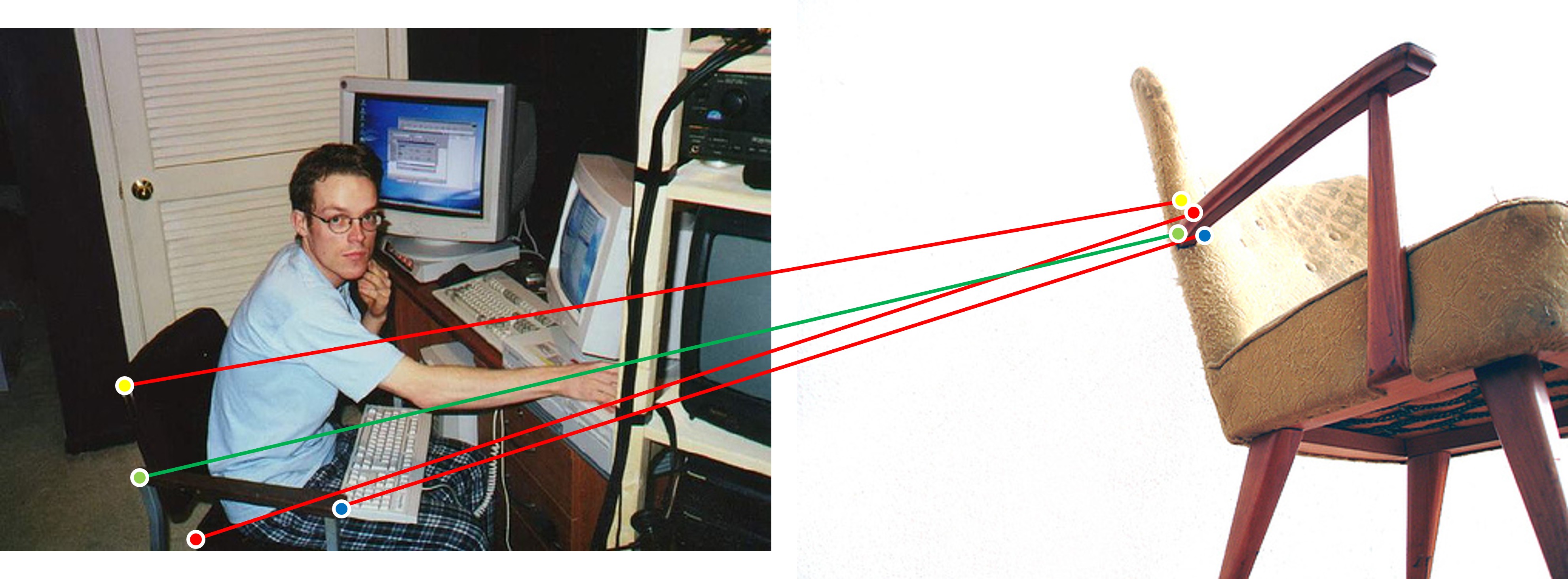}
    \caption{NN (noisy features)}
    \label{fig:failure_nn2}
\end{subfigure}
\hfill
\begin{subfigure}[b]{0.49\columnwidth}
    \centering
    \includegraphics[width=0.9\linewidth]{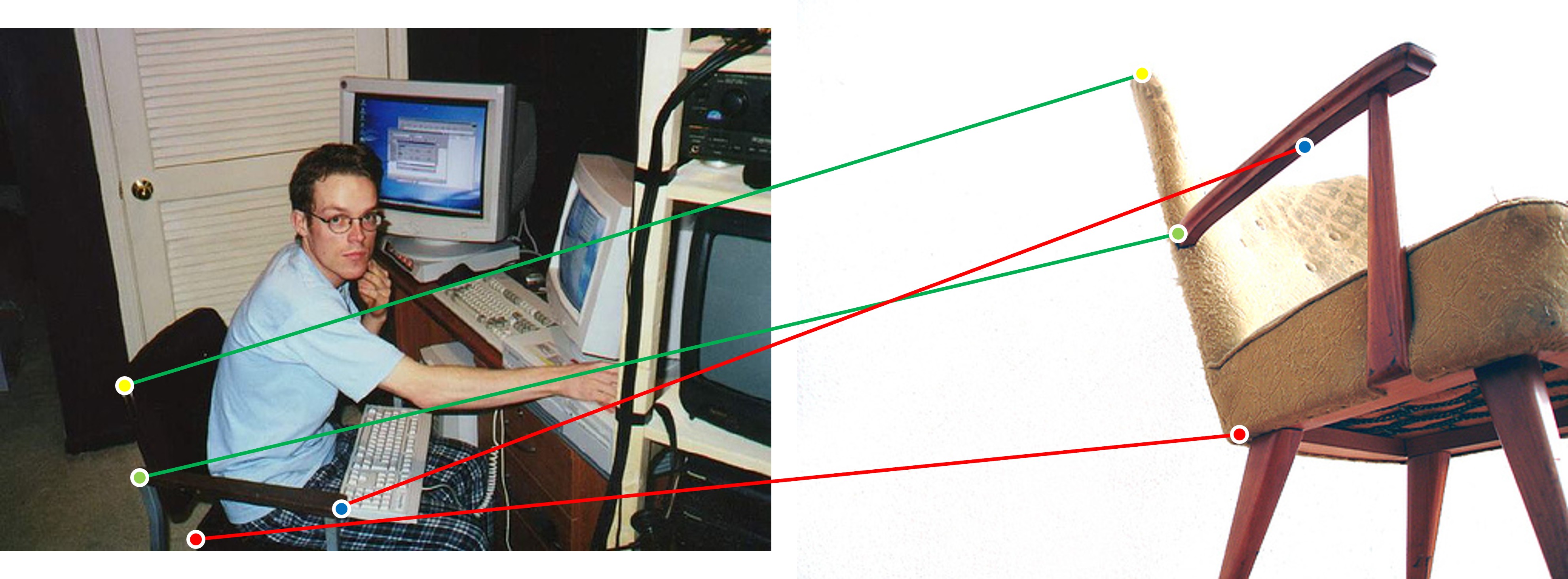}
    \caption{Ours (noisy features)}
    \label{fig:failure_ours2}
\end{subfigure}

\vspace{-2mm}
\caption{\textbf{Failure modes of pseudo-label generation.}
Top: failure when the object breaks into disconnected parts.
Bottom: failure when the features used for pseudo labels are highly noisy and all correspondences collapse to a local region.}
\vspace{-3mm}
\label{fig:failure_pseudo}
\end{figure}
\paragraph{Pseudo-label failure modes.}
\label{sec:failure}
Fig.~\ref{fig:failure_pseudo} summarizes two typical failure modes of our pseudo-labels. 
In the first row (broken structure), nearest-neighbor (NN) matching produces an incorrect correspondence along the bicycle frame, whereas our method corrects the violet keypoint by enforcing global geometric consistency. 
However, when the object is physically broken into disconnected parts (the detached pedal), the green keypoint has no structurally consistent counterpart, so our FGW refinement can no longer rely on geometry and therefore fails to update the NN pseudo-label.
In the second row (noisy features), the DINOv2+SD features used for pseudo-label generation are themselves highly ambiguous, causing multiple points to collapse onto a small spurious region on the chair. 
Since both the semantic cost and anchor selection are driven by these noisy similarities, our refinement is also misled and cannot recover the correct correspondences. 
Taken together, these examples highlight that our pseudo-label generator still depends on reasonably coherent 3D structure and sufficiently informative 2D features to produce reliable matches. 
In rare cases, such incorrectly generated pseudo-labels may provide slightly inconsistent supervision during training and can mildly bias the learned matcher, suggesting an interesting direction for making our framework more robust to pseudo-label noise in future work.

\section{Implementation details}
\label{sec:implementation_details}

\begin{table}[t]
    \centering
    \begin{adjustbox}{width=\linewidth}
    \begin{tabular}{llp{6cm}}
        \toprule
        \textbf{Group} & \textbf{Parameter} & \textbf{Value} \\
        \midrule

        \multirow{4}{*}{Semantic UOT}  
            & KL penalty $\rho$ & 0.75 (UOT marginal relaxation) \\
            & Entropy $\varepsilon$ & 1 (implicit Sinkhorn regularization) \\
            & Mass distribution & Uniform over valid patches \\
            & Cost $C_{\text{sem}}$ & $1 - \text{cosine\_sim}(f_i, f_j)$ \\

        \midrule

        \multirow{3}{*}{FGW Fusion}
            & Fusion weight $\alpha$ & 0.3 (semantic vs.\ geometric balance) \\
            & Distance metric & 3D Euclidean distance in lifted space \\
            & Normalization & Both costs normalized before fusion \\

        \midrule

        \multirow{5}{*}{FGW Refinement}
            & Anchor count $K$ & 64 anchors per iteration \\
            & Iterations $T$ & 5 refinement steps \\
            & Cycle-consistency $\delta$ & Quantile threshold $q = 0.01$ \\
            & Anchor mass & Uniform: $\hat{\pi}_{i'j'} = 1/K$ \\
            & Linearized cost & $C_{\text{geo}}(i,j)$=$\frac{1}{K}$$\sum |D_A(i,a_A)-D_B(j,a_B)|$ \\

        \midrule

        \multirow{7}{*}{Training}
            & Soft-target mixing $\beta$ & 0.5 \\
            & Temperature $\tau$ & Learnable \\
            & Dense-loss noise $\epsilon$ & Gaussian noise for regularization \\
            & Optimizer & AdamW \\
            & Optimizer args & lr = 5e-3,\; weight\_decay = 1e-3 \\
            & LR scheduler & OneCycleLR \\
            & Scheduler steps & total\_steps = 2e+5 \\

        \midrule

        \multirow{4}{*}{3D Lifting}
            & Backbone 3D model & VGGT (pretrained) \\
            & Patch grid & $60 \times 60$ \\
            & Interpolation & Bilinear interpolation of 3D maps \\
            & Intra-structure & Distance matrices $D_A$ and $D_B$ from 3D points \\

        \bottomrule
    \end{tabular}
    \end{adjustbox}
    \caption{\textbf{Hyperparameters used in the Shape-of-You (SoY) framework.}
    Values reflect the unified configuration used across all experiments.}
    \vspace{-3mm}
    \label{tab:soy_parameters}
\end{table}

We summarize all hyperparameters in \cref{tab:soy_parameters}. 
Semantic UOT uses a KL penalty $\rho=0.75$, entropy regularization $\varepsilon=1$, uniform patch masses, and a cosine-based cost $C_{\text{sem}}$.
For FGW fusion, we use the corrected semantic–geometric balance of $\alpha=0.3$, and normalize both semantic and geometric costs before combining them.

Anchor-based refinement runs for $T=5$ iterations with $K=64$ mutual anchors per iteration.  
The cycle-consistency tolerance $\delta$ is determined by a data-driven quantile threshold $q=0.01$ over the 3D cycle-error distribution and serves mainly to filter out clear outliers.  
Final anchors are ranked by a combined score favoring both high transport confidence and low geometric distortion, making the method robust to the exact choice of $q$.

Training uses soft-target mixing $\beta=0.5$, a learnable temperature~$\tau$, Gaussian noise in the dense loss, and an AdamW optimizer with a OneCycleLR schedule as specified in \cref{tab:soy_parameters}.  
For 3D lifting, we employ a pretrained VGGT backbone, lift images to a $60\times60$ grid via bilinear interpolation, and construct intra-structure distance matrices $D_A$ and $D_B$ from the resulting 3D points for use in the FGW term.

\section{Algorithm}
\label{sec:algorithm}
To make our pseudo-label generator easy to reproduce, we provide PyTorch-style pseudocode for the full FGW pipeline in Alg.~\ref{alg:pseudo-label}. 
The code explicitly shows (i) the initial semantic UOT matching, (ii) the construction of 3D distance matrices, (iii) the anchor-based linearization of the FGW structural term, and (iv) the iterative re-solving of unbalanced OT with the fused semantic–geometric cost.
This low-level view complements the high-level description in the main paper and clarifies how each component of SoY is implemented in practice.

\section{Additional visualization}
\label{sec:visualization}
\begin{figure*}[htbp]
  \centering
  \begin{tabular}{ccc}
    \multicolumn{1}{c}{DistillDIFT~\cite{fundel2025distillation}} &
    \multicolumn{1}{c}{DINOv2+SD~\cite{{zhang2023tale}}} &
    \multicolumn{1}{c}{Ours} \\[5pt]
    \includegraphics[width=0.3\textwidth]{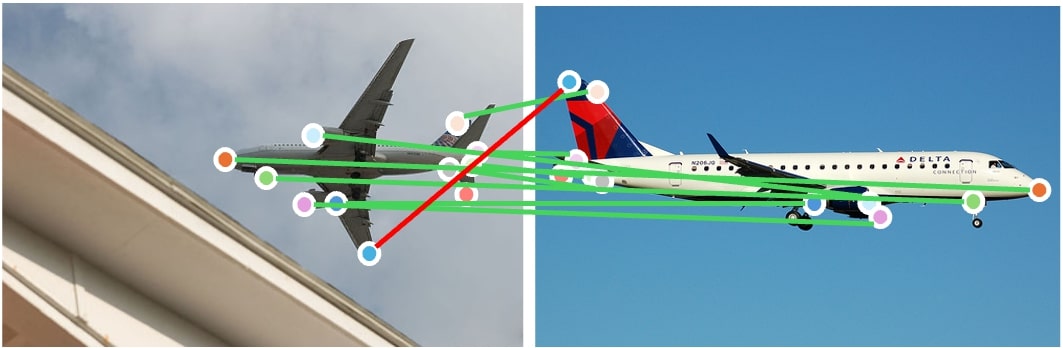} &
    \includegraphics[width=0.3\textwidth]{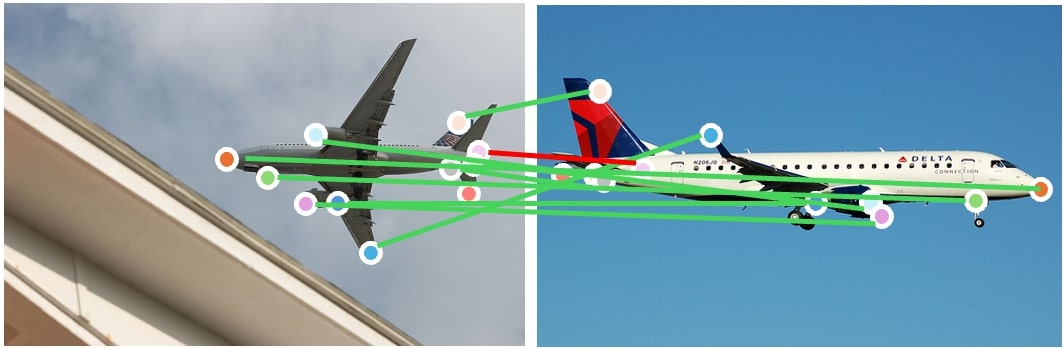} &
    \includegraphics[width=0.3\textwidth]{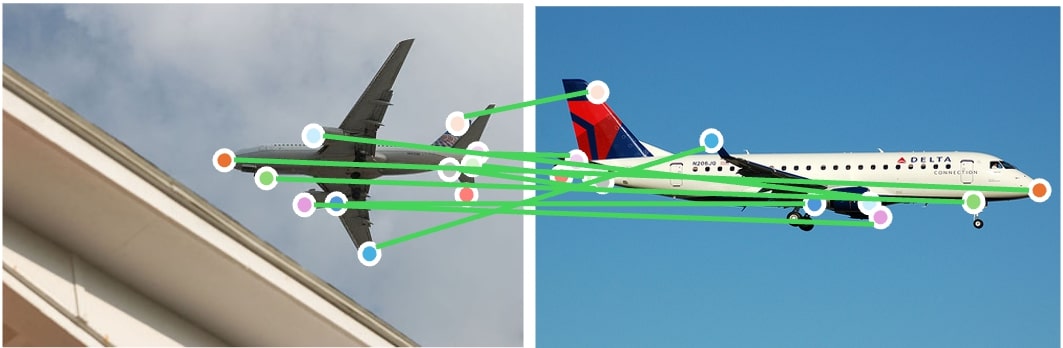} \\[0.25mm]

    \includegraphics[width=0.3\textwidth]{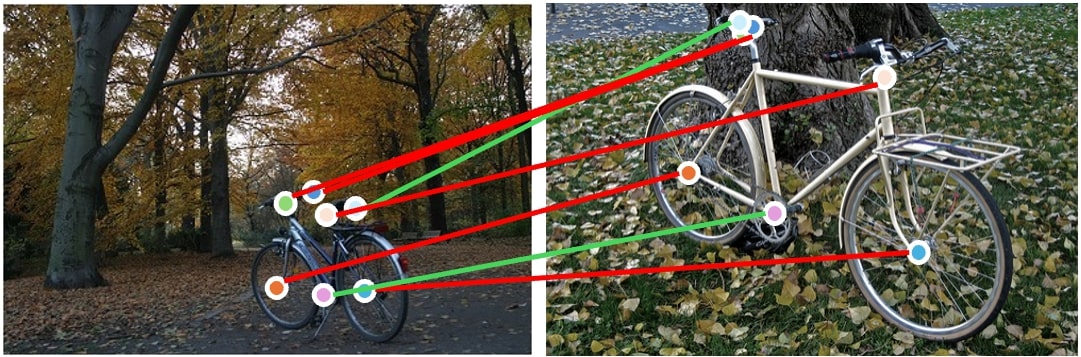} &
    \includegraphics[width=0.3\textwidth]{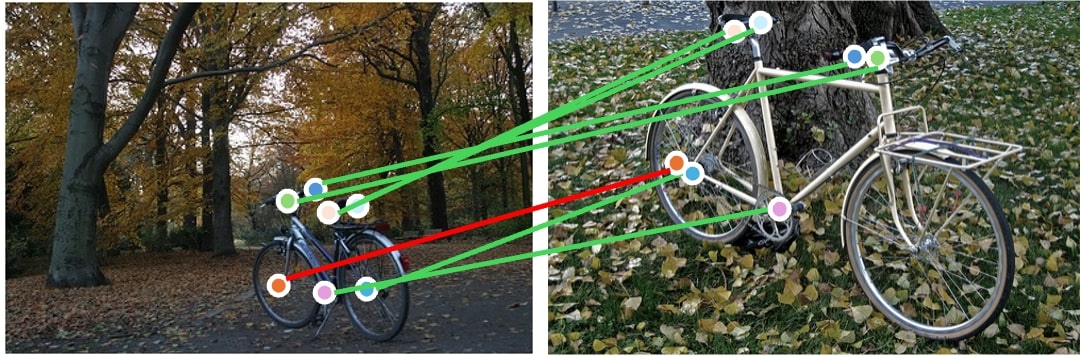} &
    \includegraphics[width=0.3\textwidth]{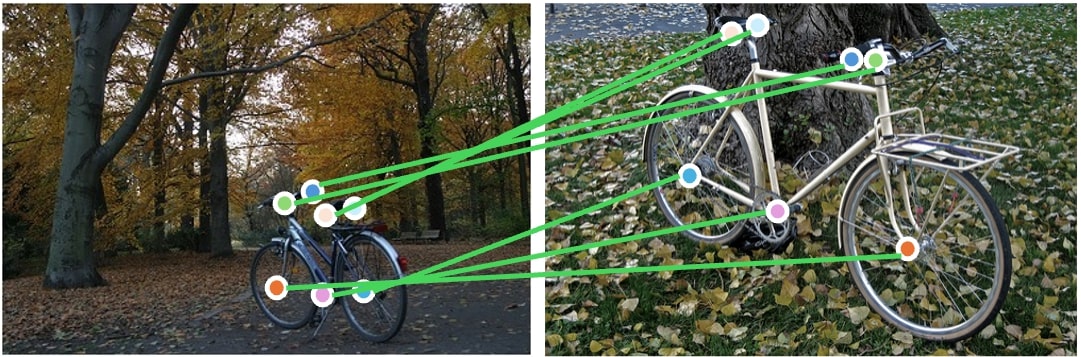} \\[0.25mm]

    \includegraphics[width=0.3\textwidth]{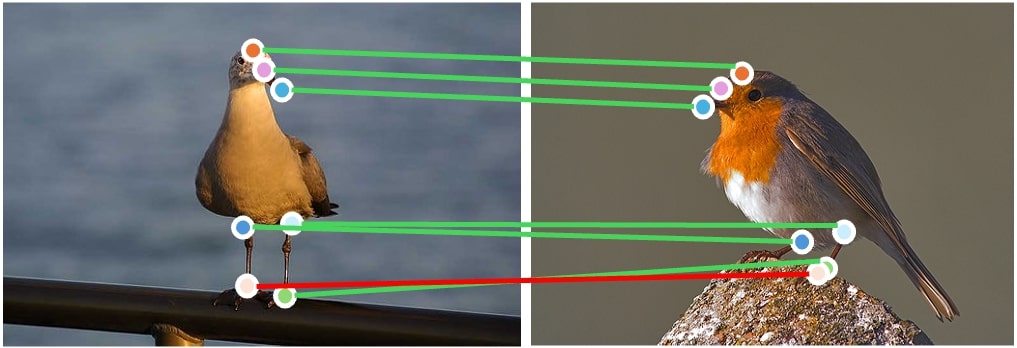} &
    \includegraphics[width=0.3\textwidth]{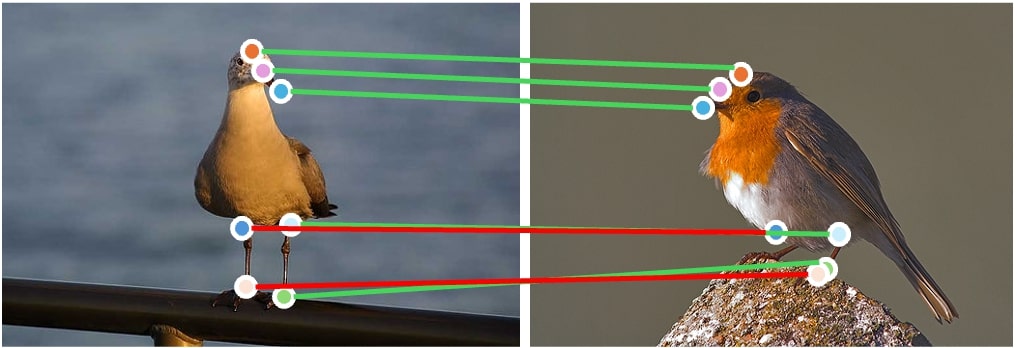} &
    \includegraphics[width=0.3\textwidth]{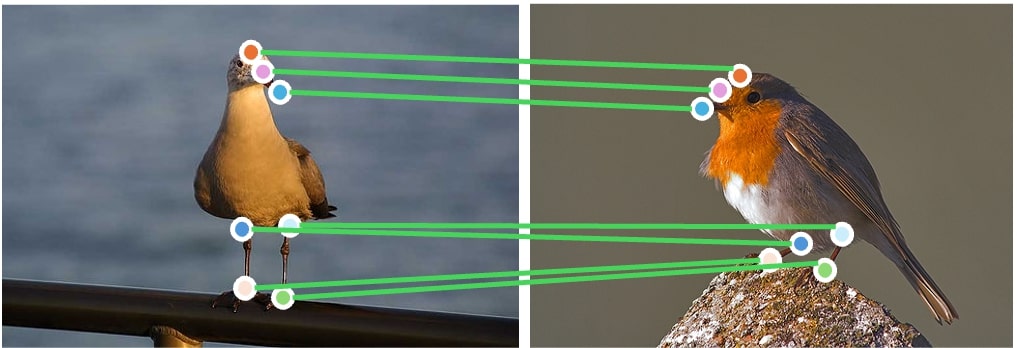} \\[0.25mm]

    \includegraphics[width=0.3\textwidth]{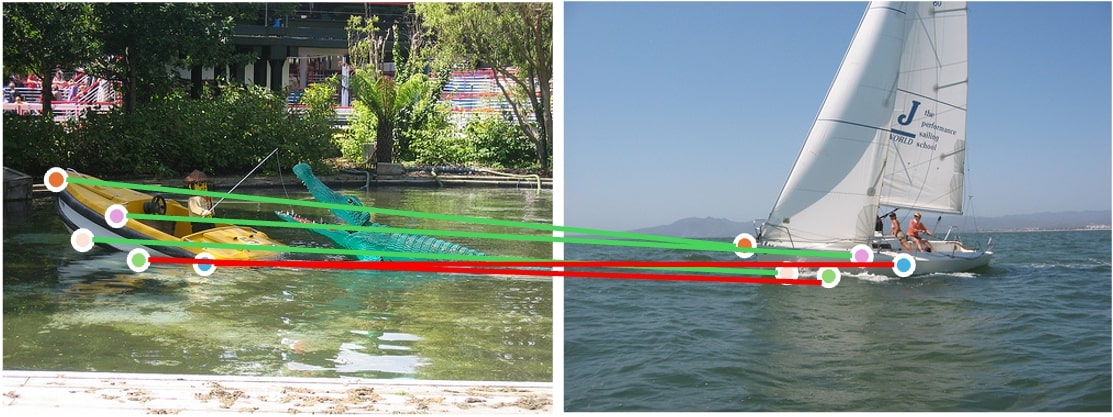} &
    \includegraphics[width=0.3\textwidth]{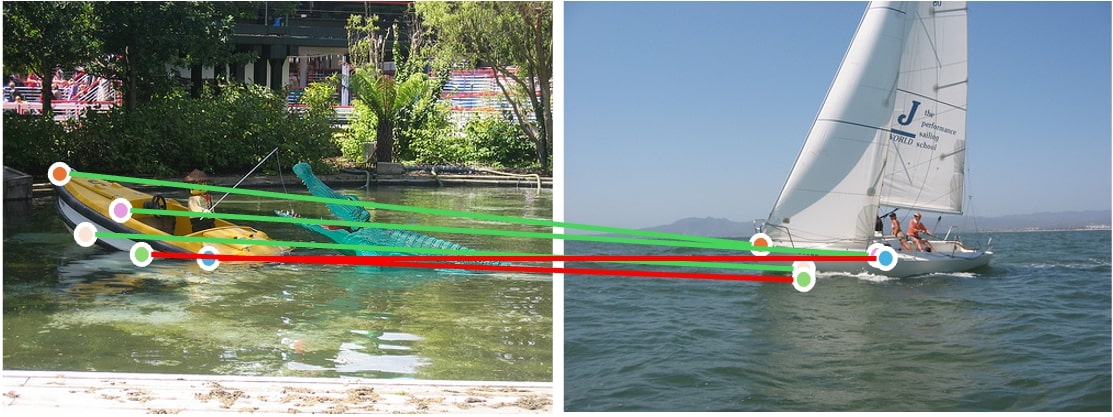} &
    \includegraphics[width=0.3\textwidth]{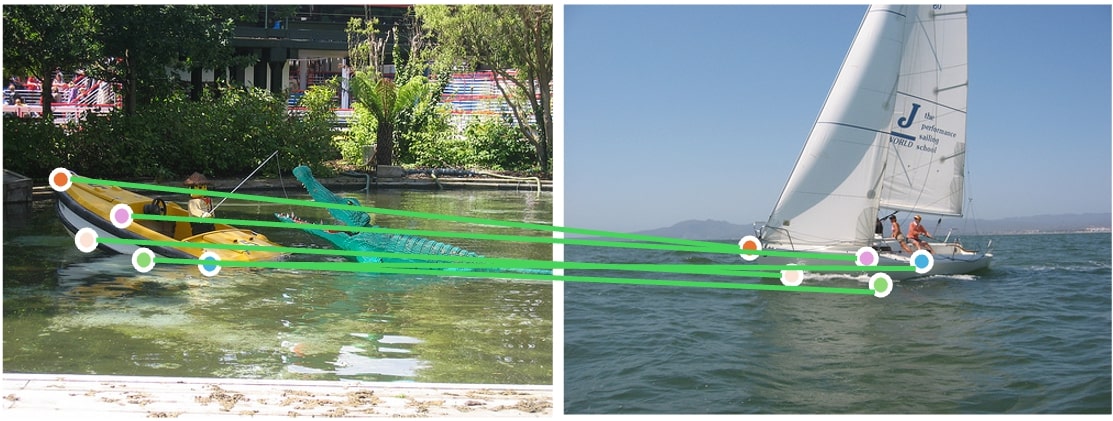} \\[0.25mm]

    \includegraphics[width=0.3\textwidth]{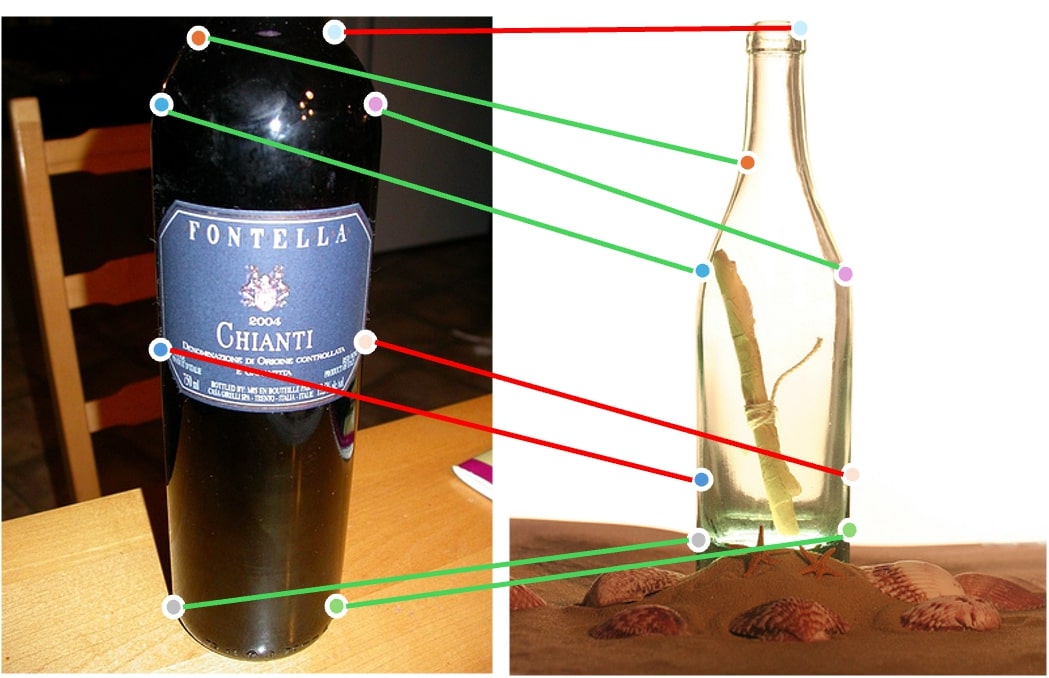} &
    \includegraphics[width=0.3\textwidth]{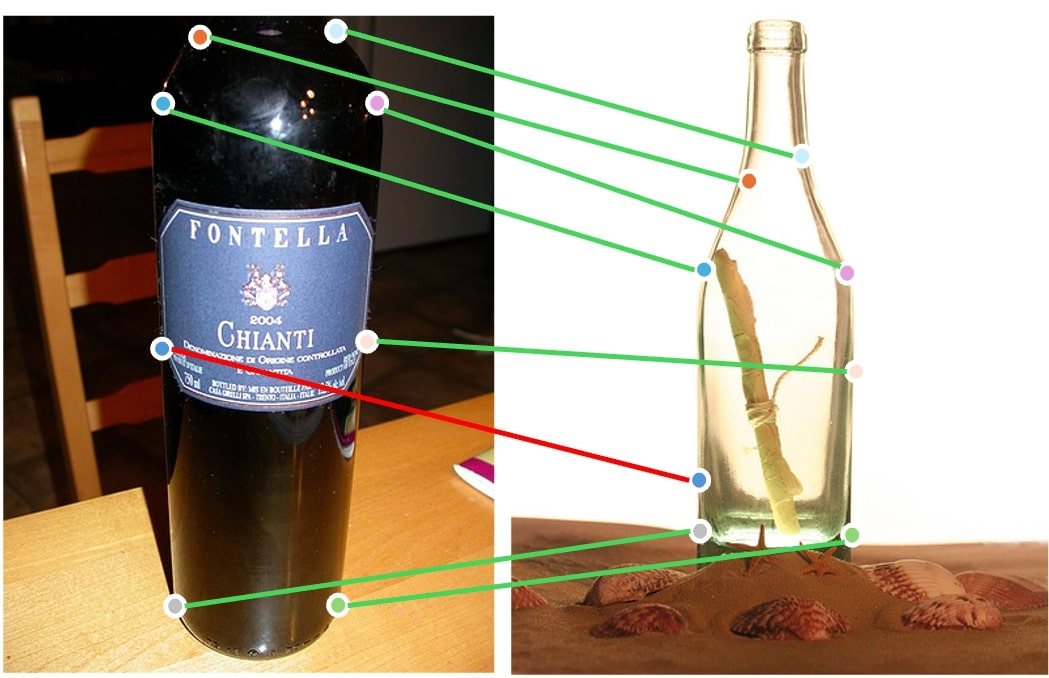} &
    \includegraphics[width=0.3\textwidth]{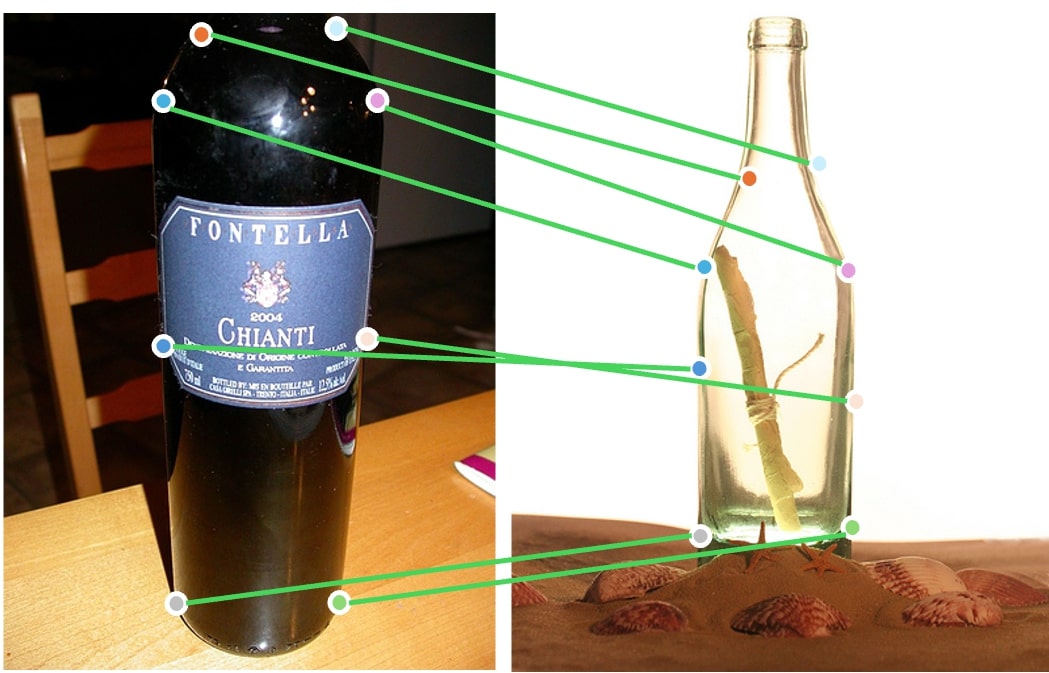} \\[0.25mm]

    \includegraphics[width=0.3\textwidth]{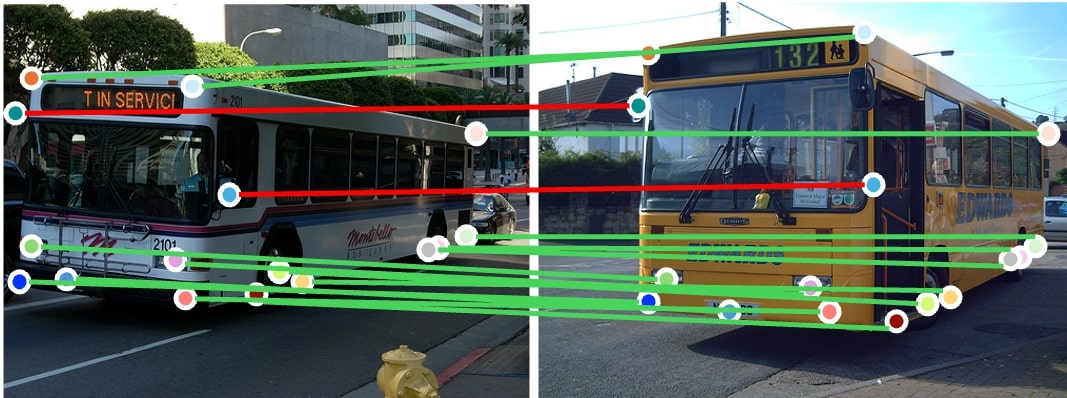} &
    \includegraphics[width=0.3\textwidth]{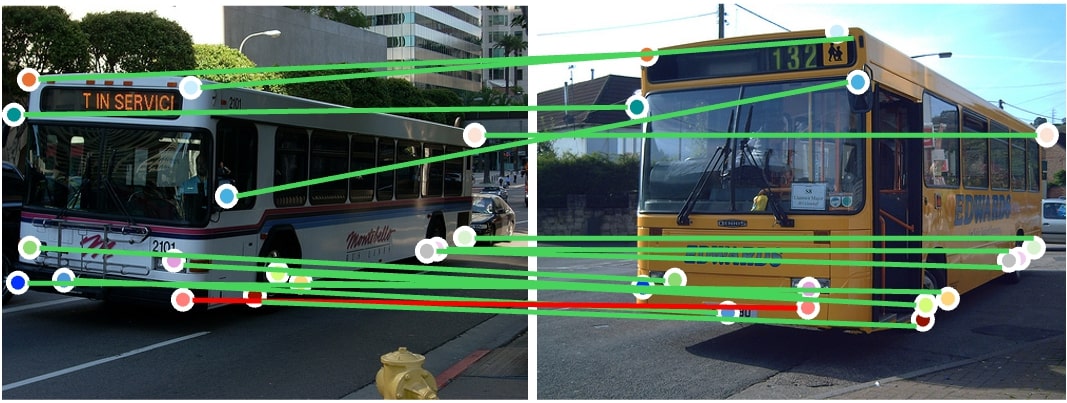} &
    \includegraphics[width=0.3\textwidth]{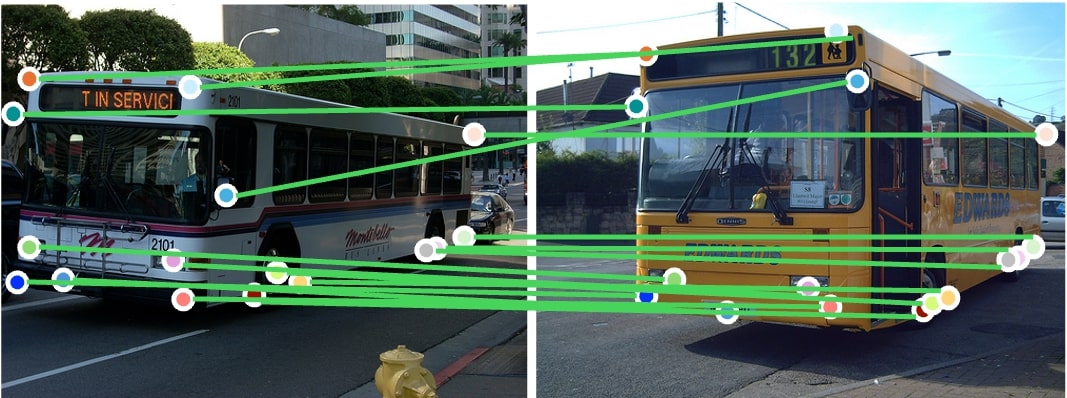} \\[0.25mm]

    \includegraphics[width=0.3\textwidth]{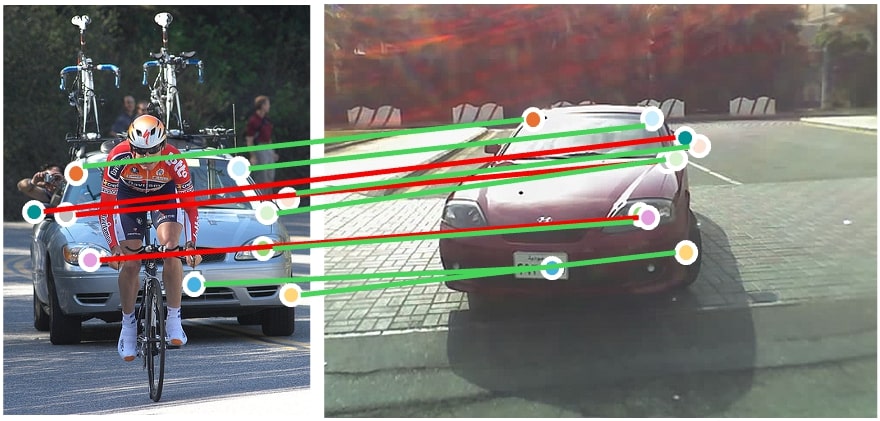} &
    \includegraphics[width=0.3\textwidth]{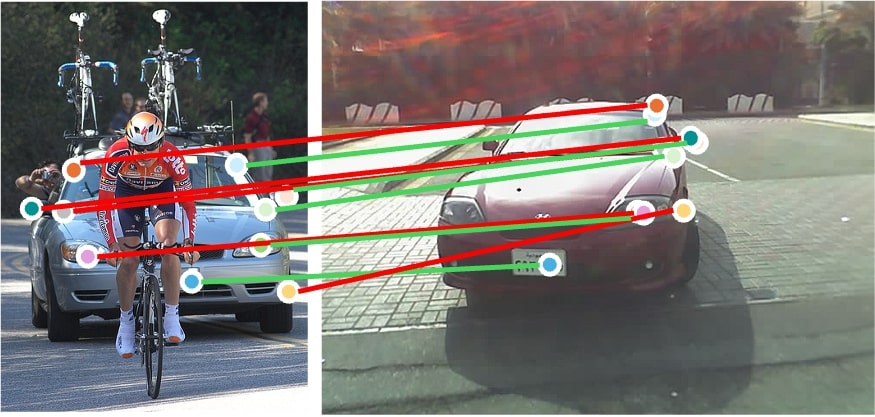} &
    \includegraphics[width=0.3\textwidth]{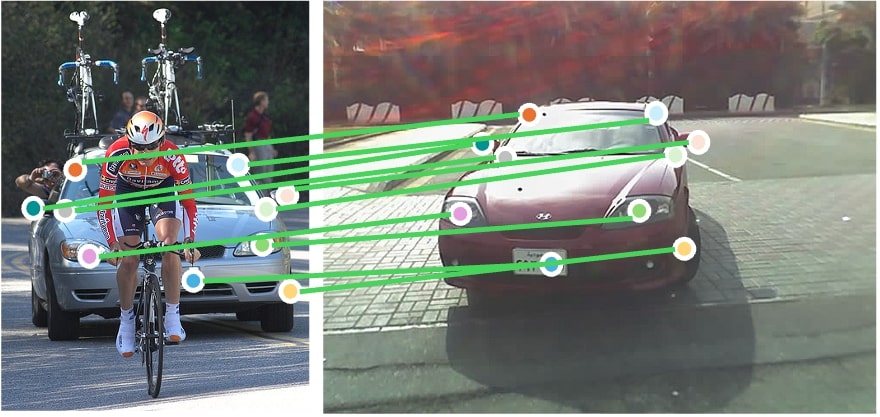} \\[0.25mm]

    \includegraphics[width=0.3\textwidth]{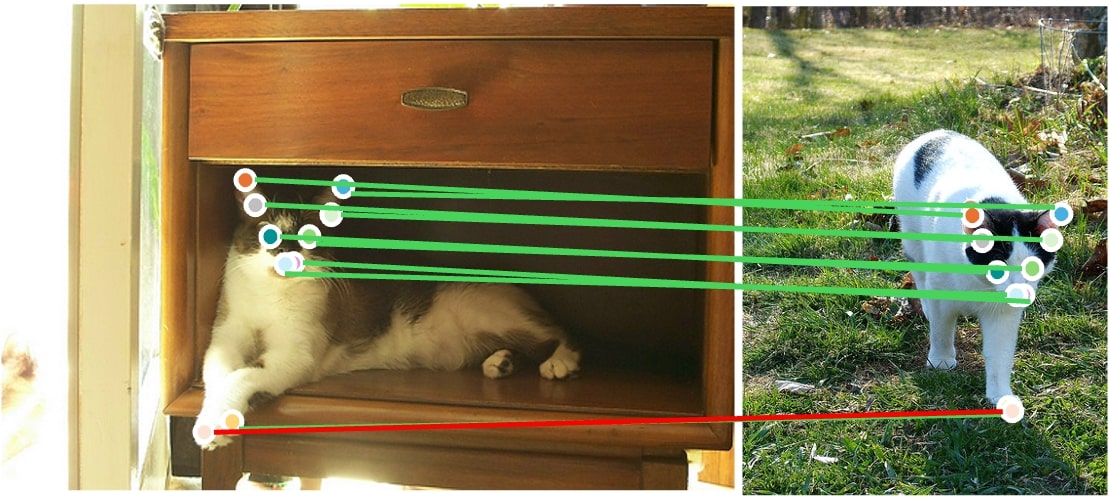} &
    \includegraphics[width=0.3\textwidth]{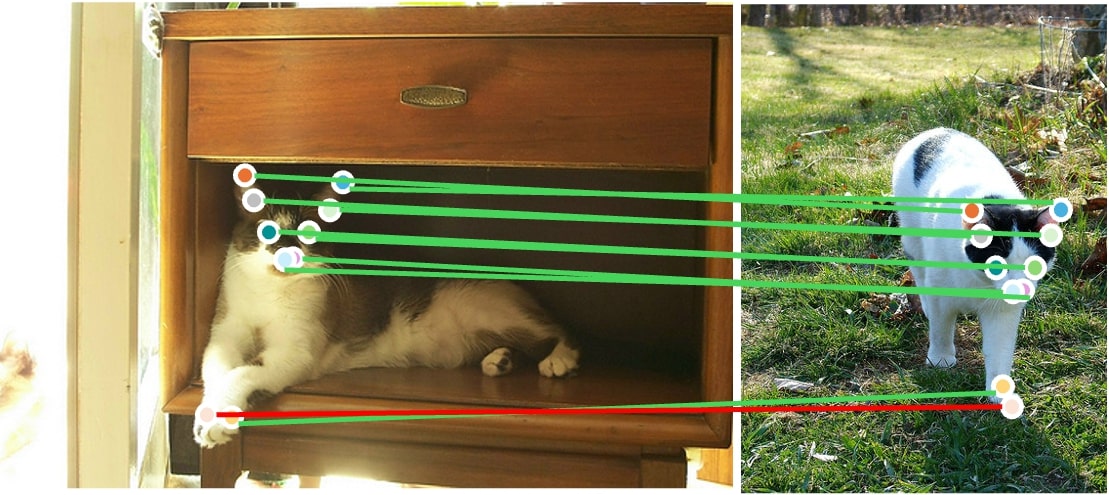} &
    \includegraphics[width=0.3\textwidth]{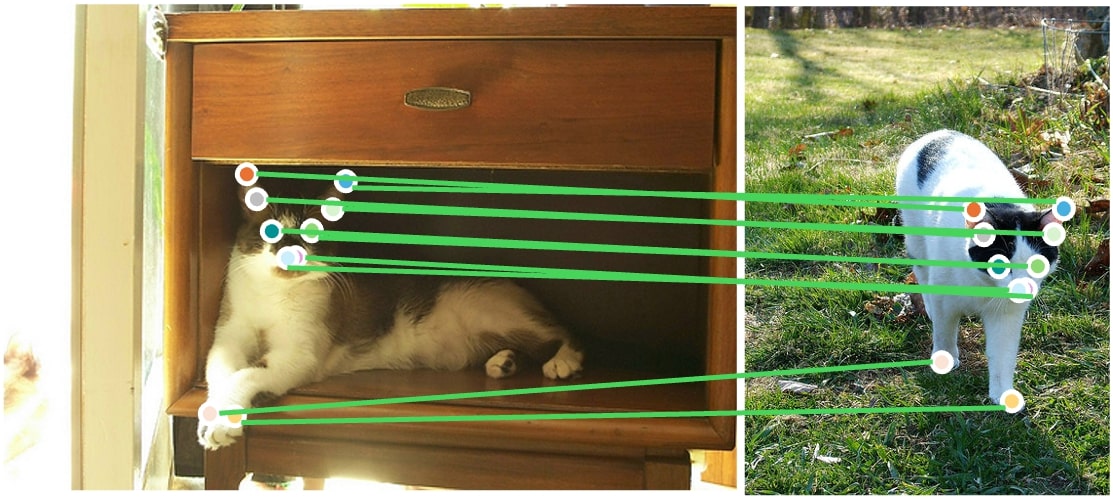} \\[0.25mm]

    \includegraphics[width=0.3\textwidth]{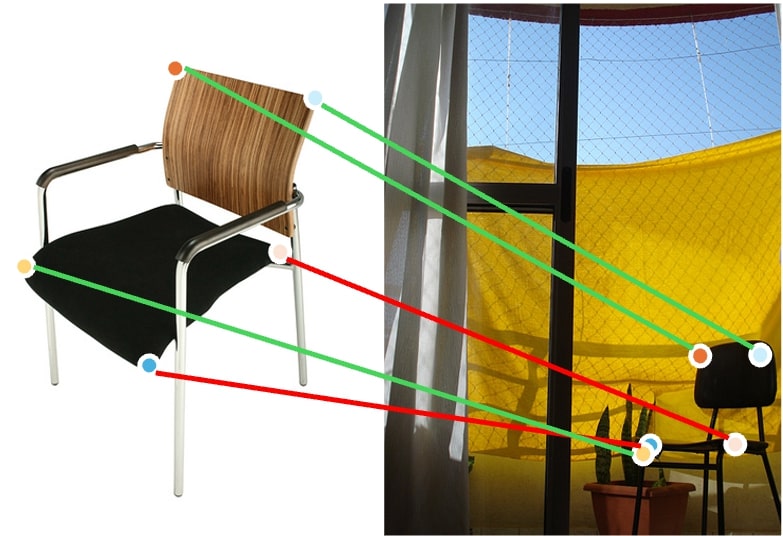} &
    \includegraphics[width=0.3\textwidth]{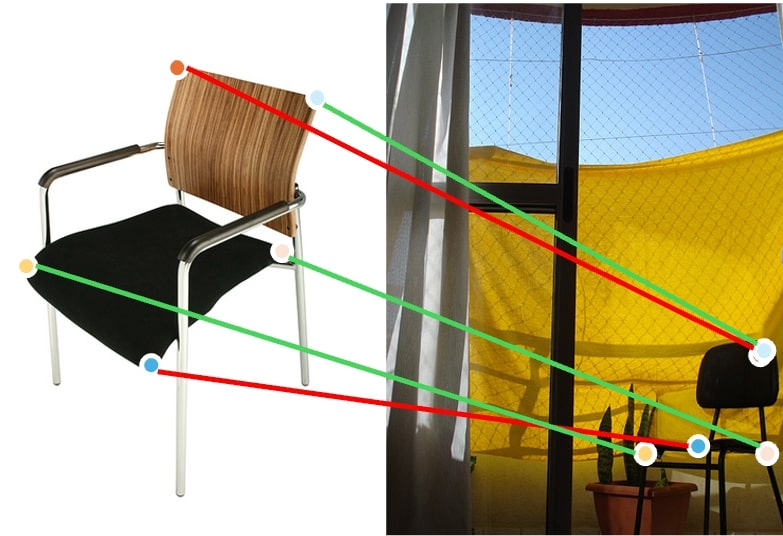} &
    \includegraphics[width=0.3\textwidth]{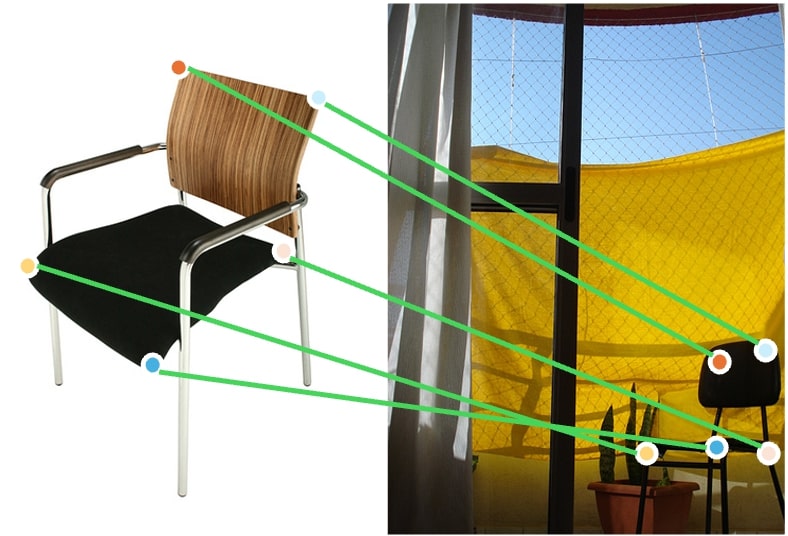} \\
  \end{tabular}

  \caption{Visual comparison of semantic correspondences on SPair-71k~\cite{min2019spair} across DistillDIFT~\cite{fundel2025distillation}, DINOv2+SD~\cite{zhang2023tale}, and our approach. Correct and incorrect matches are indicated by {\color{green}green lines} and {\color{red}red lines}, respectively.}
  \label{fig:uncurated_matches_1}
\end{figure*}

\begin{figure*}[htbp]
  \centering
  \begin{tabular}{ccc}
    \multicolumn{1}{c}{DistillDIFT~\cite{fundel2025distillation}} &
    \multicolumn{1}{c}{DINOv2+SD~\cite{zhang2023tale}} &
    \multicolumn{1}{c}{Ours} \\[5pt]
    \includegraphics[width=0.3\textwidth]{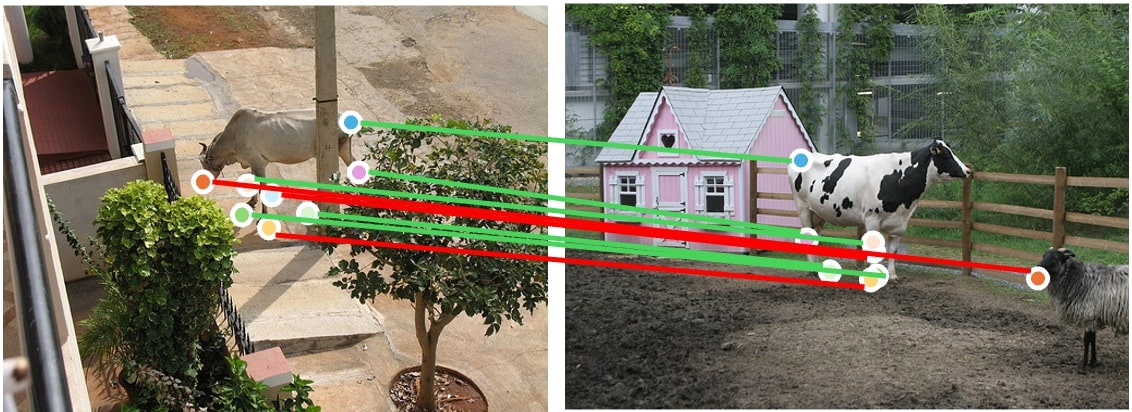} &
    \includegraphics[width=0.3\textwidth]{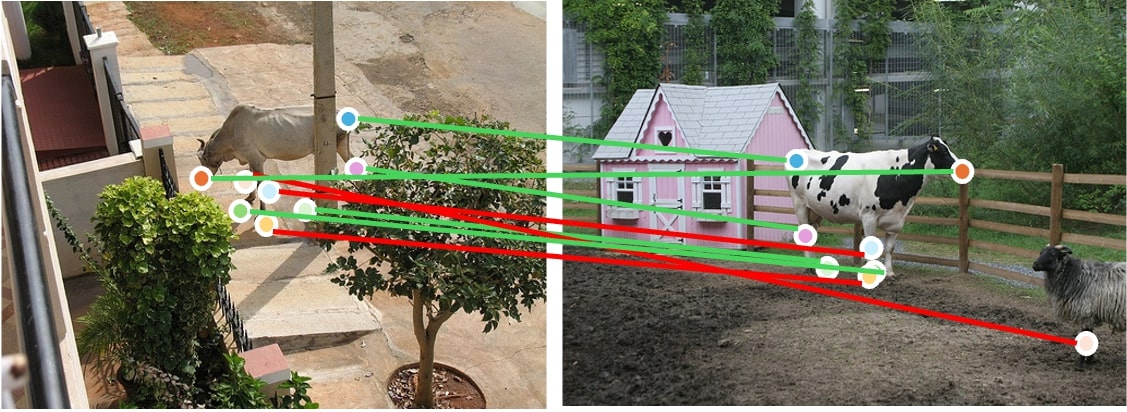} &
    \includegraphics[width=0.3\textwidth]{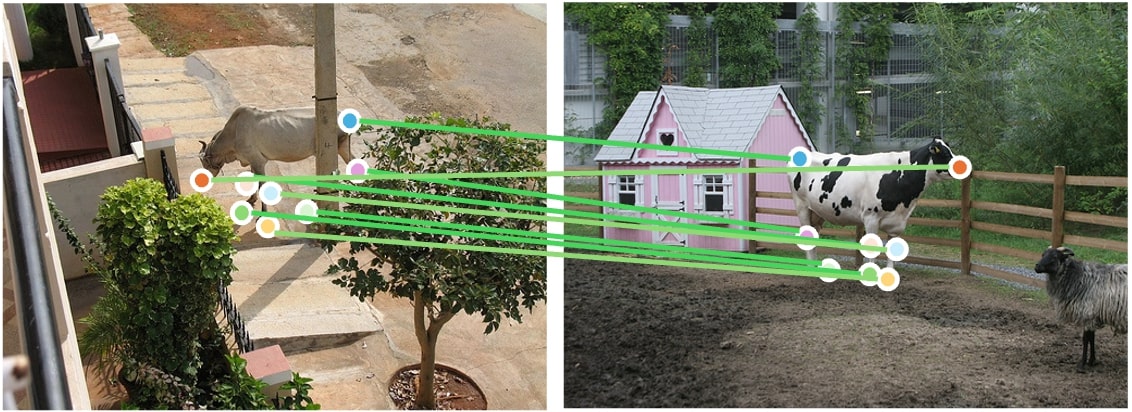} \\[2.5mm]

    \includegraphics[width=0.3\textwidth]{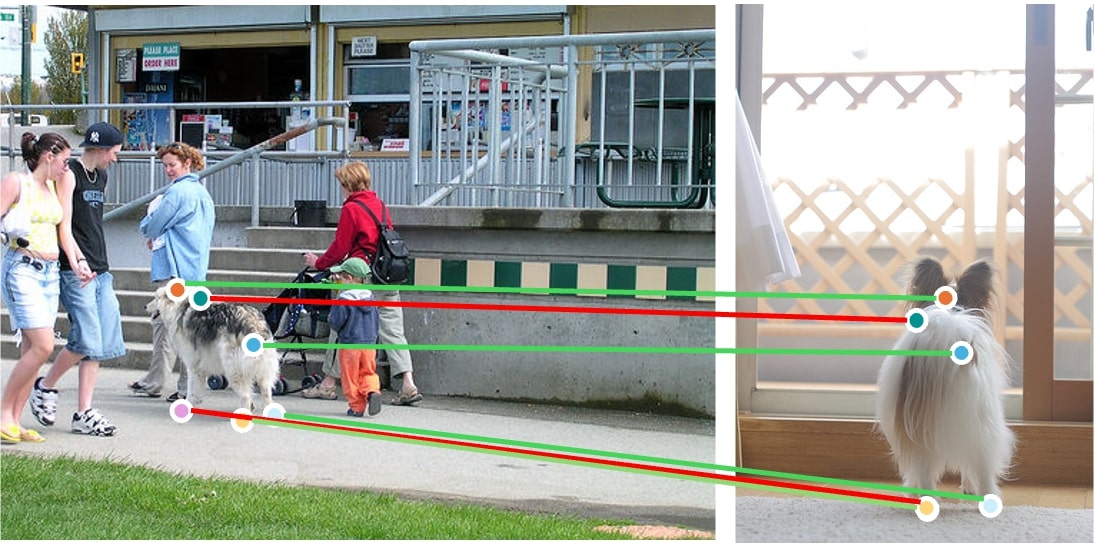} &
    \includegraphics[width=0.3\textwidth]{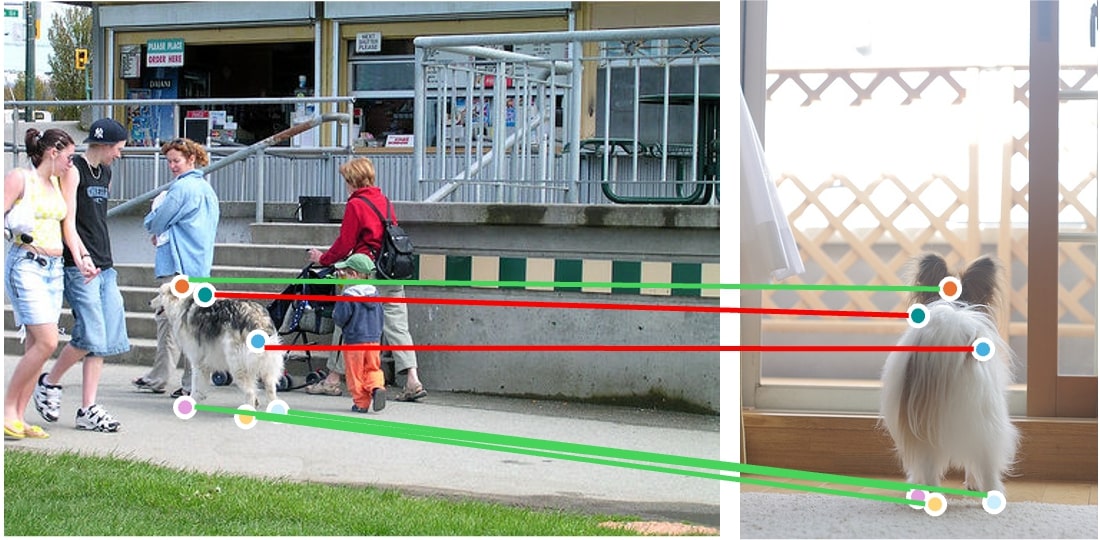} &
    \includegraphics[width=0.3\textwidth]{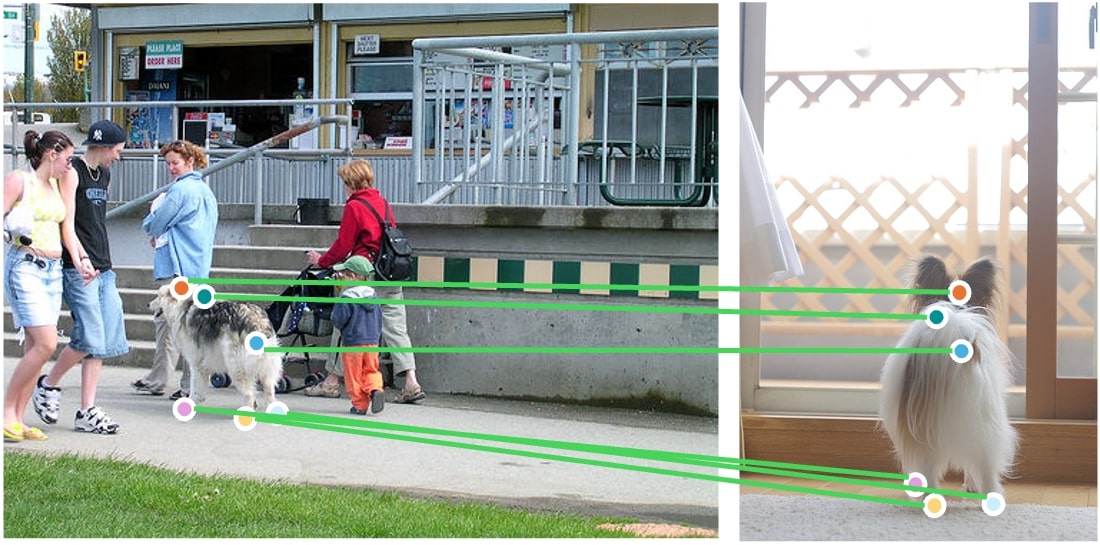} \\[2.5mm]

    \includegraphics[width=0.3\textwidth]{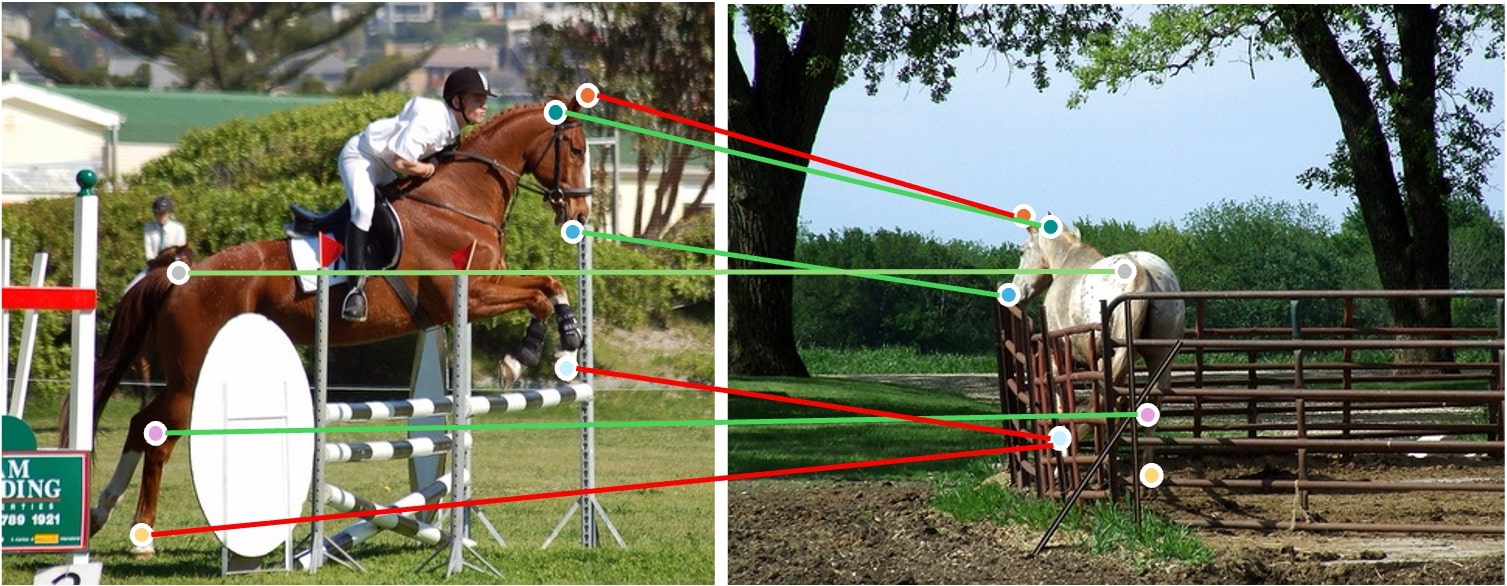} &
    \includegraphics[width=0.3\textwidth]{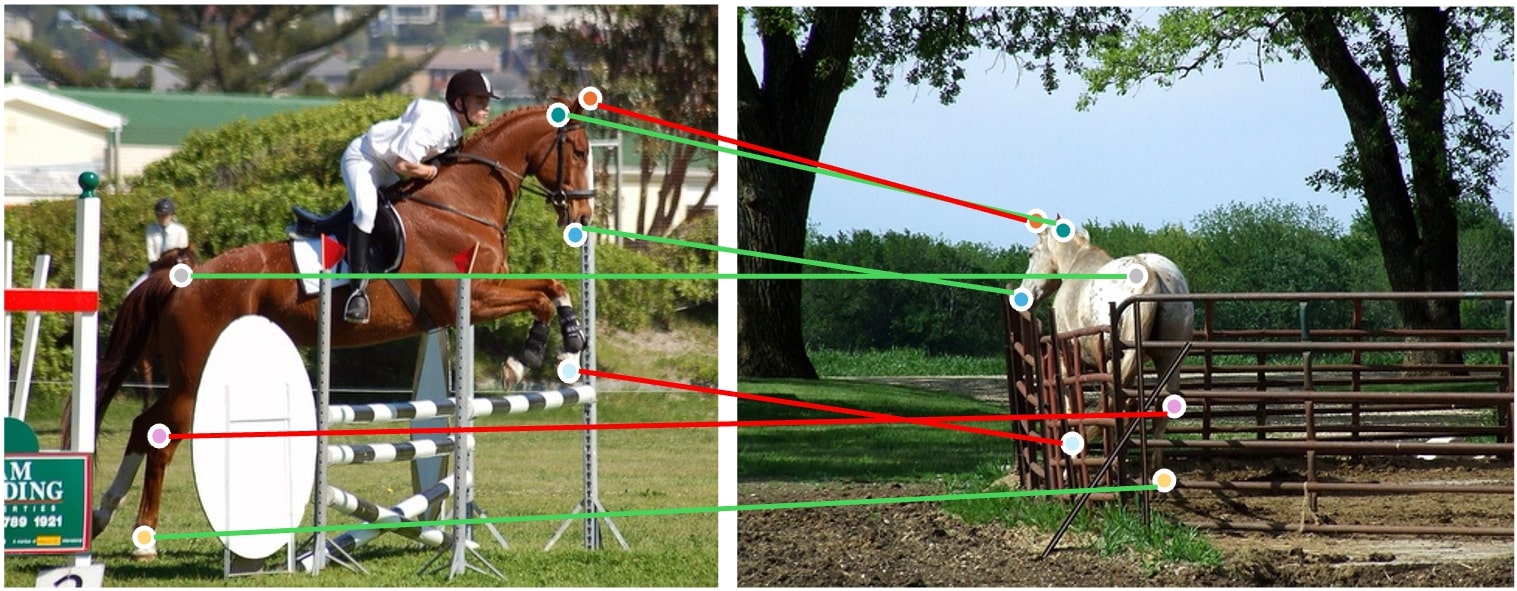} &
    \includegraphics[width=0.3\textwidth]{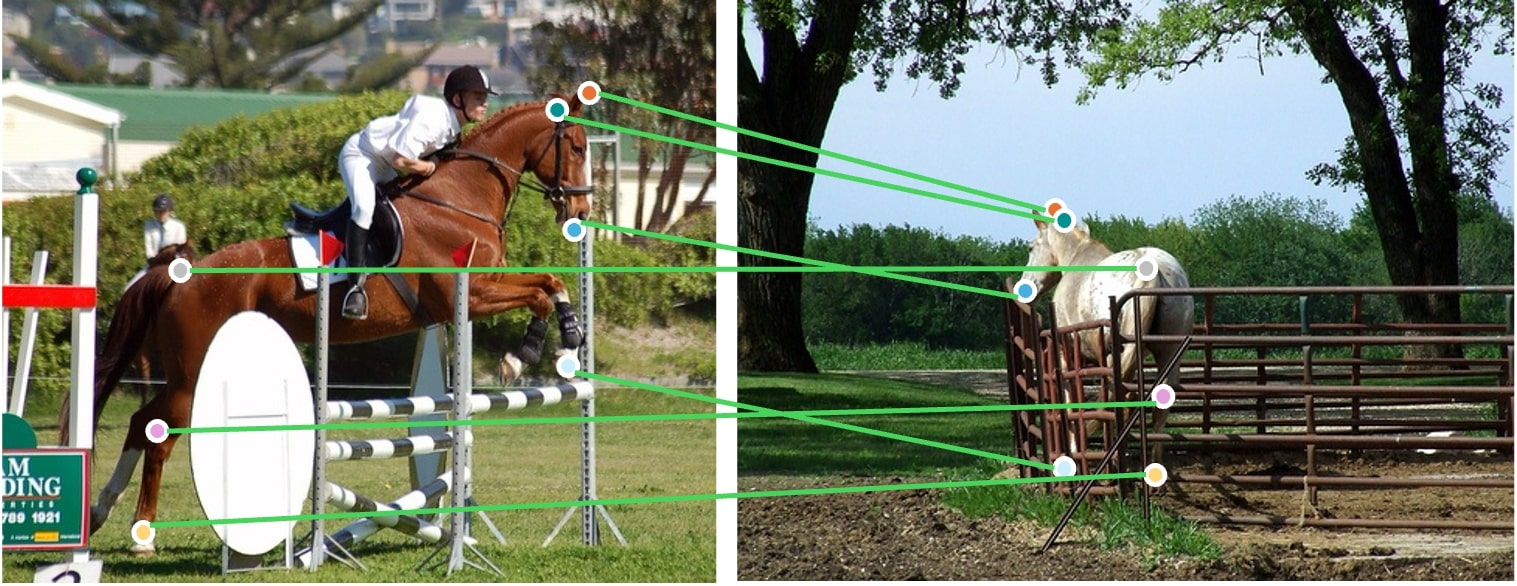} \\[2.5mm]

    \includegraphics[width=0.3\textwidth]{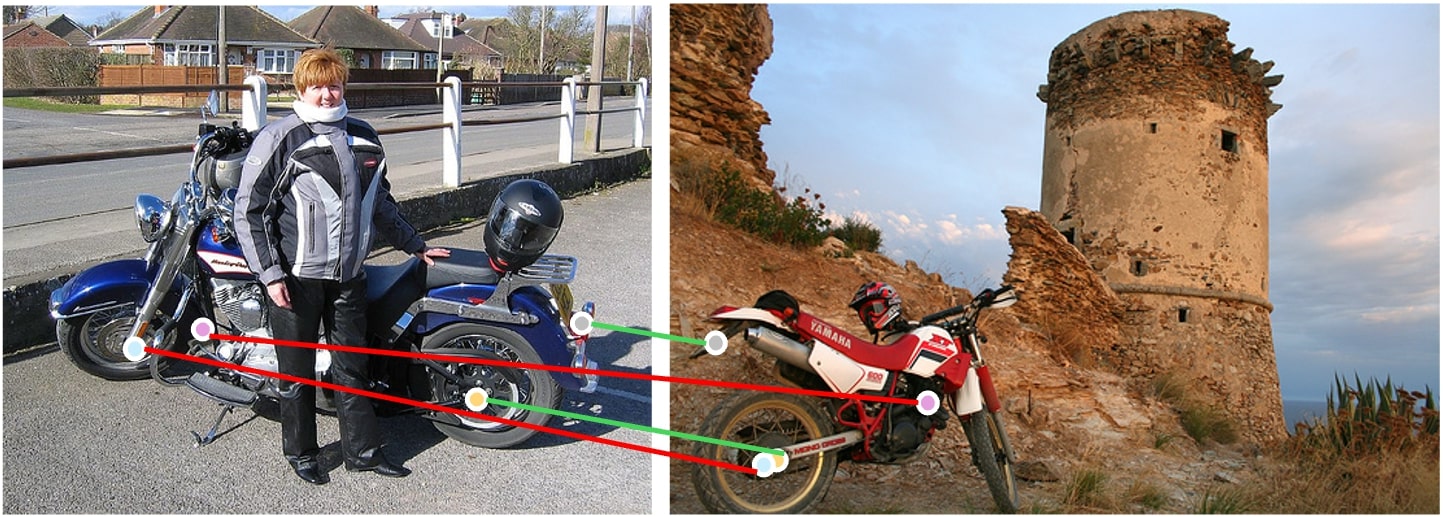} &
    \includegraphics[width=0.3\textwidth]{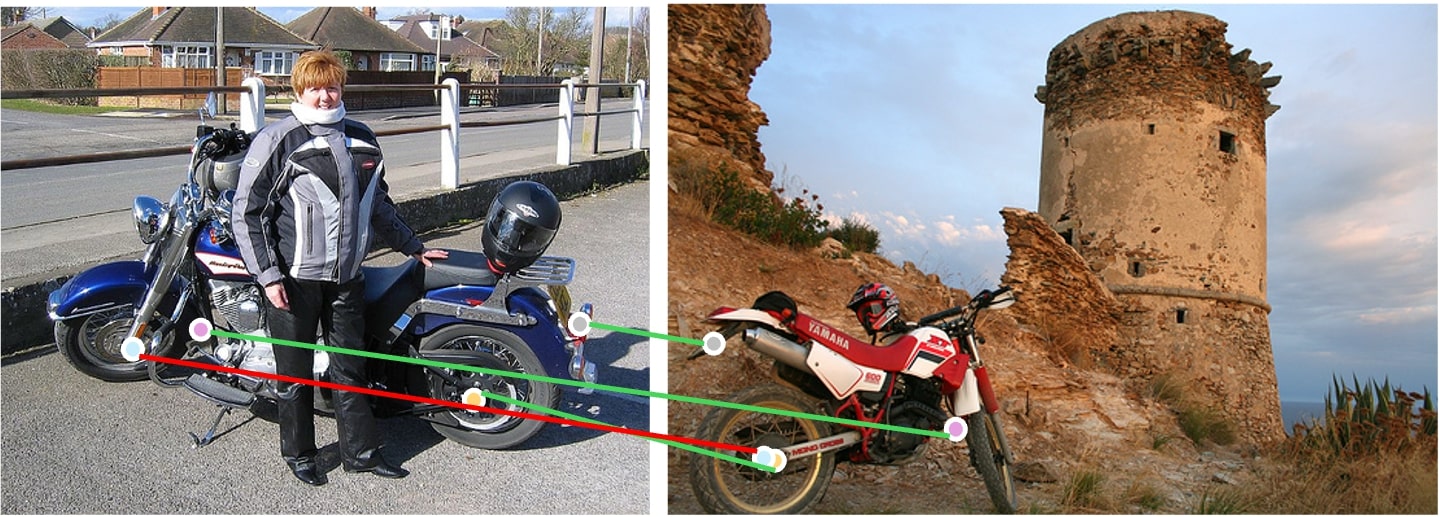} &
    \includegraphics[width=0.3\textwidth]{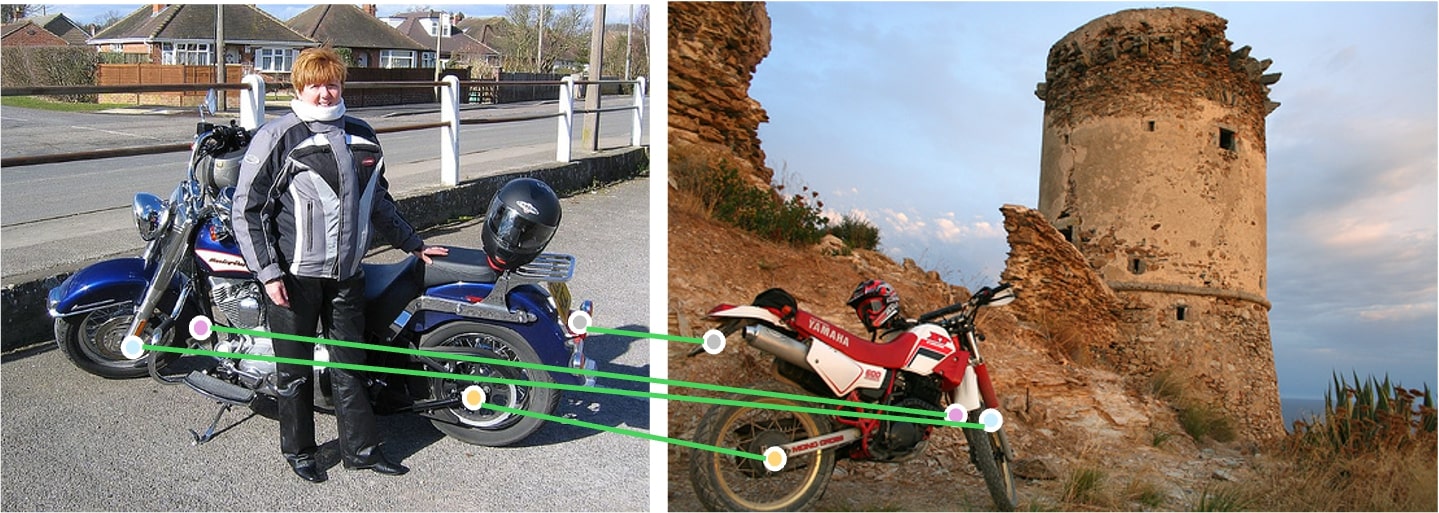} \\[2.5mm]

    \includegraphics[width=0.3\textwidth]{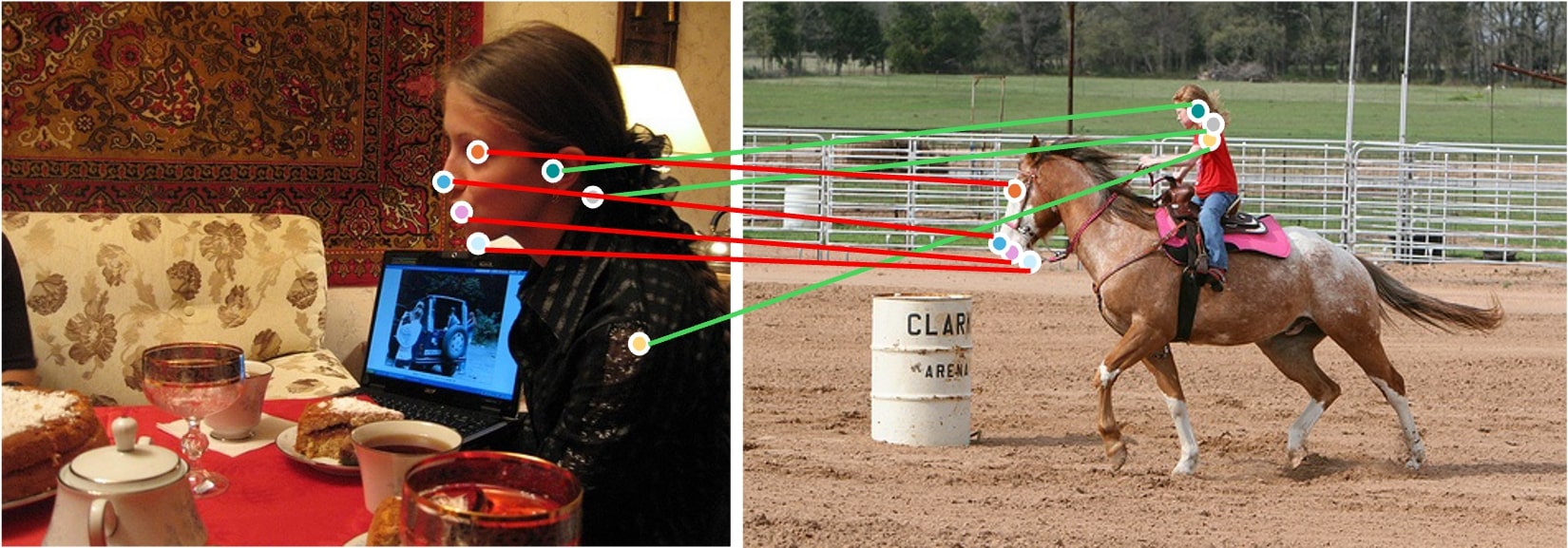} &
    \includegraphics[width=0.3\textwidth]{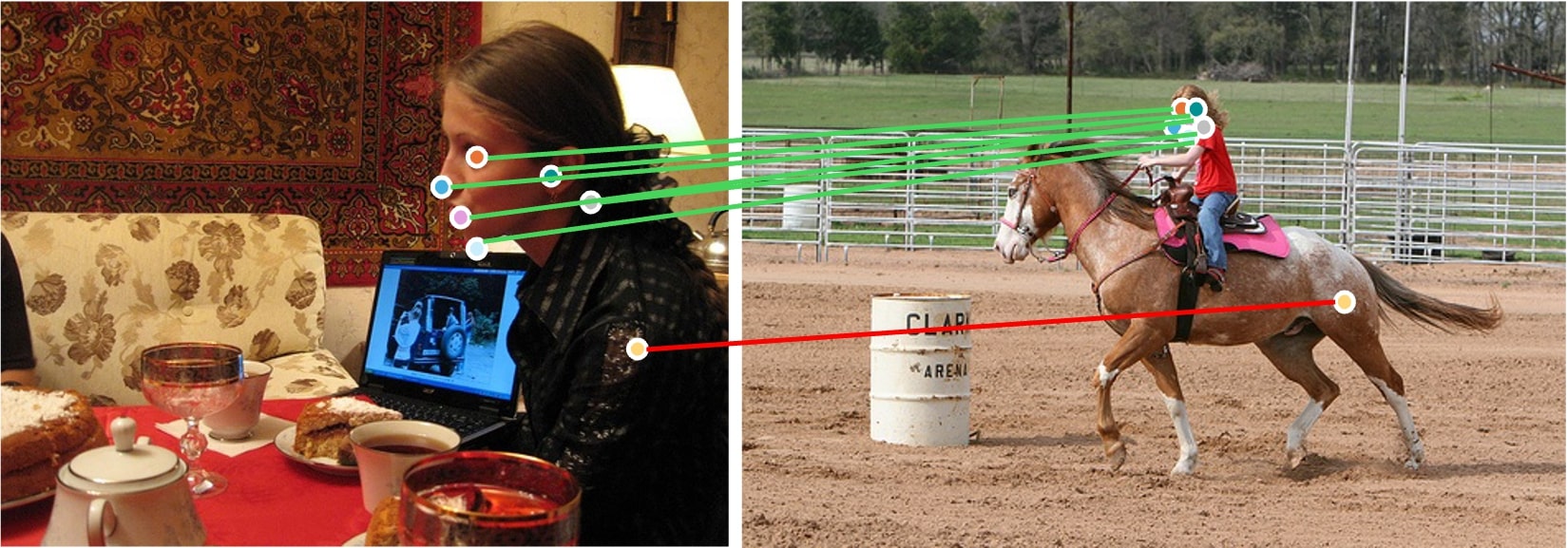} &
    \includegraphics[width=0.3\textwidth]{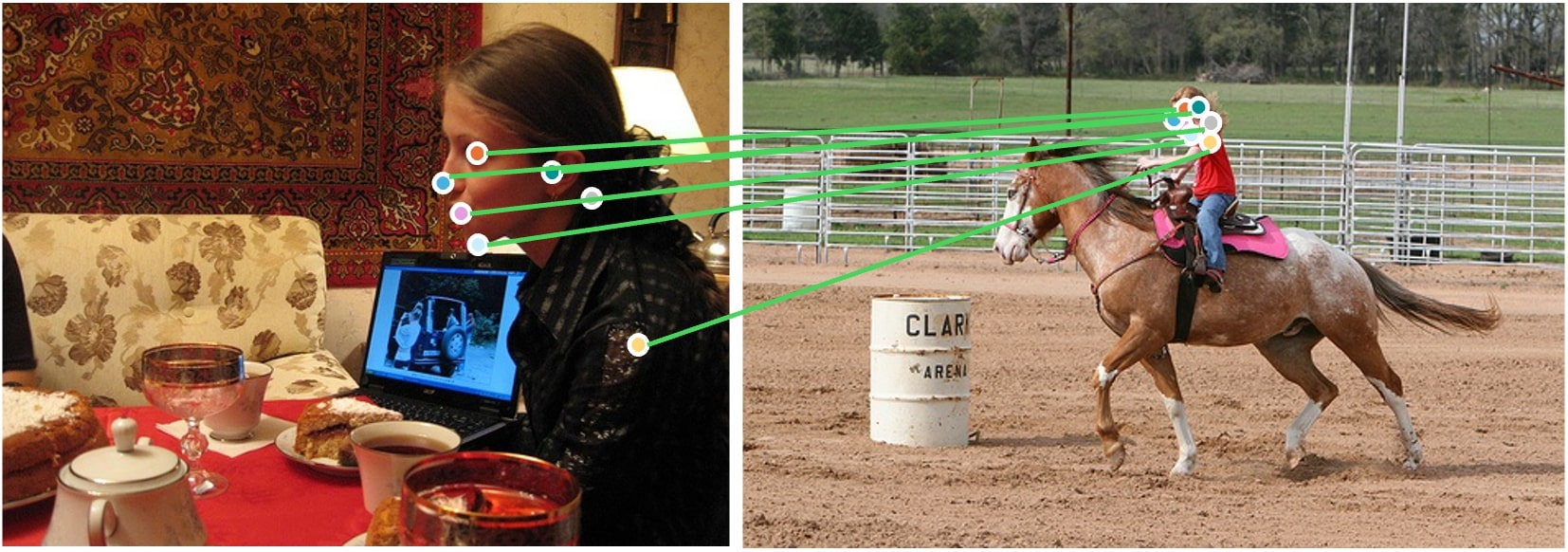} \\[2.5mm]

    \includegraphics[width=0.3\textwidth]{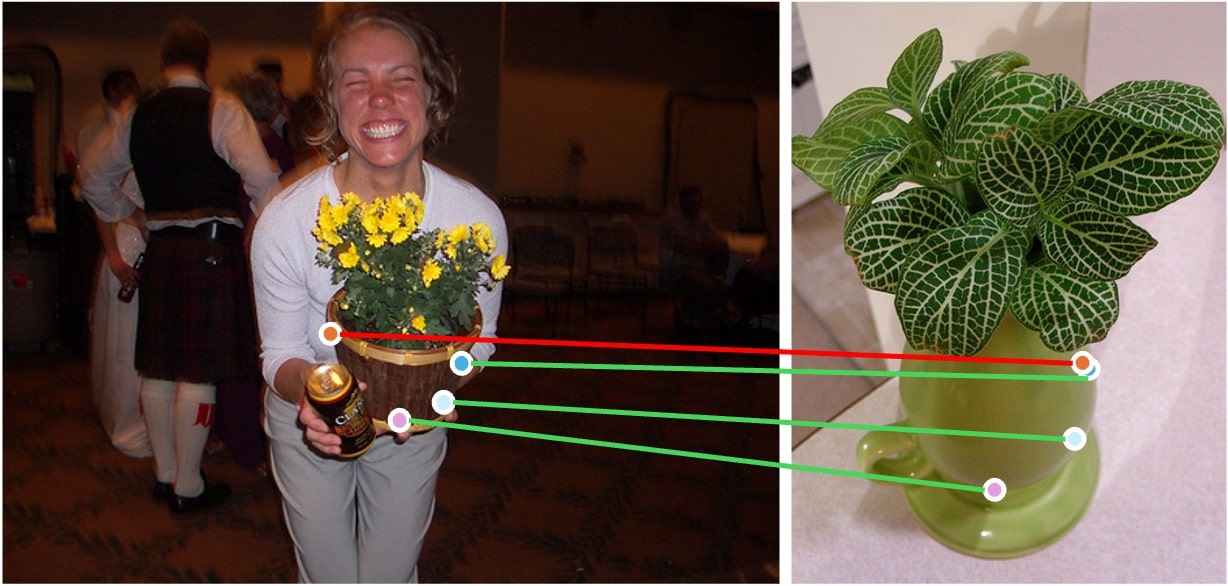} &
    \includegraphics[width=0.3\textwidth]{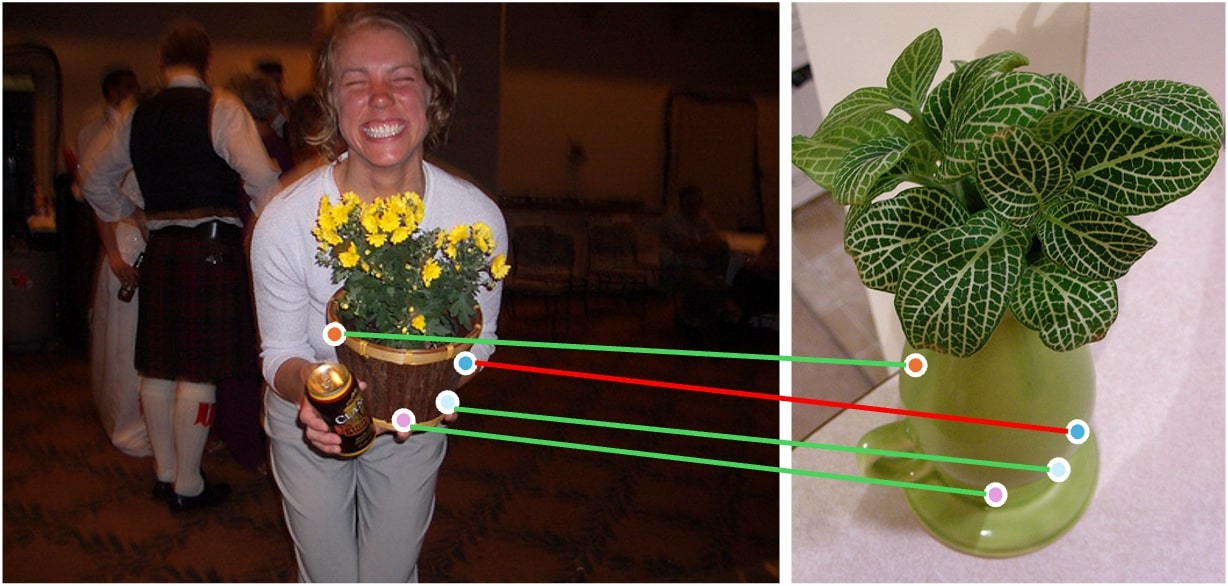} &
    \includegraphics[width=0.3\textwidth]{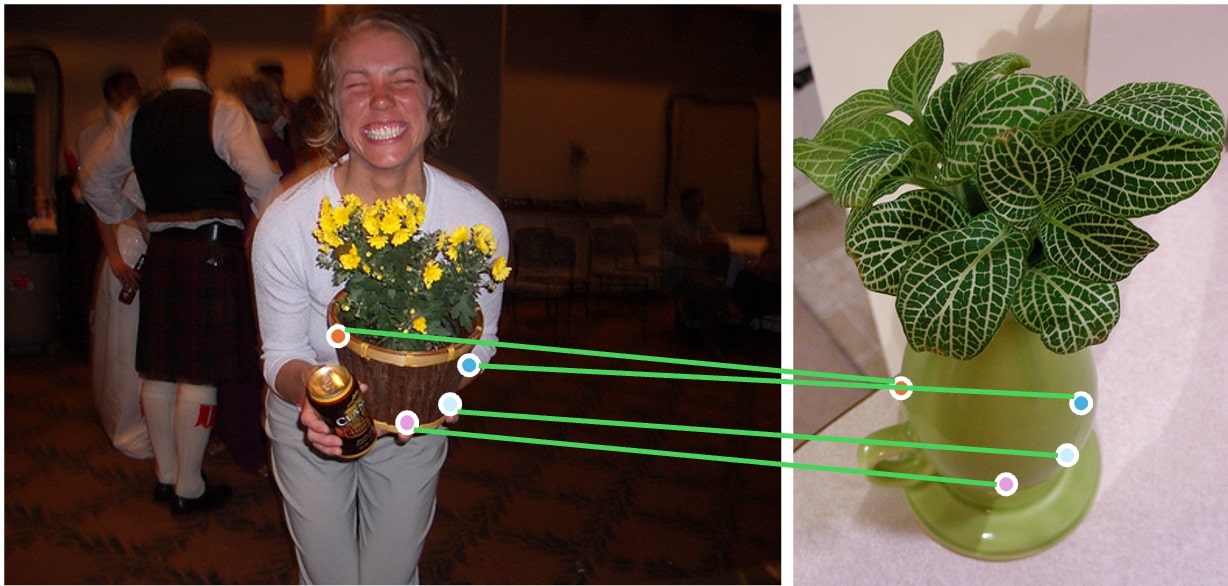} \\[2.5mm]

    \includegraphics[width=0.3\textwidth]{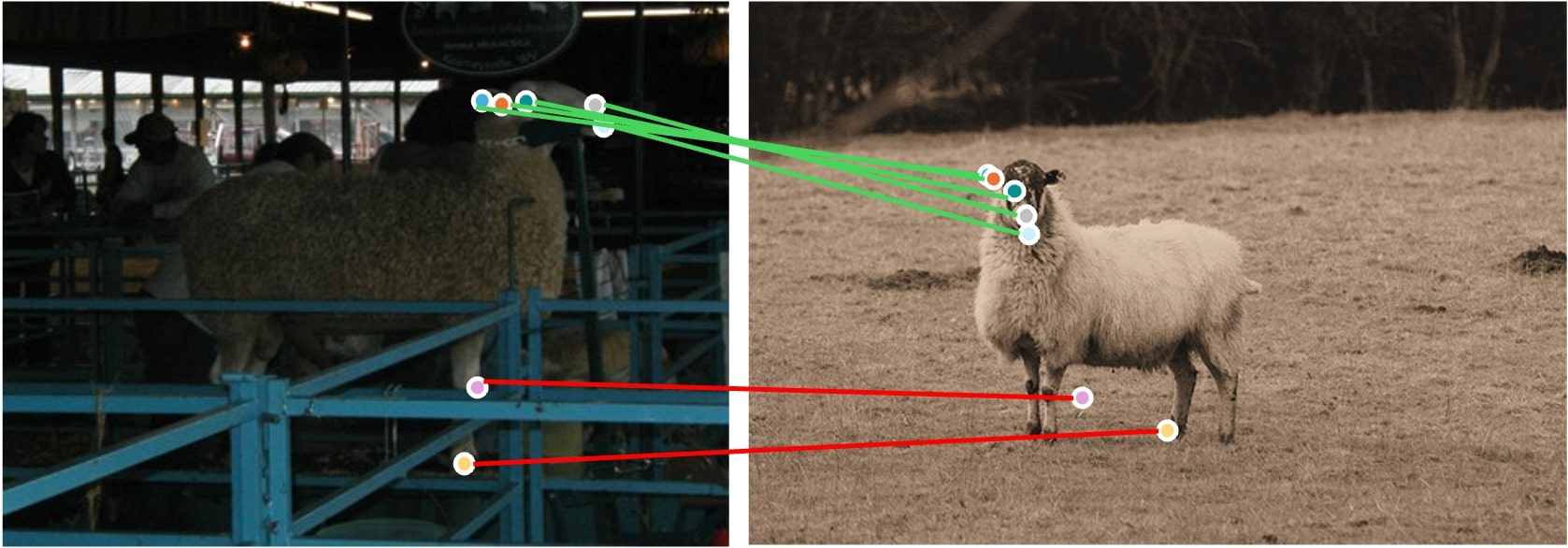} &
    \includegraphics[width=0.3\textwidth]{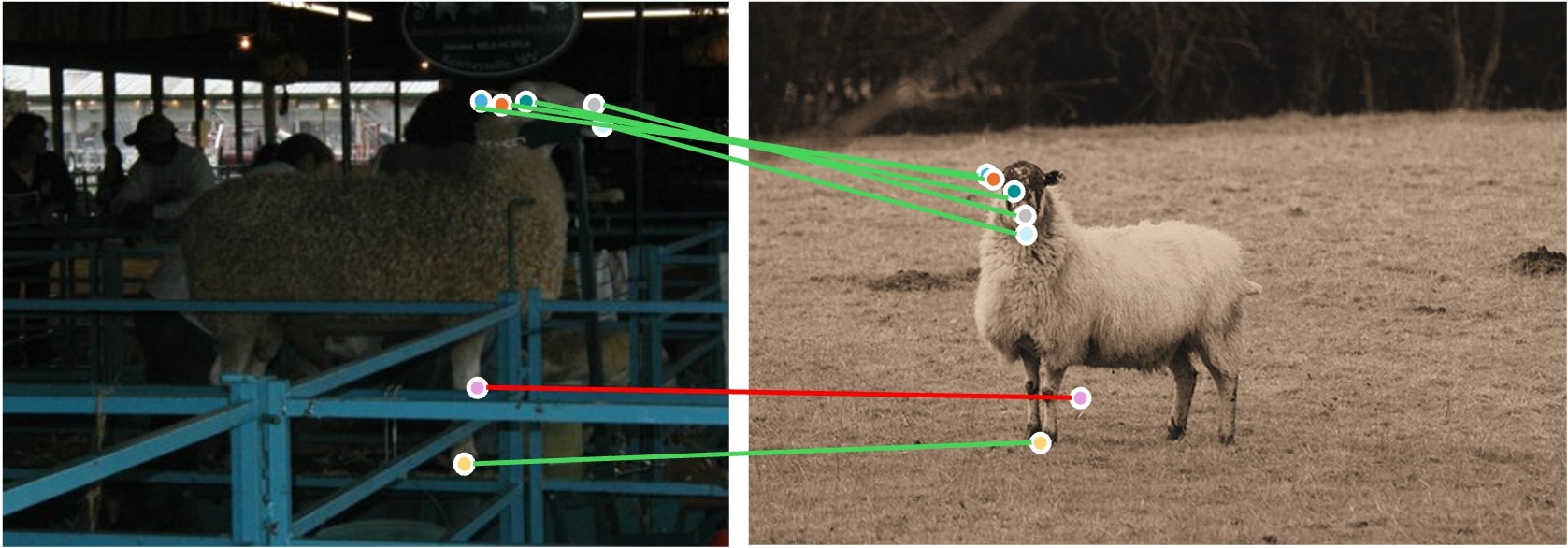} &
    \includegraphics[width=0.3\textwidth]{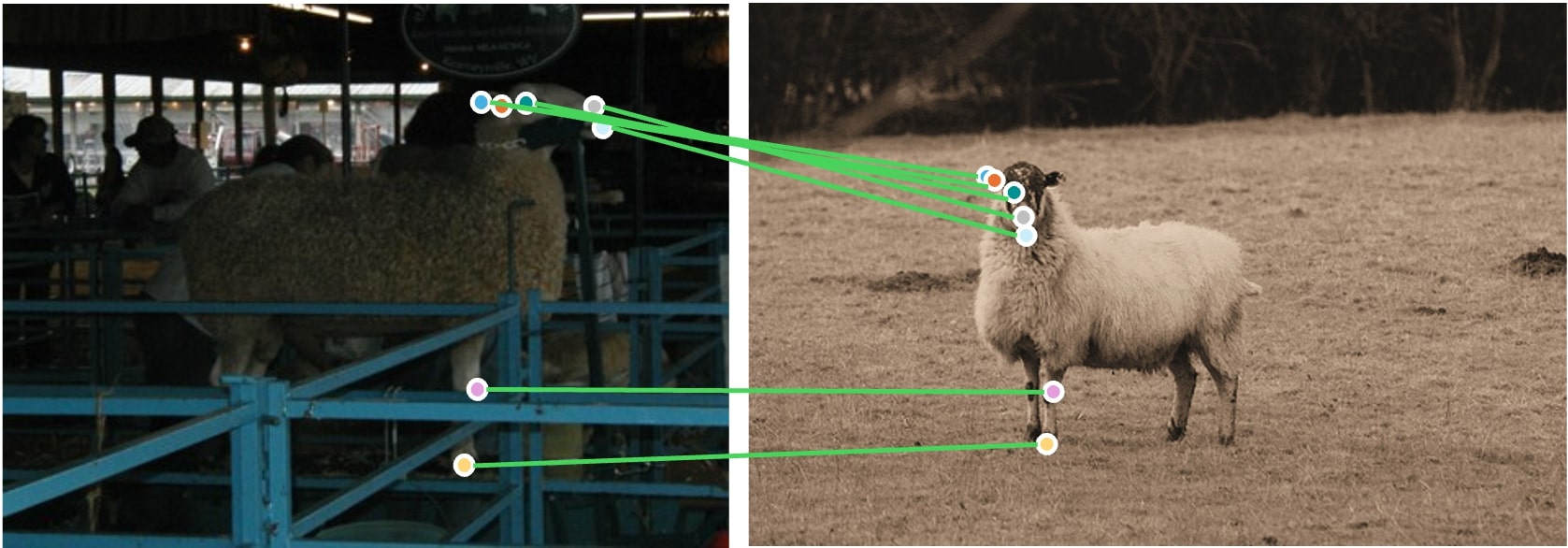} \\[2.5mm]

    \includegraphics[width=0.30\textwidth]{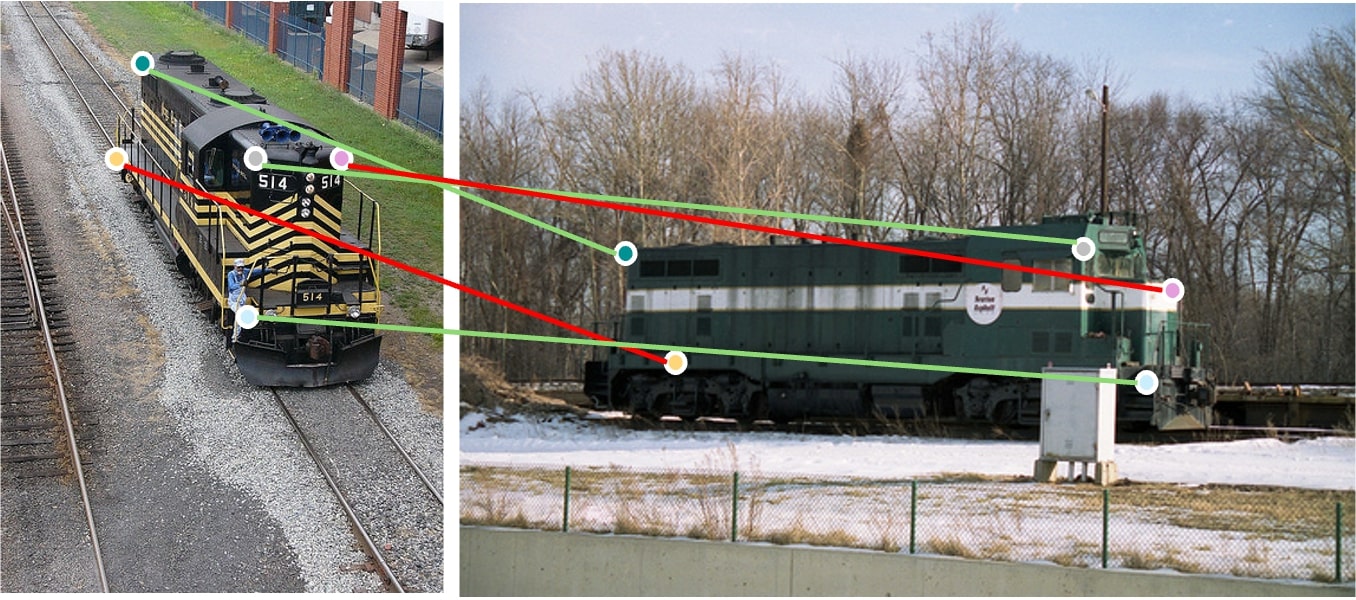} &
    \includegraphics[width=0.30\textwidth]{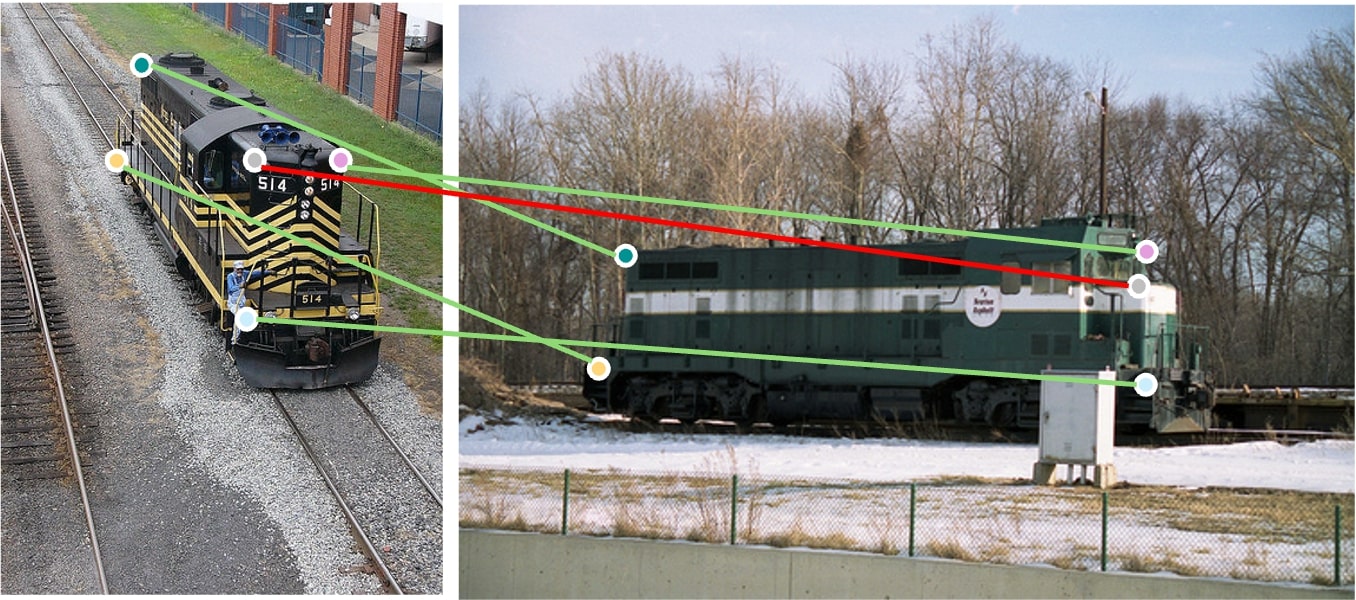} &
    \includegraphics[width=0.30\textwidth]{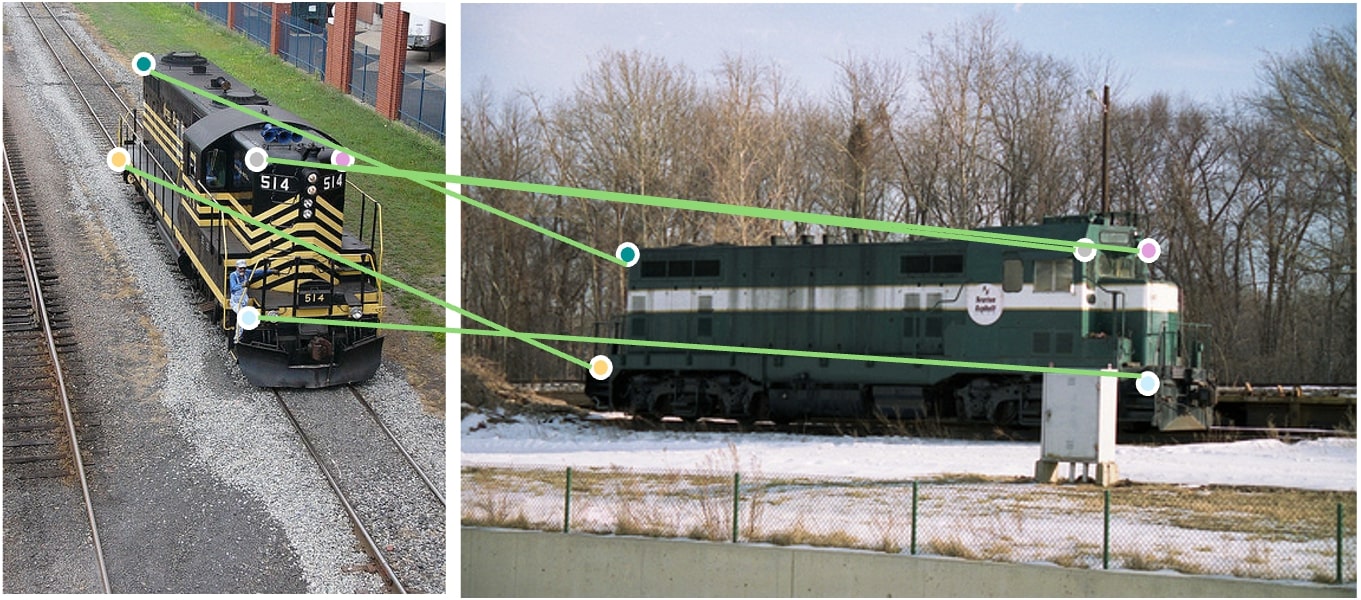} \\[2.5mm]

    \includegraphics[width=0.30\textwidth]{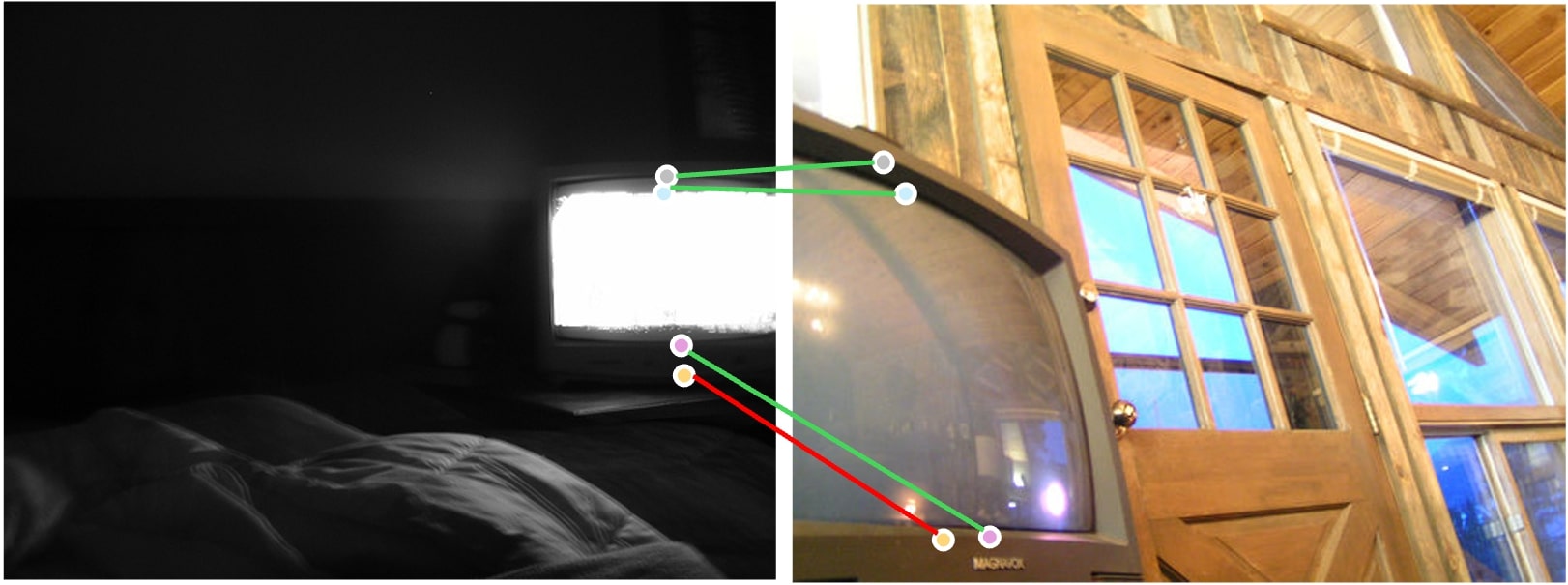} &
    \includegraphics[width=0.30\textwidth]{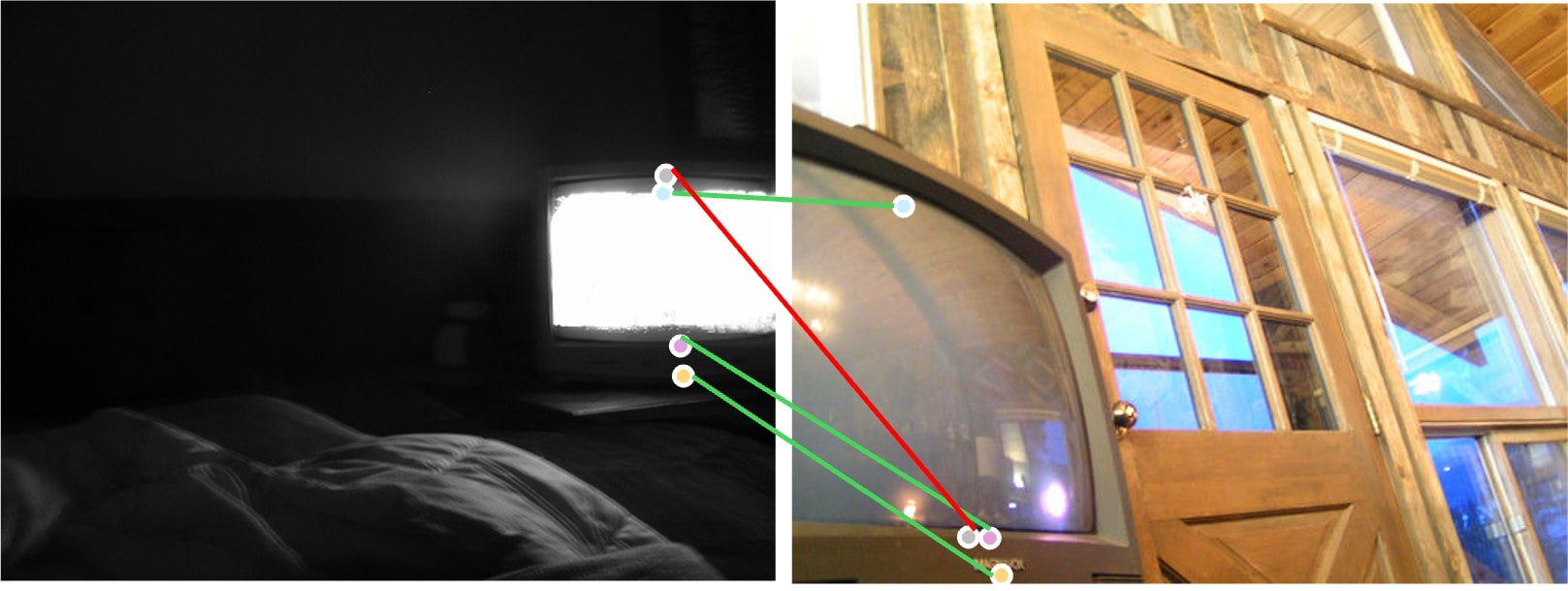} &
    \includegraphics[width=0.30\textwidth]{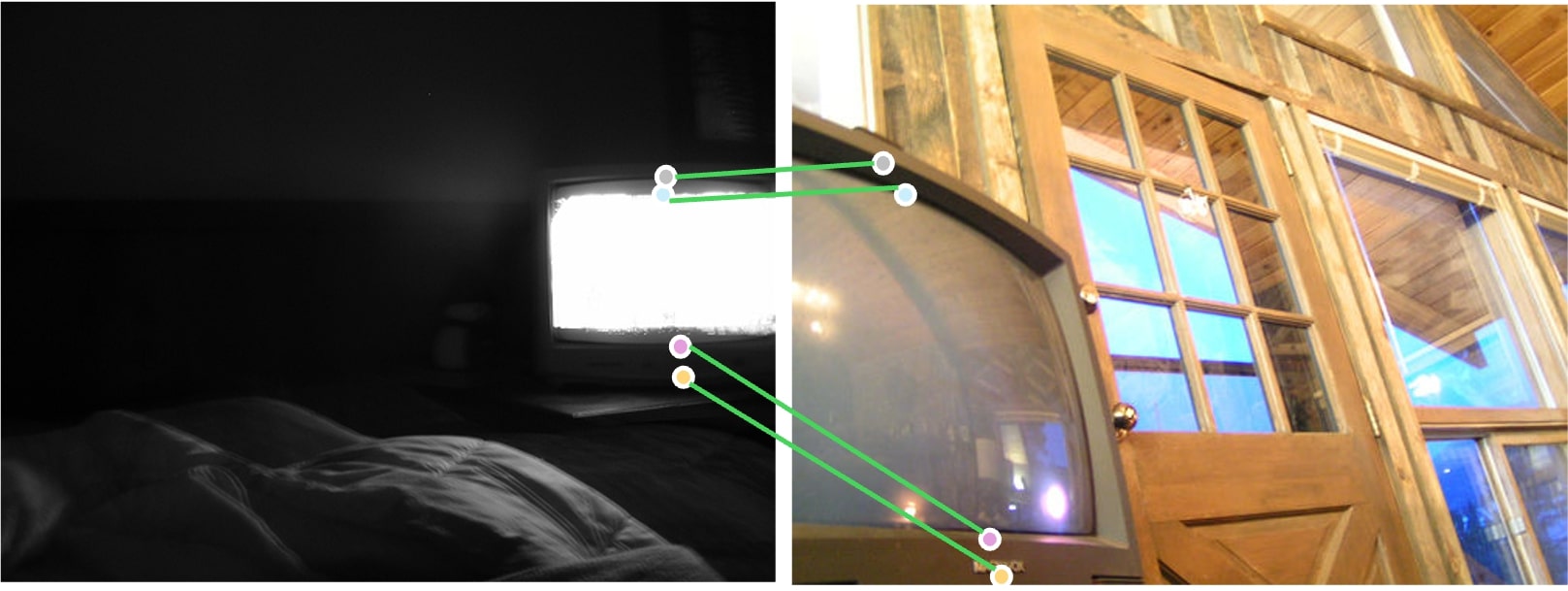} \\
  \end{tabular}

  \caption{Visual comparison of semantic correspondences on SPair-71k~\cite{min2019spair} across DistillDIFT~\cite{fundel2025distillation}, DINOv2+SD~\cite{zhang2023tale}, and our approach. Correct and incorrect matches are indicated by {\color{green}green lines} and {\color{red}red lines}, respectively.}
  \label{fig:uncurated_matches_2}
\end{figure*}

In this section, we present additional qualitative comparisons on SPair-71k~\cite{min2019spair} across all object categories. 
Figures~\ref{fig:uncurated_matches_1} and \ref{fig:uncurated_matches_2} visualize dense correspondences produced by DistillDIFT~\cite{fundel2025distillation}, DINOv2+SD~\cite{zhang2023tale}, and our method, where correct and incorrect matches are highlighted in green and red, respectively. 
Across a variety of categories, SoY tends to produce sharper and more globally consistent correspondences under large viewpoint, scale, and appearance changes, qualitatively complementing the quantitative improvements reported in the main paper.

\clearpage
\begin{algorithm*}[h!]
\caption{PyTorch-style pseudocode for our pseudo-label generation.}
\label{alg:pseudo-label}
\definecolor{codeblue}{rgb}{0.25,0.5,0.5}
\lstset{
  basicstyle=\fontsize{8pt}{8.8pt}\ttfamily\bfseries,
  commentstyle=\fontsize{8pt}{8.8pt}\color{codeblue},
  keywordstyle=\fontsize{8pt}{8.8pt},
  columns=fullflexible,
  keepspaces=true,
  breaklines=true,
  breakatwhitespace=true,
}
\begin{lstlisting}[language=python]
# Inputs:
#   F_A, F_B: feature maps of image A and B
#   V_A, V_B: 3D points (vertices) of image A and B
#   T:  number of refinement iterations
#   K:  number of anchors
#   alpha: feature-vs-geometry trade-off
#   rho:   KL strength in UOT
#   iters: #Sinkhorn iterations
# Output:
#   pi_T: final transport plan 

def unbalanced_sinkhorn(C, rho, iters):
    # 1) convert cost to log-kernel (soft affinity)
    Z = -C / rho                  # log K = - cost / rho
    m, n = C.shape

    # uniform marginals: mu_i = 1/m, nu_j = 1/n
    log_mu = torch.full((m,), -math.log(m))  # log(1/m)
    log_nu = torch.full((n,), -math.log(n))  # log(1/n)

    u = torch.zeros_like(log_mu)
    v = torch.zeros_like(log_nu)

    for _ in range(iters):
        # log-domain row / col updates (unbalanced)
        u = rho * (log_mu - torch.logsumexp(Z + v[None, :], dim=-1))
        v = rho * (log_nu - torch.logsumexp(Z + u[:, None], dim=-2))

    log_pi = Z + u[:, None] + v[None, :]
    pi = torch.exp(log_pi)        # final UOT plan
    return pi

# ---------- Stage 1: initial semantic matching ----------
# semantic cost from cosine similarity
C_sem = 1.0 - F_A @ F_B.T           # semantic cost matrix
pi = unbalanced_sinkhorn(C_sem, rho=rho, iters=iters)

# ---------- Pre-compute 3D distance matrices ----------
D_A = pairwise_dist(V_A)           # ||V_A[i] - V_A[j]||_2
D_B = pairwise_dist(V_B)           # ||V_B[i] - V_B[j]||_2

# ---------- Stage 2: iterative FGW refinement ----------
for t in range(1, T + 1):
    # 1) select 3D cycle-consistent mutual anchors
    anchors = select_anchors(pi, V_A, V_B, k=K)

    # 2) build geometric cost from anchors
    C_geo = torch.zeros_like(C_sem)
    for a_s, a_t in anchors:
        # distances to anchor on each shape
        dist_A = D_A[:, a_s][:, None].expand_as(C_geo)
        dist_B = D_B[:, a_t][None, :].expand_as(C_geo)
        # cycle-consistent structure cost
        C_geo += (dist_A - dist_B).abs()

    # 3) fuse normalized semantic & geometric costs
    C_sem_n = normalize(C_sem)     # scaling [0, 1]
    C_geo_n = normalize(C_geo)
    C_total = (1.0 - alpha) * C_sem_n + alpha * C_geo_n

    # 4) re-solve unbalanced OT with fused cost
    pi = unbalanced_sinkhorn(C_total, rho=rho, iters=iters)

pi_T = pi
\end{lstlisting}
\end{algorithm*}

\clearpage

\end{document}